\documentclass[lettersize,journal]{IEEEtran}
\usepackage{tikz}
\usepackage{amsmath}

\usepackage{filecontents}
\usepackage{soul}
\usepackage{graphicx}
\usepackage{subfigure}
\usepackage{amsmath}
\usepackage{amsthm}
\usepackage{mathrsfs}
\usepackage[vlined,ruled,linesnumbered,algo2e]{algorithm2e}
\usepackage{xcolor}
\usepackage{multicol}
\usepackage{multirow}
\usepackage{diagbox}
\usepackage{bbm}
\usepackage{makecell}
\usepackage{enumitem}
\usepackage{booktabs}
\usepackage{algorithm}
\usepackage{algorithmic}
\usepackage{hyperref}
\usepackage{amsfonts}
\usepackage{newtxmath}
\usepackage{cite}
\usepackage[numbers]{natbib}
\usepackage[most]{tcolorbox}

\begin{document}
%
\title{One Framework for All: Cross-Modal Membership Inference for Generative Models}

\author{Dayong Ye, Tianqing Zhu$^*$, Kun Gao, Junhao Liu, Yichuan Chen, \\ Shuai Zhou, Hengzhu Liu, Bo Liu, and Wanlei Zhou
\thanks{Tianqing Zhu is the corresponding author.}
}

\maketitle

\begin{abstract}
Large generative models across text-to-text, text-to-image, and image-to-text modalities have been shown to pose significant privacy risks. One fundamental threat is membership inference attacks (MIA), which aim to determine whether a given data point was used in a model’s training set. Although prior work has investigated MIAs against these three classes of generative models, existing approaches treat them in isolation and are not cross-applicable, thereby limiting their real-world utility. To address this limitation, we present the first comprehensive study of a unified membership inference framework that applies across text-to-text, text-to-image, and image-to-text modalities. Our approach is grounded in a key modality-agnostic observation: the output distribution of a generative model can approximate its training data distribution. Leveraging this property, we model the distributions of model-generated outputs and auxiliary non-member samples in a shared embedding space, and perform membership inference via likelihood ratio testing. We conduct extensive experiments in a strict black-box setting under both partial-knowledge and zero-knowledge threat models, and evaluate membership inference against both fine-tuning and pre-training data. Experimental results demonstrate our approach's superior performance in comparison to existing state-of-the-art methods, which are typically optimized for a single model class. 
\end{abstract}

\begin{IEEEkeywords}
Membership Inference, Generative Models, Cross Modality
\end{IEEEkeywords}

\section{Introduction}
\IEEEPARstart{L}{arge} generative models have become a core component of modern machine learning (ML) systems, enabling applications ranging from text generation and image synthesis to multi-modal assistants \cite{Xia24ECCV,Kara25ACL}. While these models are typically pre-trained on massive public corpora, their deployment often relies on fine-tuning with private or proprietary datasets to adapt them to downstream tasks. This practice raises significant privacy concerns, among which the risk of membership inference attacks stands out as a fundamental threat \cite{Shokri17Oakland}. 

Membership inference attacks (MIAs) aim to determine whether a given sample belongs to a model’s training set. Prior work has shown that large generative models inadvertently leak membership information through their outputs \cite{Ko23ICCV, Tran25ACL,Fu24NeurIPS,Pang25NDSS,Hu25USENIX}. However, existing studies mainly focus on a single class of generative models, for example, considering only text-to-text, text-to-image, or image-to-text models in isolation. This fragmentation raises an important question: \textbf{is there a unified membership inference approach that can effectively target all three classes of generative models?}
Addressing this question is crucial. In practice, modern AI systems integrate multiple generative components across different modalities \cite{Zhang24ACL,Sun24CVPR}, and privacy risks cannot be fully understood by analyzing each modality in isolation. From a research perspective, a unified approach would clarify whether membership leakage arises from modality-specific artifacts or from modality-agnostic properties inherent to generative modeling. Motivated by these considerations, our work aims to develop such a unified membership inference approach. 
However, achieving this aim presents \textbf{two challenges}.

First, the three classes of generative models, text-to-text, text-to-image, and image-to-text, rely on fundamentally different architectures and training paradigms. Designing a single attack strategy that applies across all these model classes requires techniques that are generalizable across heterogeneous systems.
Second, the data modalities associated with these generative models differ substantially. Text data are sequential and discrete, whereas image data are spatial and continuous. A unified attack framework must effectively accommodate these modality-specific characteristics to remain effective across different generative settings.

To address the first challenge, we leverage a fundamental property shared by generative models: their outputs are meaningful data points in the same domain as the training data. These generated samples can therefore be used to construct an output distribution that serves as an approximation of the training-data (i.e., member) distribution. By comparing a target sample against this approximate training-data distribution, we can infer its membership status.
To address the second challenge, we adopt a set of modality-appropriate feature extractors to map data samples into a shared numerical embedding space. The resulting embeddings are represented as numerical vectors that are independent of the underlying data modality, enabling unified and consistent downstream computations across multi-modal generative models.

In summary, this work has \textbf{three contributions}.
\begin{itemize}[leftmargin=*]
\item We initiate a unified study of membership inference attacks against generative models across multiple modalities, with the goal of uncovering modality-agnostic privacy leakage inherent to generative modeling.

\item Our approach is model-independent and lightweight. It exploits the fundamental property that the output distribution of a generative model approximates its training data distribution, and infers the membership status of target samples via likelihood ratio testing, without requiring the training of additional models. 

\item We conduct extensive experiments comparing our method with state-of-the-art baselines across a wide range of generative models and datasets, demonstrating consistently superior membership inference performance. Beyond inferring membership with respect to fine-tuning data, we further extend our evaluation to pre-training data, highlighting the generality and extensibility of our approach.

\end{itemize}

\section{Preliminaries and Threat Model}
\noindent\textbf{Generative Models.}
We focus on three representative categories of generative models: large language models (LLMs), diffusion models (DMs), and vision-language models (VLMs), which correspond to the core application domains of text-to-text, text-to-image, and image-to-text, respectively. 

\vspace{1mm}
\noindent\textbf{Large language models} \cite{GPT4} define a likelihood distribution over token sequences. Given a text record $x = [t_1, \dots, t_{|x|}]$ of length $|x|$, an LLM estimates the conditional probability of each token based on its preceding context: $p_{\theta}(t_i|t_1,...,t_{i-1})$, where $\theta$ denotes the model parameters. By applying the chain rule, the joint probability of the entire sequence is factorized into the product of these conditional probabilities. Accordingly, LLMs are trained to minimize the negative log-likelihood of the training corpus:
\begin{equation}\nonumber
    \mathcal{L}_{LLM}=-\frac{1}{N}\sum^N_{j=1}\sum^{|x^{(j)}|}_{i=1}\mathrm{log} p_{\theta}(t_i|t_1,...,t_{i-1}),
\end{equation}
where $N$ is the number of training records. 

During generation, the model produces tokens one at a time in an autoregressive manner, thereby constructing coherent text sequences. From the user’s perspective, an LLM takes a text record $x$ (e.g., a prompt) as input and returns a corresponding text response $y$, denoted as $y = LLM(x)$.

\vspace{1mm}
\noindent\textbf{Diffusion models} \cite{DiffusionModel} generate data by gradually reversing a noise-adding process. An image is corrupted by incrementally adding Gaussian noise, and the model is trained to recover the original image by denoising step by step. Formally, given an image $x_0$, the forward diffusion process adds noise over $T$ steps. At time step $t$, the noisy image $x_t$ is given by:
\begin{equation}\nonumber
    x_t=\sqrt{\overline{\alpha}_t}x_0+\sqrt{1-\overline{\alpha}_t}\epsilon_t,
\end{equation}
where $\overline{\alpha}_t = \prod_{i=1}^t \alpha_i$, each $\alpha_i \in (0,1)$ is a predefined parameter controlling the noise schedule, and $\epsilon_t$ is Gaussian noise obtained via the reparameterization trick.

The reverse diffusion process aims to invert this corruption. Starting from $\hat{x}_T=x_T$, the model denoises $\hat{x}_t$ to recover $\hat{x}_{t-1}$. A neural network $\mathcal{M}_\theta$ is trained to predict the removed noise at each step. The training objective is therefore:
\begin{equation}
    \mathcal{L}_{DM}=\mathbb{E}[||\epsilon_t-\mathcal{M}_{\theta}(\sqrt{\overline{\alpha}_t}x_0+\sqrt{1-\overline{\alpha}_t}\epsilon_t,t)||^2_2].
\end{equation}

Conditional diffusion models extend the standard diffusion framework to generate high-quality images guided by text prompts. Given a prompt $x$, the diffusion model conditions the denoising process on a learned representation of $x$, ensuring that the generated sample aligns semantically with the input prompt.
Formally, let $x_0$ denote the original image and $z_x$ denote the embedding of the prompt $x$. The forward process remains unchanged, namely Gaussian noise is gradually added to $x_0$ over $T$ steps, producing noisy intermediate states $x_t$. The reverse process is then parameterized by a neural network $\mathcal{M}_\theta$, which predicts the noise at each step conditioned not only on the noisy image $x_t$, but also on the prompt embedding $z_x$: $\epsilon_t\approx\mathcal{M}_\theta(x_t,t,z_x)$. The training objective becomes:
\begin{equation}
    \mathcal{L}_{CDM}=\mathbb{E}[||\epsilon_t-\mathcal{M}_\theta(x_t,t,z_x)||^2_2].
\end{equation}
From the user’s perspective, a conditional diffusion model takes a prompt $x$ as input and produces an image $y$ as output, i.e., $y = DM(x)$.

\vspace{1mm}
\noindent\textbf{Vision language models} \cite{VLModel} integrate visual and textual inputs to perform multimodal reasoning. An input image $x_v$ is first processed by a vision encoder to extract its visual feature representation $e_v$. This feature is then projected into the language model's embedding space, yielding $t_v$. In parallel, a text prompt $x_q$ is tokenized into a sequence of text embeddings $t_q$. The combined token sequence $[t_v, t_q]$ is subsequently fed into an LLM, which generates the final text response $y_a$.

The training dataset of a VLM is typically represented as $D = {(x_v^i, x_q^i, y_a^i)}_{i=1}^N$, where each triplet consists of an image, a text query, and a corresponding textual answer. The learning objective is to maximize the likelihood of the model generating $y_a$ given the paired inputs $x_v$ and $x_q$, i.e.,
\begin{equation}\nonumber
    \mathcal{L}_{VLM}=-\frac{1}{N}\sum^N_{i=1}\mathrm{log}p_\theta(y^i_a|x^i_v,x^i_q).
\end{equation}
From the user's perspective, a VLM takes an image $x_v$ together with a text prompt $x_q$ as input, and outputs a text response $y$, formally expressed as $y = VLM(x_v, x_q)$.

\vspace{2mm}
\noindent\textbf{Membership Inference.} The goal of membership inference is to determine whether a given data sample $x$ was included in the training set $D_{\mathrm{train}}$ of a target model $\mathcal{G}$. Formally, the membership inference task can be expressed as:
\begin{equation}\nonumber
    \mathcal{A}_{D_{\mathrm{train}}}: (x,\mathcal{G})\rightarrow\{0,1\},
\end{equation}
where $\mathcal{A}_{D_{\mathrm{train}}}$ outputs $1$ if $x \in D_{\mathrm{train}}$ and $0$ otherwise.

Unlike conventional models, membership inference against generative models also applies to the fine-tuning set $D_{\mathrm{tune}}$. This distinction arises because many generative models, especially open-source ones, are first trained on large public datasets and then fine-tuned on smaller, often private, datasets tailored to specific downstream tasks. In this setting, the inference problem becomes: 
\begin{equation}\nonumber
    \mathcal{A}_{D_{\mathrm{tune}}}: (x,\mathcal{G})\rightarrow\{0,1\},
\end{equation}
where $\mathcal{A}_{D_{\mathrm{tune}}}$ outputs $1$ if $x \in D_{\mathrm{tune}}$ and $0$ otherwise.

From a privacy perspective, inferring membership in the fine-tuning set poses a greater threat than inferring membership in the pre-training data. This is because pre-training datasets are typically collected from public sources with relatively lower privacy sensitivity, while fine-tuning datasets often consist of proprietary or sensitive information. For this reason, our work primarily focuses on membership inference against fine-tuning data $D_{\mathrm{tune}}$. Nevertheless, as we will demonstrate experimentally, our proposed method is also applicable to inferring membership in pre-training data $D_{\mathrm{train}}$.

\vspace{2mm}
\noindent\textbf{Threat Model.} We consider the most stringent black-box setting, where the adversary can only query the target generative model $\mathcal{G}$ without accessing or modifying its internal parameters, and can observe only the model's generated outputs, not its logits or intermediate representations. Formally, given a target sample $x^*$, the adversary queries $\mathcal{G}$ and obtains the corresponding response $y^*$. Assuming that $\mathcal{G}$ has been fine-tuned on a dataset $D_{\mathrm{tune}}$, the adversary's goal is to determine whether $x^* \in D_{\mathrm{tune}}$. The adversary knows the task domain of $D_{\mathrm{tune}}$ but has no access to the dataset itself. Note that a training sample typically consists of an input–output pair $(x, y)$, where $y$ denotes the ground-truth output. For simplicity of expression, we slightly abuse notation and use $x$ to denote the corresponding input–output pair when the context is clear. 

For the capability of collecting external data, we distinguish between two adversarial scenarios:
\begin{itemize}[leftmargin=*]
    \item Partial knowledge. The adversary has the ability to gather data from real-world sources. In this case, the adversary constructs a dataset $D_{\mathrm{real}}$ that is drawn from the same task domain as the fine-tuning dataset $D_{\mathrm{tune}}$ of $\mathcal{G}$, but does not overlap with $D_{\mathrm{tune}}$. 
    \item Zero knowledge. The adversary has no access to any external data beyond the target generative model itself. This setting represents the most restrictive scenario, where real-world data relevant to $\mathcal{G}$ are either inaccessible due to privacy restrictions, or simply do not exist in a usable form.
\end{itemize}
We design tailored inference strategies for each scenario. 

\section{Membership Inference with Partial Knowledge}

\begin{figure}[ht]
\centering
	\includegraphics[scale=0.3]{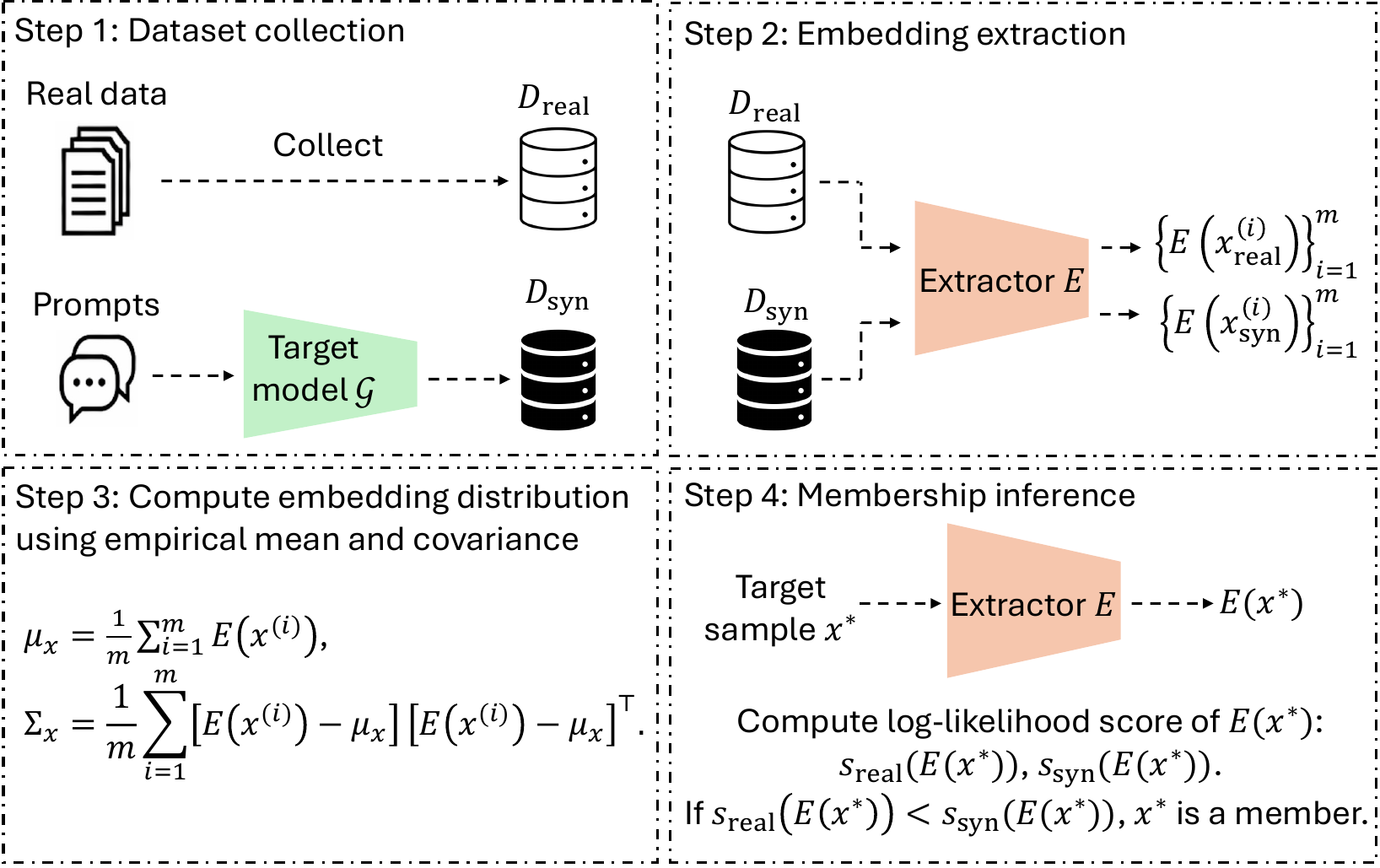}
	\caption{Overview of our attack strategy under the partial-knowledge setting.
It consists of four steps: (1) collecting an auxiliary dataset from the real world and generating a synthetic dataset using the target model; (2) extracting embeddings from both datasets; (3) estimating the corresponding embedding distributions; and (4) inferring the membership status of a target sample by comparing its embedding against the two distributions, where greater proximity to the synthetic distribution indicates higher likelihood of membership.}
    \vspace{-0mm}
	\label{fig:Overview1}
\end{figure}

This section introduces the inference approach, which applies when the adversary has partial knowledge of the target generative model's fine-tuning dataset $D_{\mathrm{tune}}$ and can collect an auxiliary dataset $D_{\mathrm{real}}$ from real-world sources. This approach consists of four steps summarized in Figure \ref{fig:Overview1}.

\vspace{1mm}
\noindent\textbf{Step 1: Dataset Collection.} Given partial knowledge of the fine-tuning dataset $D_{\mathrm{tune}}$, the adversary collects an auxiliary dataset $D_{\mathrm{real}}$ drawn from the same task domain but without overlapping with $D_{\mathrm{tune}}$. In parallel, the adversary queries the target generative model $\mathcal{G}$ to construct a synthetic dataset $D_{\mathrm{syn}}$. 
For example, suppose $\mathcal{G}$ is a text-to-image diffusion model fine-tuned on a private medical dataset of chest X-rays paired with diagnostic captions. The adversary, while not having access to $D_{\mathrm{tune}}$, can still gather publicly available chest X-ray images from open datasets such as NIH ChestX-ray \cite{Wang17CVPR} to form $D_{\mathrm{real}}$. At the same time, the adversary queries $\mathcal{G}$ with generic medical prompts (e.g., ``X-ray of lungs with mild abnormalities'') to generate $D_{\mathrm{syn}}$. 

\vspace{1mm}
\noindent\textbf{Step 2: Embedding Extraction.} For each data point in the real and synthetic datasets, i.e., $x^{(i)}_{\mathrm{real}} \in D_{\mathrm{real}}$ and $x^{(i)}_{\mathrm{syn}} \in D_{\mathrm{syn}}$, the adversary feeds the sample into an encoder $E$ to obtain its embedding representation, denoted as $E(x^{(i)}_{\mathrm{real}})$ and $E(x^{(i)}_{\mathrm{syn}})$, respectively.
The choice of encoder $E$ is aligned with the architecture and training objective of the target generative model $\mathcal{G}$. Specifically, the encoder should operate in the same semantic space that $\mathcal{G}$ relies on for generation, so that the extracted embeddings faithfully capture the model-relevant features.
For example, if $\mathcal{G}$ is a text-to-image diffusion model (e.g., Stable Diffusion), we adopt the BLIP \cite{Li22ICML} encoder as $E$. This choice is motivated by the fact that BLIP is trained with joint vision–language objectives and is designed to align visual content with its corresponding textual semantics.

\vspace{1mm}
\noindent\textbf{Step 3: Compute Embedding Distribution.} We characterize the embedding distribution by computing the empirical mean vector and covariance matrix of the extracted embeddings. Specifically, given $m$ sampled embeddings $\{E(x^{(i)})\}_{i=1}^m$, the empirical mean and covariance are defined as:
\begin{equation}\nonumber
\begin{aligned}
    \mu_x&=\frac{1}{m}\sum^m_{i=1}E(x^{(i)});\\
    \Sigma_x&=\frac{1}{m}\sum^m_{i=1}[E(x^{(i)})-\mu_x][E(x^{(i)})-\mu_x]^{\top}.
\end{aligned}
\end{equation}
Applying this procedure to the embeddings extracted from the real and synthetic datasets yields their respective distributional representations, namely $(\mu_{x_{\mathrm{real}}}, \Sigma_{x_{\mathrm{real}}})$ for real data and $(\mu_{x_{\mathrm{syn}}}, \Sigma_{x_{\mathrm{syn}}})$ for synthetic data.

We now justify why the empirical mean and covariance are sufficient to represent the embedding distribution. From a probabilistic perspective, let $z = E(x)$ denote the embedding of $x$. The distribution of $z$ is characterized by its true mean vector $\mu = \mathbb{E}[z]$ and true covariance matrix $\Sigma = \mathbb{E}[(z-\mu)(z-\mu)^\top]$. By the law of large numbers, we have 
\begin{equation}\nonumber
    \frac{1}{m}\sum^m_{i=1}E(x^{(i)})\xrightarrow[m\rightarrow\infty]{a.s.}\mathbb{E}[z]=\mu,
\end{equation}
which guarantees that the empirical mean $\mu_x$ converges almost surely to the true mean $\mu$ as the number of queries $m$ increases. Similarly, by the multivariate strong law of large numbers, the empirical covariance $\Sigma_x$ converges almost surely to the true covariance $\Sigma$:
\begin{equation}\nonumber
    \frac{1}{m}\sum^m_{i=1}[E(x^{(i)})-\mu_x][E(x^{(i)})-\mu_x]^{\top}\xrightarrow[m\rightarrow\infty]{a.s.}\Sigma.
\end{equation}
Moreover, the central limit theorem states that the sampling distribution of $\mu_x$ approaches a multivariate normal distribution centered at $\mu$ with covariance $\Sigma/m$. This implies that the estimation error for $\mu_x$ decreases at a rate of $1/m$, meaning that the uncertainty in estimating the true mean shrinks rapidly as $m$ grows. 

More specifically, a non-asymptotic bound is provided by Hoeffding’s inequality. If each coordinate $z_j$ of the embedding vector $z$ is bounded within $[a_j, b_j]$, the sample mean satisfies:
\begin{equation}\nonumber
    \mathrm{Pr}(||\mu_x-\mu||_{\infty}>\epsilon)\leq\sum^d_{j=1}2\mathrm{exp}[-\frac{2m\epsilon^2}{(b_j-a_j)^2}],
\end{equation}
where $\epsilon$ denotes the tolerance for estimation error and $d$ is the dimensionality of $z$. This inequality shows that the probability of a large deviation decays exponentially as the number of queries $m$ increases. 
This result establishes that both $\mu_x$ and $\Sigma_x$ are statistically consistent estimators of the true embedding distribution. In practice, this means that even with a moderate number of queries, the adversary can obtain a reliable approximation of the embedding distribution. 

\vspace{1mm}
\noindent\textbf{Step 4: Membership Inference.} Given a target sample $x^*$, the adversary feeds it into the encoder $E$ and obtains its embedding representation $E(x^*)$\footnote{Strictly speaking, a training sample consists of an input–output pair $(x^*, y^*)$. In practice, the adversary applies the encoder to $y^*$ and obtains $E(y^*)$. However, for notational consistency, we write this as $E(x^*)$.}. Then, given an embedding distribution $(\mu,\Sigma)$, the adversary computes the log-likelihood score of $E(x^*)$ under a multivariate Gaussian assumption \cite{Lee18NIPS}:
\begin{equation}\label{eq:score}
    s(E(x^*))=-\frac{1}{2}[(E(x^*)-\mu)^\top\Sigma^{-1}(E(x^*)-\mu)+\mathrm{log}|\Sigma|].
\end{equation}
Using Eq. \ref{eq:score}, the adversary evaluates two likelihood scores: $s_{\mathrm{real}}(E(x^*))$, computed under the real-data distribution $(\mu_{x_{\mathrm{real}}}, \Sigma_{x_{\mathrm{real}}})$, and $s_{\mathrm{syn}}(E(x^*))$, computed under the synthetic-data distribution $(\mu_{x_{\mathrm{syn}}}, \Sigma_{x_{\mathrm{syn}}})$.
The membership inference decision rule is then defined as: 
\begin{equation}\label{eq:rule}
    s_{\mathrm{real}}(E(x^*))<s_{\mathrm{syn}}(E(x^*)).
\end{equation}
If the inequality in Eq. \ref{eq:rule} holds, the sample $x^*$ is inferred to be a member of the fine-tuning dataset; otherwise, it is classified as a non-member.
We now justify why this decision rule yields reliable membership inference. 

This decision rule can be interpreted as a likelihood-ratio test between two hypotheses:

\vspace{1mm}
\noindent $\mathbf{H_0}$: $x^*$ is a non-member, and its embedding is drawn from the real-data distribution;

\noindent $\mathbf{H_1}$: $x^*$ is a member, and its embedding is drawn from the synthetic-data distribution induced by the fine-tuned model.

\vspace{1mm}
Under the Gaussian approximation, Eq. \ref{eq:rule} is equivalent to comparing the log-likelihood ratio
\begin{equation}\nonumber
    \mathrm{log}\frac{p[E(x^*)|\mu_{x_{\mathrm{syn}}},\Sigma_{x_{\mathrm{syn}}}]}{p[E(x^*)|\mu_{x_{\mathrm{real}}},\Sigma_{x_{\mathrm{real}}}]}
\end{equation}
against zero, which is the optimal decision rule in the Neyman–Pearson sense when both class-conditional distributions are known.
The rationale behind this test is that samples in the fine-tuning dataset are expected to induce outputs that are better aligned with the generative model’s learned distribution, and thus yield embeddings with higher likelihood under the synthetic distribution $(\mu_{x_{\mathrm{syn}}}, \Sigma_{x_{\mathrm{syn}}})$ than under the real-data distribution $(\mu_{x_{\mathrm{real}}}, \Sigma_{x_{\mathrm{real}}})$. In contrast, non-members are more likely to follow the real-data distribution and therefore achieve higher likelihood under $(\mu_{x_{\mathrm{real}}}, \Sigma_{x_{\mathrm{real}}})$.

\section{Membership Inference with Zero Knowledge}
\begin{figure}[ht]
\centering
	\includegraphics[scale=0.3]{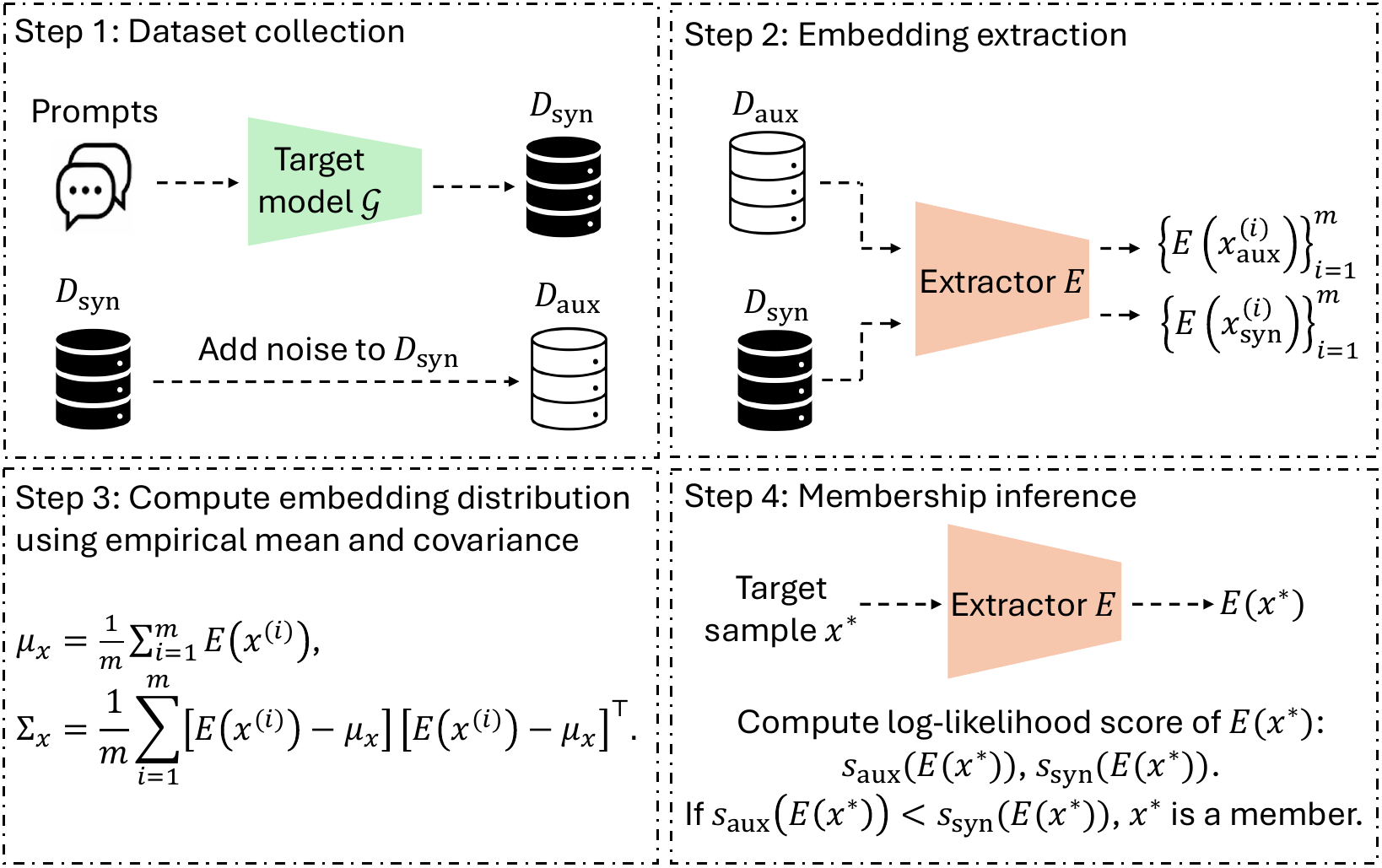}
	\caption{Overview of our attack strategy under the zero knowledge setting: (1) generating a synthetic dataset using the target model and constructing an auxiliary dataset by perturbing synthetic samples; (2) extracting embeddings from both datasets; (3) estimating the corresponding embedding distributions; and (4) inferring the membership status of a target sample by comparing its embedding against the two distributions, where greater proximity to the synthetic distribution indicates higher likelihood of membership.}
    \vspace{-0mm}
	\label{fig:Overview2}
\end{figure}

This section introduces the membership inference approach for the zero-knowledge scenario. The adversary is aware only of the general domain of the fine-tuning dataset $D_{\mathrm{tune}}$ of the target model and cannot access real-world data from that domain. As in the partial-knowledge setting, the proposed approach consists of four steps summarized in Figure \ref{fig:Overview2}. The key distinction lies in the first step, while the remaining three steps are identical to those in the partial-knowledge scenario. 

\begin{figure}[ht]
\centering
	\includegraphics[scale=0.34]{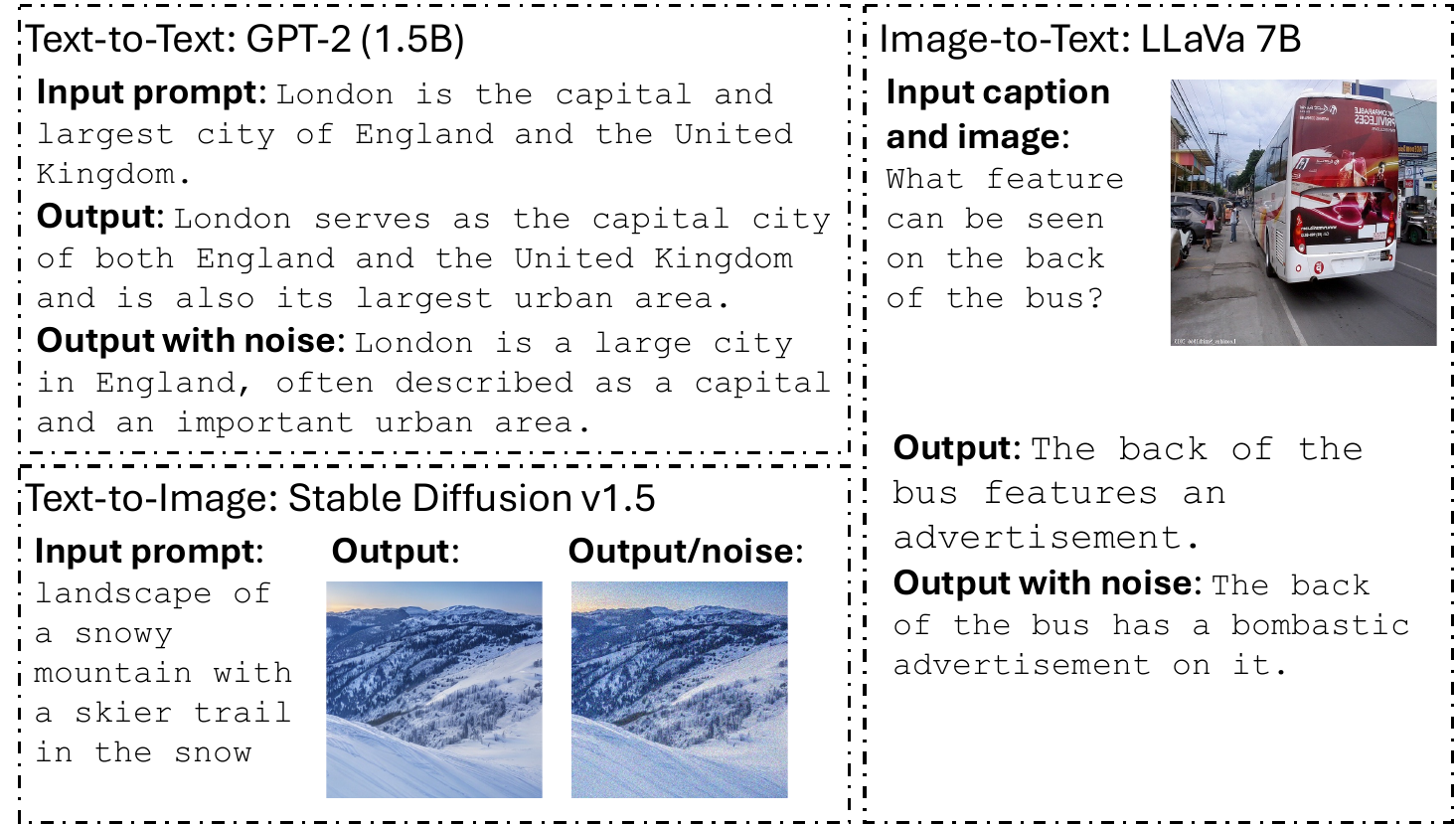}
	\caption{Examples of auxiliary data construction under three classes of generative models.}
    \vspace{-0mm}
	\label{fig:NoiseExample}
\end{figure}

\vspace{1mm}
\noindent\textbf{Step 1: Auxiliary Set Creation.} In the zero-knowledge setting, the adversary cannot collect any external data from the real world. Instead, based solely on the domain knowledge of $D_{\mathrm{tune}}$, the adversary queries the target generative model $\mathcal{G}$ to construct a synthetic dataset $D_{\mathrm{syn}}$. For example, if the domain of $D_{\mathrm{tune}}$ consists of news articles, the adversary can issue generic prompts such as “Write a short news article” to induce $\mathcal{G}$ to generate news-related samples, thereby approximating the distribution of $D_{\mathrm{tune}}$ without requiring any detailed knowledge of the data in $D_{\mathrm{tune}}$. Subsequently, an auxiliary dataset $D_{\mathrm{aux}}$ is generated by perturbing samples in $D_{\mathrm{syn}}$ with controlled noise. For image data, the adversary applies small transformations such as Gaussian noise; for text data, the adversary generates paraphrases via back-translation or synonym replacement, which preserve semantic meaning while altering token sequences. The goal of this process is to produce auxiliary samples that (i) remain semantically close to those in $D_{\mathrm{syn}}$, thereby probing the same local region of the data manifold, and (ii) are extremely unlikely to be members of the target fine-tuning dataset $D_{\mathrm{tune}}$. Figure~\ref{fig:NoiseExample} illustrates representative examples of auxiliary data construction across the three classes of generative models.

We now analyze the probability that auxiliary samples are non-members of $D_{\mathrm{tune}}$. Suppose $D_{\mathrm{tune}}$ contains $n$ examples drawn from some unknown data distribution. Let $p$ denote the probability that a single random draw from this distribution exactly matches an auxiliary sample $x'$. Although $p$ is unknown, probability theory provides useful guidance: in continuous domains, exact equality implies $p=0$, while in discrete domains, $p$ is nonzero but extremely small. Under this formulation, the probability that $x'$ is not in $D_{\mathrm{tune}}$ is:
\begin{equation}\nonumber
    \mathrm{Pr}(x'\notin D_{\mathrm{tune}})=(1-p)^n.
\end{equation}
When $p \ll 1$ and $np$ is moderate, it can be approximated as:
\begin{equation}\nonumber
    \mathrm{Pr}(x'\notin D_{\mathrm{tune}})\approx e^{-np}.
\end{equation}
If the adversary generates $m$ independent auxiliary samples ${x'_1, \dots, x'_m}$ with corresponding match probabilities $p_1, \dots, p_m$, the probability that none of the $m$ samples appear in $D_{\mathrm{tune}}$ is:
\begin{equation}\label{eq:probability}
    \mathrm{Pr}(x'_i\notin D_{\mathrm{tune}}|\forall i)=\prod^m_{i=1}(1-p_i)^n\approx\mathrm{exp}(-n\sum^m_{i=1} p_i).
\end{equation}
This result shows that the non-membership probability decreases as $n$, $m$, or any $p_i$ increase. Since $n$ (the size of $D_{\mathrm{tune}}$) and $p_i$ (the match probability) are neither known nor controllable by the adversary, the only adjustable parameter is $m$. To maintain a high probability of non-membership, $m$ should therefore be kept small. In practice, we treat $m$ as a tunable hyperparameter in the experiments. 

\vspace{1mm}
\noindent\textbf{Step 2: Embedding Extraction.} This step follows the same procedure as in the partial-knowledge scenario, except that the real-world dataset $D_{\mathrm{real}}$ is replaced with the auxiliary dataset $D_{\mathrm{aux}}$. Specifically, the adversary extracts embedding representations for samples from both $D_{\mathrm{aux}}$ and $D_{\mathrm{syn}}$, denoted as $E(x^{(i)}_{\mathrm{aux}})$ and $E(x^{(i)}_{\mathrm{syn}})$, respectively.

\vspace{1mm}
\noindent\textbf{Step 3: Compute Embedding Distribution.} This step follows the same computation procedure as in the partial-knowledge scenario. Applying it to the embeddings $E(x^{(i)}_{\mathrm{aux}})$ and $E(x^{(i)}_{\mathrm{syn}})$ yields their respective distributional representations: $(\mu_{x_{\mathrm{aux}}}, \Sigma_{x_{\mathrm{aux}}})$ for auxiliary data and $(\mu_{x_{\mathrm{syn}}}, \Sigma_{x_{\mathrm{syn}}})$ for synthetic data.

\vspace{1mm}
\noindent\textbf{Step 4: Membership Inference.} This step follows the same procedure as in the partial-knowledge setting. Using Eq.~\ref{eq:score}, the adversary computes two log-likelihood scores for the target embedding: $s_{\mathrm{aux}}(E(x^*))$ under the auxiliary data distribution $(\mu_{x_{\mathrm{aux}}}, \Sigma_{x_{\mathrm{aux}}})$ and $s_{\mathrm{syn}}(E(x^*))$ under the synthetic data distribution $(\mu_{x_{\mathrm{syn}}}, \Sigma_{x_{\mathrm{syn}}})$. The adversary then infers that $x^*$ is a member if $s_{\mathrm{aux}}(E(x^*)) < s_{\mathrm{syn}}(E(x^*))$; otherwise, $x^*$ is classified as a non-member.

\section{Theoretical Analysis}
Our analysis consists of three components: (i) why the output distribution of a generative model can approximate its training data distribution, (ii) why the log-likelihood score provides a  basis for making membership inference decisions, and (iii) why the decision threshold in Eq.~\ref{eq:rule} is implicitly set to zero, i.e., $s_{\mathrm{real}}(E(x^*))-s_{\mathrm{syn}}(E(x^*))<0$. Note that we focus the analysis on the partial-knowledge setting. The analysis for the zero-knowledge setting is similar.

\vspace{1mm}
\noindent\textbf{(i) Training Data Distribution Approximation.} Training data typically consist of pairs of input prompts and corresponding ground-truth responses. Therefore, our analysis focuses on the approximation between the distribution of ground-truth responses and that of the model’s generated responses, rather than the relationship between the model’s input prompts and output responses. Moreover, this analysis naturally extends to the fine-tuning setting. This is because both pre-training and fine-tuning share the same underlying mechanism: likelihood-based optimization encourages the model to approximate the empirical distribution of the data it is trained on, regardless of whether the data come from pre-training or fine-tuning. 

Given a generative model $\mathcal{G}_\theta$, let $p_{\mathrm{train}}$ denote the distribution of its training data. Training a likelihood-based generative model typically amounts to minimizing the empirical negative log-likelihood:
\[
L(\theta) = - \frac{1}{n} \sum_{i=1}^n \mathrm{log} p_\theta(x_i).
\]
In expectation, this objective corresponds to
\[
\mathbb{E}_{x \sim p_{\mathrm{train}}}[-\mathrm{log} p_\theta(x)] = \int p_{\mathrm{train}}(x)(-\mathrm{log} p_\theta(x)) \mathrm{d}x.
\]
This quantity measures how well the model’s output distribution $p_\theta$ explains samples drawn from the true training data distribution $p_{\mathrm{train}}$.
By adding and subtracting $\mathrm{log} p_{\mathrm{train}}(x)$ inside the expectation, we obtain
\begin{equation}\nonumber
\begin{aligned}
    &\mathbb{E}_{x \sim p_{\mathrm{train}}}[-\mathrm{log} p_\theta(x)]\\
    &=\mathbb{E}_{x \sim p_{\mathrm{train}}}[-\mathrm{log} p_\theta(x) + \mathrm{log} p_{\mathrm{train}}(x) - \mathrm{log} p_{\mathrm{train}}(x)]\\
    &=\mathbb{E}_{x \sim p_{\mathrm{train}}}[- \mathrm{log} p_{\mathrm{train}}(x)]\cdot\mathbb{E}_{x \sim p_{\mathrm{train}}}[\mathrm{log} p_{\mathrm{train}}(x)-\mathrm{log} p_\theta(x)],
\end{aligned}
\end{equation}
where the first term is exactly the entropy of the training data distribution and is independent of $p_\theta$, while the second term corresponds to the Kullback–Leibler (KL) divergence:
\[
\mathbb{E}_{x \sim p_{\mathrm{train}}}[\mathrm{log} p_{\mathrm{train}}(x)-\mathrm{log} p_\theta(x)]=D_{\mathrm{KL}}(p_{\mathrm{train}}||p_\theta).
\]
Since the entropy term is constant with respect to $p_\theta$, minimizing the expected negative log-likelihood is equivalent to minimizing $D_{\mathrm{KL}}(p_{\mathrm{train}} || p_\theta)$. Consequently, likelihood-based training drives the model’s output distribution $p_\theta$ toward the training data distribution $p_{\mathrm{train}}$, i.e., $D_{\mathrm{KL}}(p_{\mathrm{train}} || p_\theta) \rightarrow 0$. 

\begin{figure}[ht]
\centering
	\includegraphics[scale=0.42]{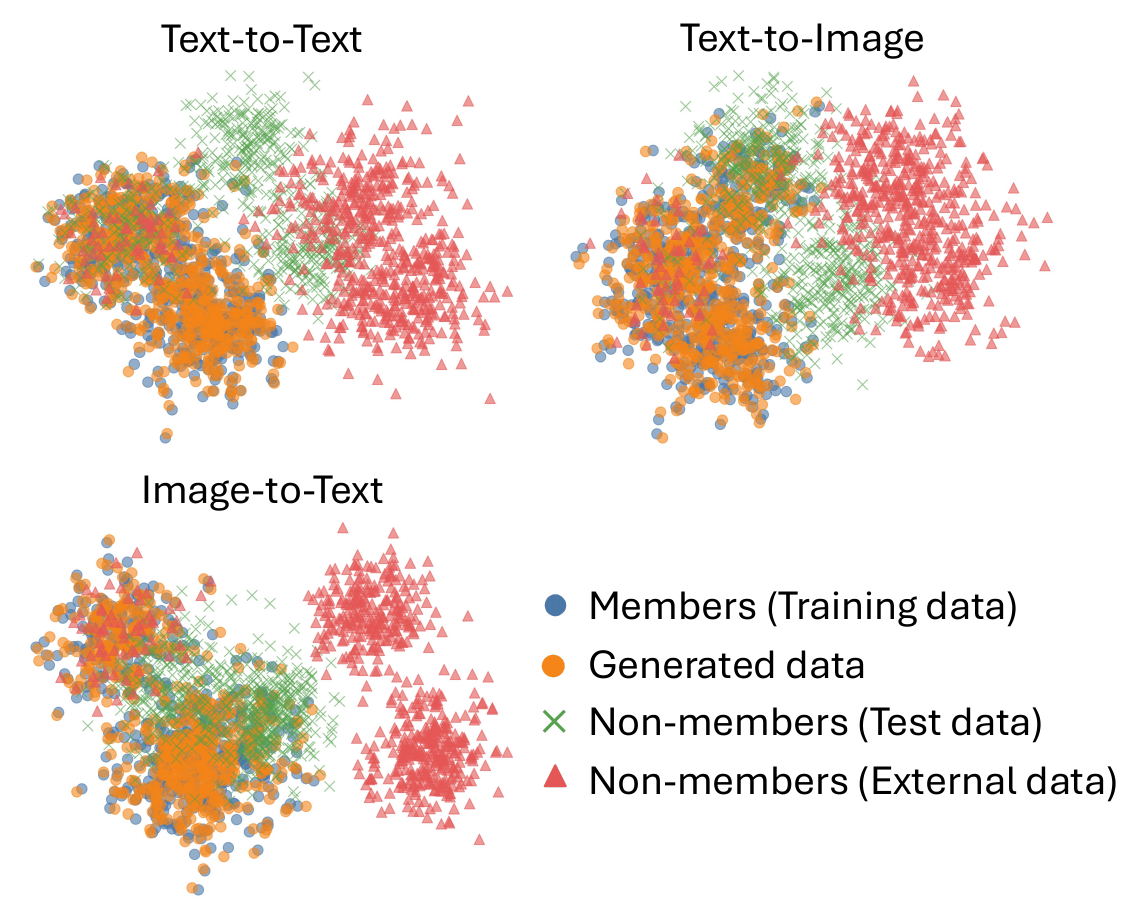}
	\caption{t-SNE Visualization of Data Distribution.}
    \vspace{-0mm}
	\label{fig:tsne}
\end{figure}

To qualitatively illustrate the data distributions, we use t-SNE to visualize the relationships among them \cite{t-SNE}. Figure \ref{fig:tsne} shows the distributions of training data (member data), generated data, test data (non-member data drawn from the same dataset as the training data), and external data (non-member data from the same task domain but from a different dataset) across the three categories of generative models. We observe a consistent trend across all three categories: the distribution of generated data aligns much more closely with that of the member data than with either type of non-member data. Moreover, although the test data are drawn from the same dataset as the training data, their distribution remains farther from the training data distribution than the generated data does. This indicates that generative models capture distributional characteristics that are more strongly influenced by the specific samples seen during training, rather than merely reflecting the overall data domain, supporting our analysis.

\vspace{1mm}
\noindent\textbf{(ii) Likelihood Ratio for Membership Inference.} Let $z^* = E(x^*)$ denote the embedding of a target sample $x^*$. We consider the following binary hypotheses:
\begin{equation}\nonumber
\begin{aligned}
    &\mathbf{H_0}:z^*\sim p_{\mathrm{real}},\\
    &\mathbf{H_1}:z^*\sim p_{\mathrm{syn}},
\end{aligned}
\end{equation}
where $p_{\mathrm{real}}=(\mu_{x_{\mathrm{real}}}, \Sigma_{x_{\mathrm{real}}})$ denotes the embedding distribution of real, non-member samples drawn from the same domain as $D_{\mathrm{tune}}$, and $p_{\mathrm{syn}}=(\mu_{x_{\mathrm{syn}}}, \Sigma_{x_{\mathrm{syn}}})$ denotes the embedding distribution of model-generated (synthetic) samples.

We define the log likelihood ratio as
\begin{equation}\nonumber
\Lambda(z)=\mathrm{log}\frac{p[z|\mu_{x_{\mathrm{syn}}},\Sigma_{x_{\mathrm{syn}}}]}{p[z|\mu_{x_{\mathrm{real}}},\Sigma_{x_{\mathrm{real}}}]}=\mathrm{log}\frac{p_{\mathrm{syn}}(z)}{p_{\mathrm{real}}(z)}.
\end{equation}
The proposed decision rule infers $x^*$ as a member if $\Lambda(z^*) > 0$. Taking expectations of $\Lambda(z)$ under each hypothesis yields 
\begin{equation}\nonumber
\begin{aligned}
    \mathbb{E}_{z\sim p_{\mathrm{real}}}[\Lambda(z)]&=-D_{\mathrm{KL}}(p_{\mathrm{real}}||p_{\mathrm{syn}}),\\
    \mathbb{E}_{z\sim p_{\mathrm{syn}}}[\Lambda(z)]&=D_{\mathrm{KL}}(p_{\mathrm{syn}}||p_{\mathrm{real}}).
\end{aligned}
\end{equation}
Since KL divergence is non-negative and equals zero if and only if the two distributions coincide, we have
\begin{equation}\nonumber
    \mathbb{E}_{z\sim p_{\mathrm{real}}}[\Lambda(z)]\leq 0, \mathbb{E}_{z\sim p_{\mathrm{syn}}}[\Lambda(z)]\geq 0, 
\end{equation}
with strict inequalities whenever $p_{\mathrm{real}} \neq p_{\mathrm{syn}}$. These results imply that the likelihood ratio $\Lambda(z)$ is, in expectation, positive for member samples and negative for non-members. Hence, any nonzero KL divergence between the real and synthetic embedding distributions induces a statistical bias that can be exploited for membership inference. Moreover, the magnitude of the KL divergence directly controls the separability of the two hypotheses: larger divergence implies a larger expectation gap and, thus, a lower achievable classification error.

\vspace{1mm}
\noindent\textbf{(iii) Optimality of the Decision Rule.} Given the two hypotheses $\mathbf{H_0}$ and $\mathbf{H_1}$, consider a test $\delta(z)\in\{0,1\}$, where $\delta(z)=0$ corresponds to deciding $\mathbf{H_0}$ and $\delta(z)=1$ corresponds to deciding $\mathbf{H_1}$. Under equal misclassification costs (0-1 loss), the Bayes risk is defined as 
\begin{equation}\nonumber
    R(\delta)=\pi_0\mathbb{P}_0[\delta(z)=1]+\pi_1\mathbb{P}_1[\delta(z)=0],
\end{equation}
where $\pi_0=\mathbb{P}(\mathbf{H_0})$ and $\pi_1=\mathbb{P}(\mathbf{H_1})$ denote the prior probabilities of the two hypotheses. Here, $\mathbb{P}_0[\delta(z)=1]$ represents the probability of deciding $\mathbf{H_1}$ when $\mathbf{H_0}$ is true (false positive rate), and $\mathbb{P}_1[\delta(z)=0]$ represents the probability of deciding $\mathbf{H_0}$ when $\mathbf{H_1}$ is true (false negative rate).

The Bayes risk can equivalently be written in integral form
\begin{equation}\nonumber
\begin{aligned}
    R(\delta)&=\int\Big(\pi_0\delta(z)p_{\mathrm{real}}(z)+\pi_1(1-\delta(z))p_{\mathrm{syn}}(z)\Big)\mathrm{d}z\\
    &=\int\Big(\pi_1 p_{\mathrm{syn}}(z)+\delta(z)(\pi_0 p_{\mathrm{real}}(z)-\pi_1 p_{\mathrm{syn}}(z))\Big)\mathrm{d}z.
\end{aligned}
\end{equation}
Since the first term, $\int[\pi_1 p_{\mathrm{syn}}(z)]\mathrm{d}z=\pi_1$, is independent of $\delta$, minimizing $R(\delta)$ implies minimizing the integrand pointwise for each $z$. This yields the Bayes-optimal decision rule
\[
\delta^*(z) =
\begin{cases}
1, & \text{if } \pi_0 p_{\mathrm{real}}(z) - \pi_1 p_{\mathrm{syn}}(z) < 0; \\
0, & \text{otherwise}.
\end{cases}
\]
The above rule can be equivalently expressed as 
\begin{equation}\nonumber
\begin{aligned}
\delta^*(z) &= \mathbb{I}\!\left[\pi_1 p_{\mathrm{syn}}(z) > \pi_0 p_{\mathrm{real}}(z)\right]
= \mathbb{I}\!\left[\frac{p_{\mathrm{syn}}(z)}{p_{\mathrm{real}}(z)} > \frac{\pi_0}{\pi_1}\right]\\
&= \mathbb{I}\!\left[\Lambda(z) > \log\!\left(\frac{\pi_0}{\pi_1}\right)\right].
\end{aligned}
\end{equation}
Thus, under equal prior probabilities, $\pi_0=\pi_1$, the threshold simplifies to $\log\!\left(\frac{\pi_0}{\pi_1}\right)=0$, and the Bayes-optimal rule becomes
\[
\delta^*(z) =\mathbb{I}\!\left[\Lambda(z) > 0\right].
\]
That means the optimal decision rule is obtained by thresholding the log-likelihood ratio $\Lambda(z)$ at zero, which is exactly the decision rule adopted in our method.

\begin{table*}[ht!]\footnotesize
	\centering
 	\caption{Overall Results of Our Method and Baselines Across Three Categories of Generative Models in the Partial Knowledge Setting}
    
\setlength{\tabcolsep}{3.5pt}
\renewcommand{\arraystretch}{1.0}
\begin{tabular} {ccccccc|cccccc|cccccc}
\toprule
& \multicolumn{6}{c|}{\textbf{Text-to-Text}}\centering & \multicolumn{6}{c|}{\textbf{Text-to-Image}}\centering & \multicolumn{6}{c}{\textbf{Image-to-Text}}\\\cline{2-19}
& \multicolumn{3}{c}{GPT2+Wiki103} & \multicolumn{3}{c|}{Falcon+XSum} & \multicolumn{3}{c}{SD1.5+MSCOCO} & \multicolumn{3}{c|}{SD2.1+CelebA-D.} & \multicolumn{3}{c}{LLaVA+COCO} & \multicolumn{3}{c}{MiniGPT4+CC\_SBU} \\\cline{2-19}
& \textbf{Ours} & SPV & ICP & \textbf{Ours} & SPV & ICP & \textbf{Ours} & Score & CLiD & \textbf{Ours} & Score & CLiD & \textbf{Ours} & Temp. & MaxRényi & \textbf{Ours} & Temp. & MaxRényi \\
\midrule
ASR $\uparrow$        & $0.90$ & $0.87$ & $0.89$ & $0.98$ & $0.95$ & $0.96$ & $0.86$ & $0.85$ & $0.84$ & $0.96$ & $0.88$ & $0.92$ & $0.91$ & $0.88$ & $0.89$ & $0.85$ & $0.83$ & $0.82$ \\%
AUC $\uparrow$        & $0.93$ & $0.88$ & $0.91$ & $0.99$ & $0.97$ & $0.97$ & $0.95$ & $0.94$ & $0.95$ & $0.99$ & $0.95$ & $0.96$ & $0.97$ & $0.95$ & $0.94$ & $0.94$ & $0.92$ & $0.92$ \\
T@1\%F $\uparrow$ & $0.52$ & $0.48$ & $0.51$ & $0.63$ & $0.61$ & $0.60$ & $0.62$ & $0.56$ & $0.57$ & $0.59$ & $0.44$ & $0.48$ & $0.55$ & $0.53$ & $0.52$ & $0.59$ & $0.55$ & $0.53$ \\
\bottomrule
\end{tabular}
	\label{tab:OverallPartial}
\end{table*}

\begin{table*}[ht!]\footnotesize
	\centering
 	\caption{Overall Results of Our Method and Baselines Across Three Categories of Generative Models in the Zero Knowledge Setting}
    \setlength{\tabcolsep}{3.5pt}
\renewcommand{\arraystretch}{1.0}
\begin{tabular} {ccccccc|cccccc|cccccc}
\toprule
& \multicolumn{6}{c|}{\textbf{Text-to-Text}}\centering & \multicolumn{6}{c|}{\textbf{Text-to-Image}}\centering & \multicolumn{6}{c}{\textbf{Image-to-Text}}\\\cline{2-19}
& \multicolumn{3}{c}{GPT2+Wiki103} & \multicolumn{3}{c|}{Falcon+XSum} & \multicolumn{3}{c}{SD1.5+MSCOCO} & \multicolumn{3}{c|}{SD2.1+CelebA-D.} & \multicolumn{3}{c}{LLaVA+COCO} & \multicolumn{3}{c}{MiniGPT4+CC\_SBU} \\\cline{2-19}
& \textbf{Ours} & SPV & ICP & \textbf{Ours} & SPV & ICP & \textbf{Ours} & Score & CLiD & \textbf{Ours} & Score & CLiD & \textbf{Ours} & Temp. & MaxRényi & \textbf{Ours} & Temp. & MaxRényi \\
\midrule
ASR $\uparrow$        & $0.83$ & $0.70$ & $0.73$ & $0.88$ & $0.85$ & $0.82$ & $0.78$ & $0.50$ & $0.54$ & $0.89$ & $0.52$ & $0.56$ & $0.88$ & $0.60$ & $0.58$ & $0.85$ & $0.58$ & $0.60$ \\%
AUC $\uparrow$        & $0.90$ & $0.79$ & $0.80$ & $0.96$ & $0.95$ & $0.93$ & $0.90$ & $0.46$ & $0.48$ & $0.94$ & $0.52$ & $0.64$ & $0.96$ & $0.58$ & $0.55$ & $0.91$ & $0.63$ & $0.61$ \\
T@1\%F $\uparrow$ & $0.48$ & $0.40$ & $0.39$ & $0.61$ & $0.54$ & $0.55$ & $0.45$ & $0.15$ & $0.18$ & $0.53$ & $0.21$ & $0.20$ & $0.53$ & $0.23$ & $0.19$ & $0.57$ & $0.31$ & $0.25$ \\
\bottomrule
\end{tabular}
	\label{tab:OverallZero}
\end{table*}

\section{Experimental Setup}\label{sec:experimental setup}
\subsection{Models and Datasets}
The choice of models and datasets follows prior work to ensure a fair comparison. Detailed descriptions of these adopted datasets are provided in the Appendix.

\vspace{1mm}
\noindent\textbf{Text-to-Text Models.} Following the setup in \cite{Fu24NeurIPS}, we adopt two open-source LLMs: GPT-2 (1.5B) \cite{GPT2} and Falcon-7B \cite{Falcon}. The fine-tuning datasets include Wikitext-103 \cite{WikiText} and XSum \cite{XSum}. GPT-2 was pre-trained on the WebText corpus, while Falcon-7B was pre-trained on the RefinedWeb dataset. Neither model’s pre-training documentation lists Wikitext-103 or XSum as part of their training data. Thus, it is considered that these benchmark datasets were not included in their pre-training corpora. In addition, we adopt DistilBERT \cite{Sanh19NeurIPS} as the embedding extractor $E$ due to its strong semantic representation capability. 

\vspace{1mm}
\noindent\textbf{Text-to-Image Models.} Following the setup in \cite{Pang25NDSS}, we use Stable Diffusion v1.5 and Stable Diffusion v2.1 \cite{SD1.5} as our text-to-image generative models. These models were pre-trained on large-scale LAION datasets, including LAION-2B and LAION-5B \cite{LAION5B}. To avoid any overlap between the models’ pre-training data and our fine-tuning datasets, we fine-tune them on MS-COCO \cite{MSCoco} and CelebA-Dialog \cite{CelebADialog}. Additionally, we adopt BLIP \cite{Li22ICML} as the embedding extractor $E$ due to its ability to align textual prompts with generated images in a shared embedding space.

\vspace{1mm}
\noindent\textbf{Image-to-Text Models.} Following the setup in \cite{Hu25USENIX}, we employ LLaVA-7B \cite{Liu23NeurIPS} and MiniGPT-4 \cite{Zhu24ICLR} as our image-to-text generative models, and fine-tune them on COCO 2017 \cite{MSCoco} and CC\_SBU\_ALIGN \cite{Zhu24ICLR}. These datasets are not listed in the pre-training data descriptions provided in the official documentation of LLaVA-7B and MiniGPT-4. Therefore, we consider that COCO 2017 and CC\_SBU\_ALIGN were not included in their pre-training corpora. Moreover, we use all-MiniLM-L6-v2 \cite{MiniLM} as the embedding extractor $E$ due to its strong capability in sentence-level similarity measurement.


\subsection{Evaluation Metrics}
Following prior membership inference studies \cite{Fu24NeurIPS, Hu25USENIX, Pang25NDSS, Wang25CCS}, we evaluate attack performance using three metrics: Attack Success Rate (ASR), which measures the overall accuracy of membership inference; Area Under the Receiver Operating Characteristic Curve (AUC); and True Positive Rate (TPR) evaluated at a low False Positive Rate (FPR).

\subsection{Baseline Attacks}
For each category of generative models, we select two state-of-the-art membership inference attacks as baselines. Specifically, for \textbf{text-to-text models}, we adopt \textbf{SPV-MIA} (Self-calibrated Probabilistic Variation–based Membership Inference Attack) \cite{Fu24NeurIPS}, which exploits probabilistic variation induced by LLM memorization as a membership signal, and \textbf{ICP-MIA} (In-Context Probing-based MIA) \cite{Lu26NDSS}, which utilizes the optimization gap signal where member samples exhibit minimal remaining loss-reduction potential, while non-members retain significant potential for further optimization. 

For \textbf{text-to-image models}, we adopt the \textbf{score-based MIA} \cite{Pang25NDSS}, which leverages the diffusion training objective to quantify a model’s memorization of query samples through similarity scores. We also consider \textbf{CLiD-MIA} \cite{Zhai24NeurIPS}, which exploits the phenomenon of conditional overfitting, where the model tends to overfit the conditional distribution of images given their corresponding text prompts, rather than the marginal distribution of images alone.

For \textbf{image-to-text models}, we adopt the \textbf{temperature-based MIA} \cite{Hu25USENIX}, which infers membership by analyzing the sensitivity of generated outputs to temperature variations across multiple samples. We also consider \textbf{MaxRényi-K\% MIA} \cite{Li24NeurIPS}, which exploits the Rényi entropy of the next-token probability distribution over image or text tokens: if the model has seen a sample during training, it is typically more confident in predicting the next token, resulting in lower Rényi entropy.

\section{Experimental Results}
\noindent\textbf{Overall Results.} 
The overall results of our method and the baseline approaches across the three classes of generative models under the partial-knowledge and zero-knowledge settings are reported in Tables~\ref{tab:OverallPartial} and \ref{tab:OverallZero}. We observe that, in the partial-knowledge setting (Table \ref{tab:OverallPartial}), our method achieves performance comparable to the baselines across all three metrics: ASR, AUC, and TPR@1\%FPR. In contrast, in the zero-knowledge setting (Table \ref{tab:OverallZero}), our method consistently outperforms the baselines, with notable improvements in terms of AUC.
This performance gap arises because existing baselines heavily rely on prior knowledge of the training data distribution or model-specific signals, which become unavailable in the zero-knowledge scenario. By contrast, our method leverages a modality-agnostic, distributional perspective that exploits the approximation between the model’s output distribution and its training data distribution, enabling effective membership inference even in the limited knowledge setting.

\begin{figure}[ht]
\centering
	\begin{minipage}{1\textwidth}
    \subfigure[\small{T2T GPT2+Wiki103}]{
    \includegraphics[scale=0.18]{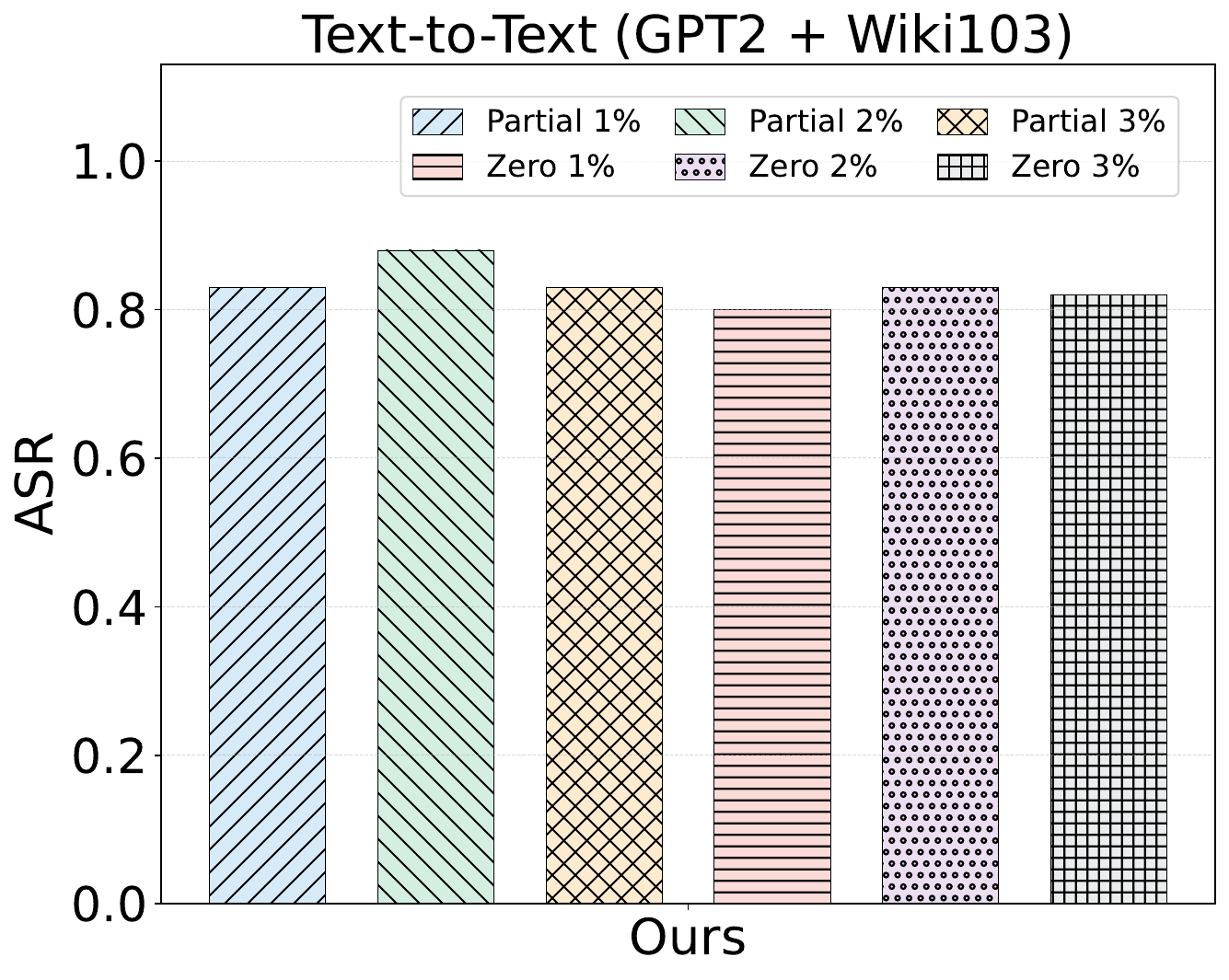}
			\label{fig:DataSizeT2TModel1ASR}}
	\subfigure[\small{T2T Falcon+XSum}]{
    \includegraphics[scale=0.18]{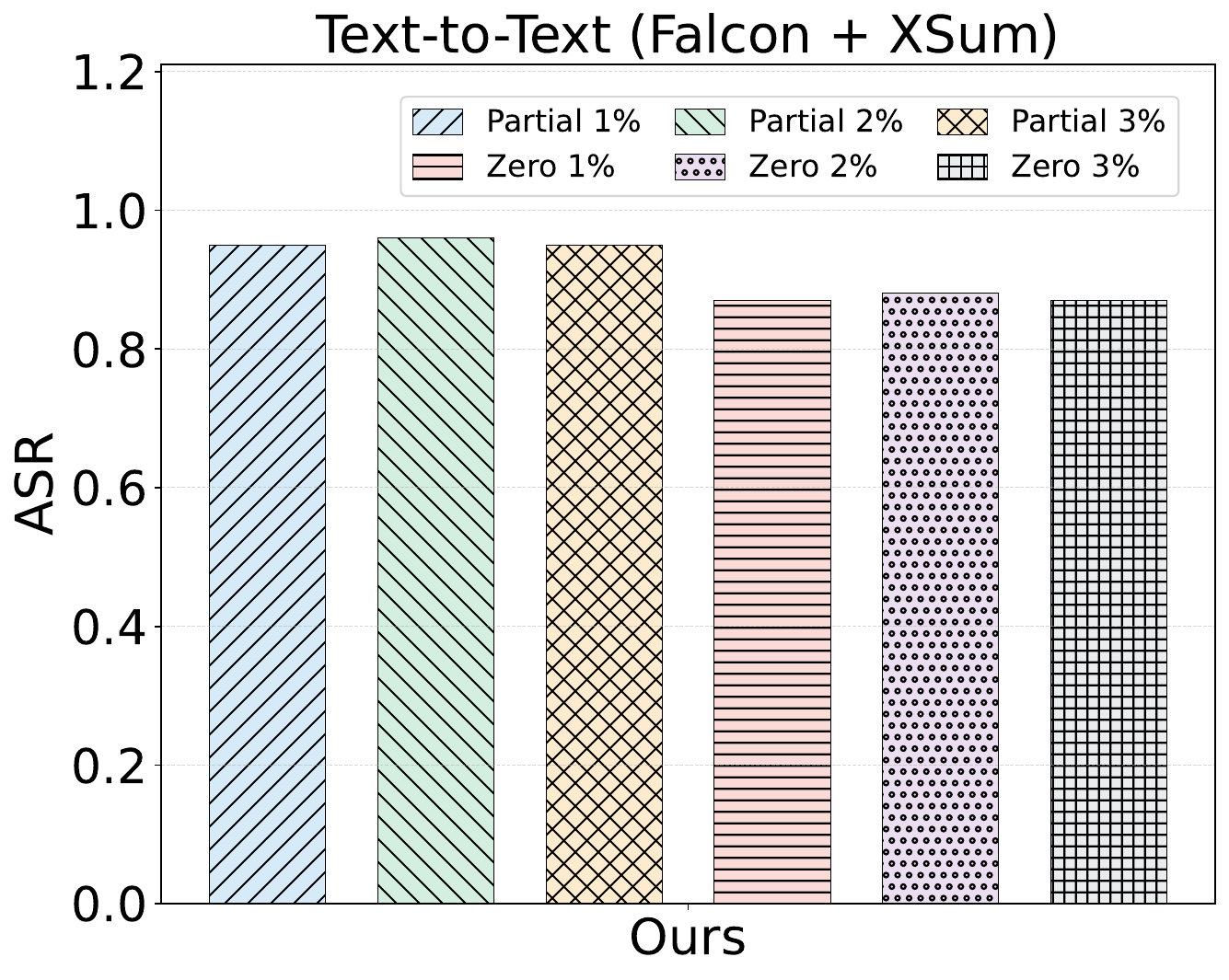}
			\label{fig:DataSizeT2TModel2ASR}}\\
    \subfigure[\small{T2I SD1.5+MSCOCO}]{
    \includegraphics[scale=0.18]{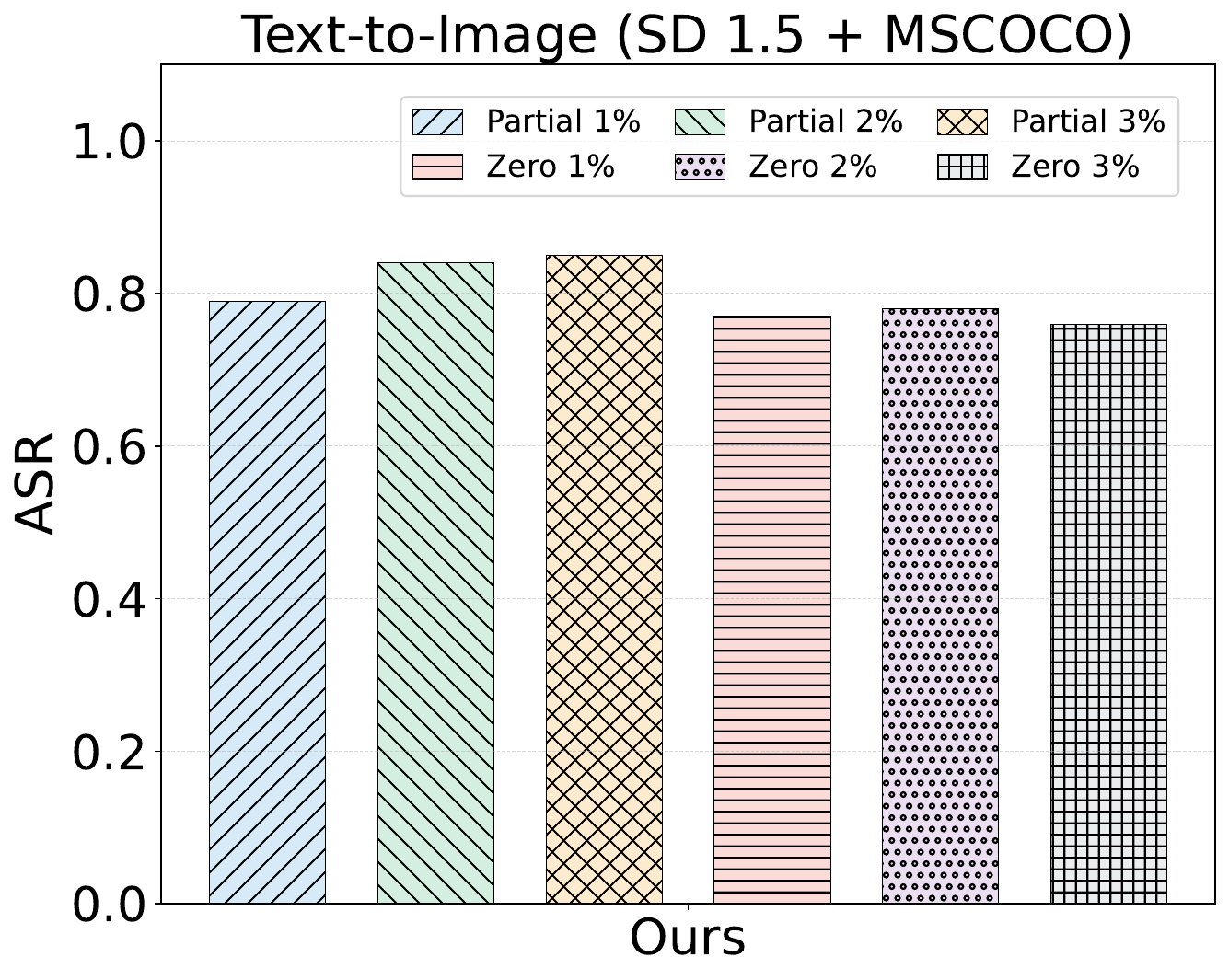}
			\label{fig:DataSizeT2IModel1ASR}}
	\subfigure[\small{T2I SD2.1+CelebA-D.}]{
    \includegraphics[scale=0.18]{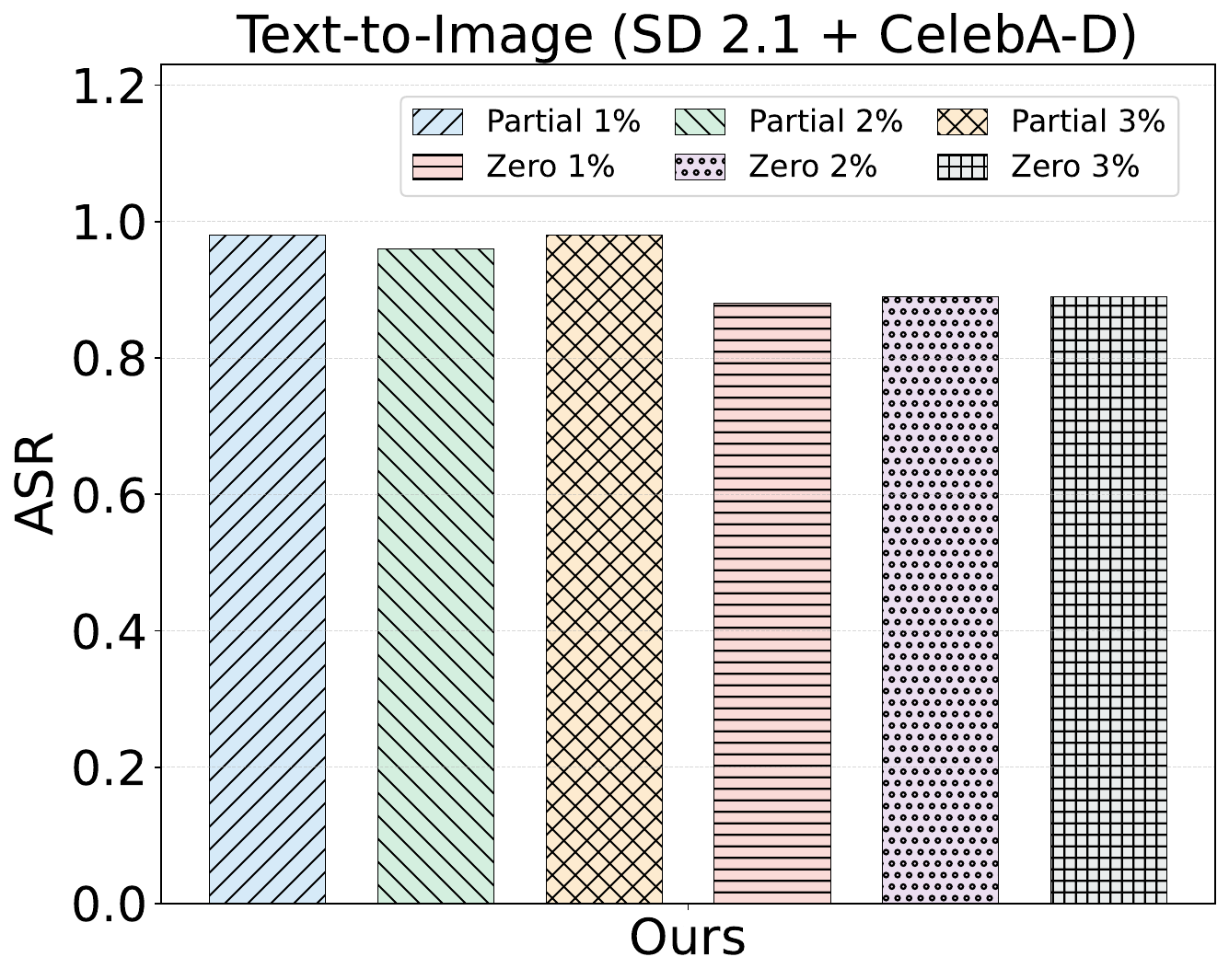}
			\label{fig:DataSizeT2IModel2ASR}}\\
    \subfigure[\small{I2T LLaVa+COCO}]{
    \includegraphics[scale=0.18]{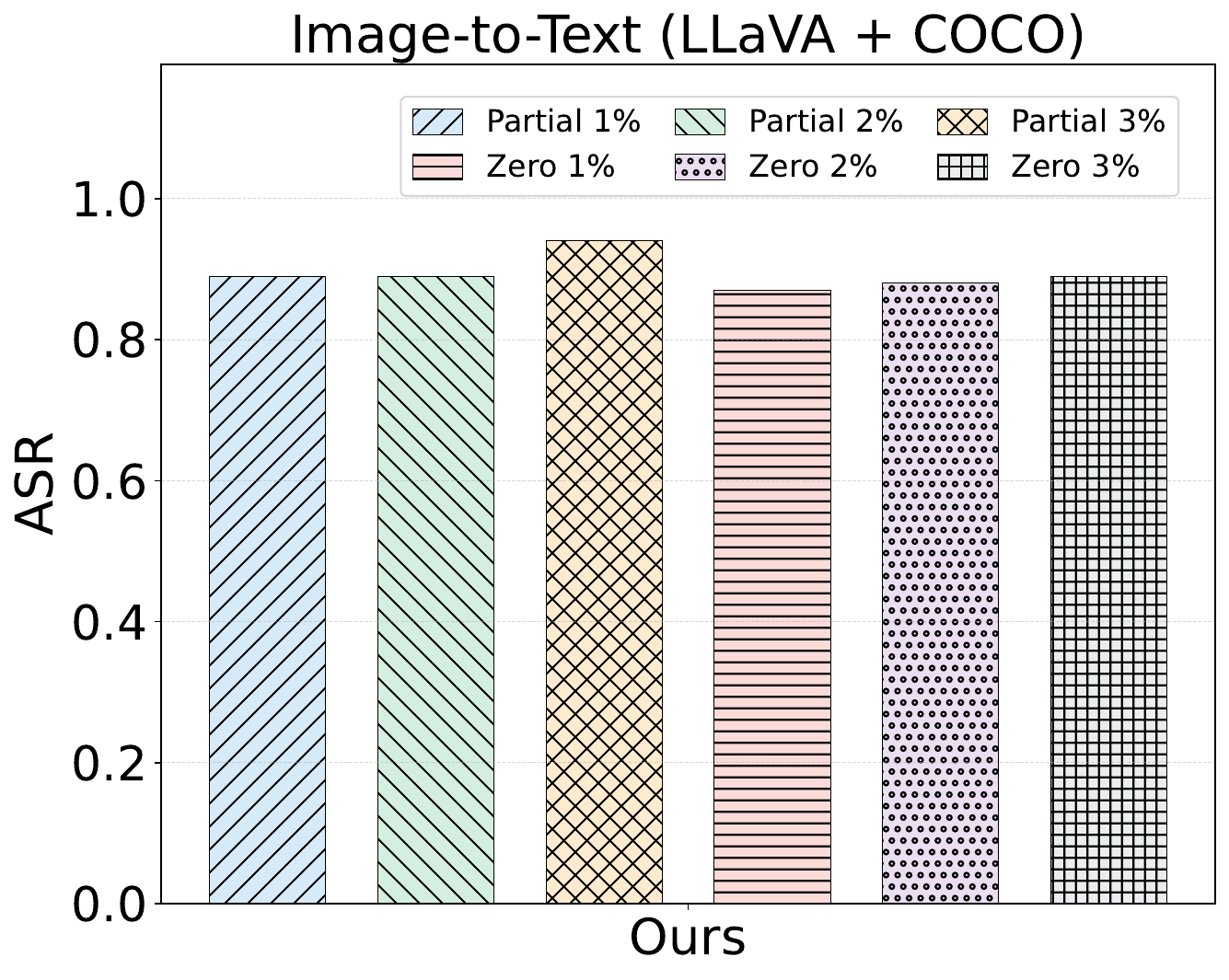}
			\label{fig:DataSizeI2TModel1ASR}}
	\subfigure[\small{I2T MiniGPT-4+CC\_SUB}]{
    \includegraphics[scale=0.18]{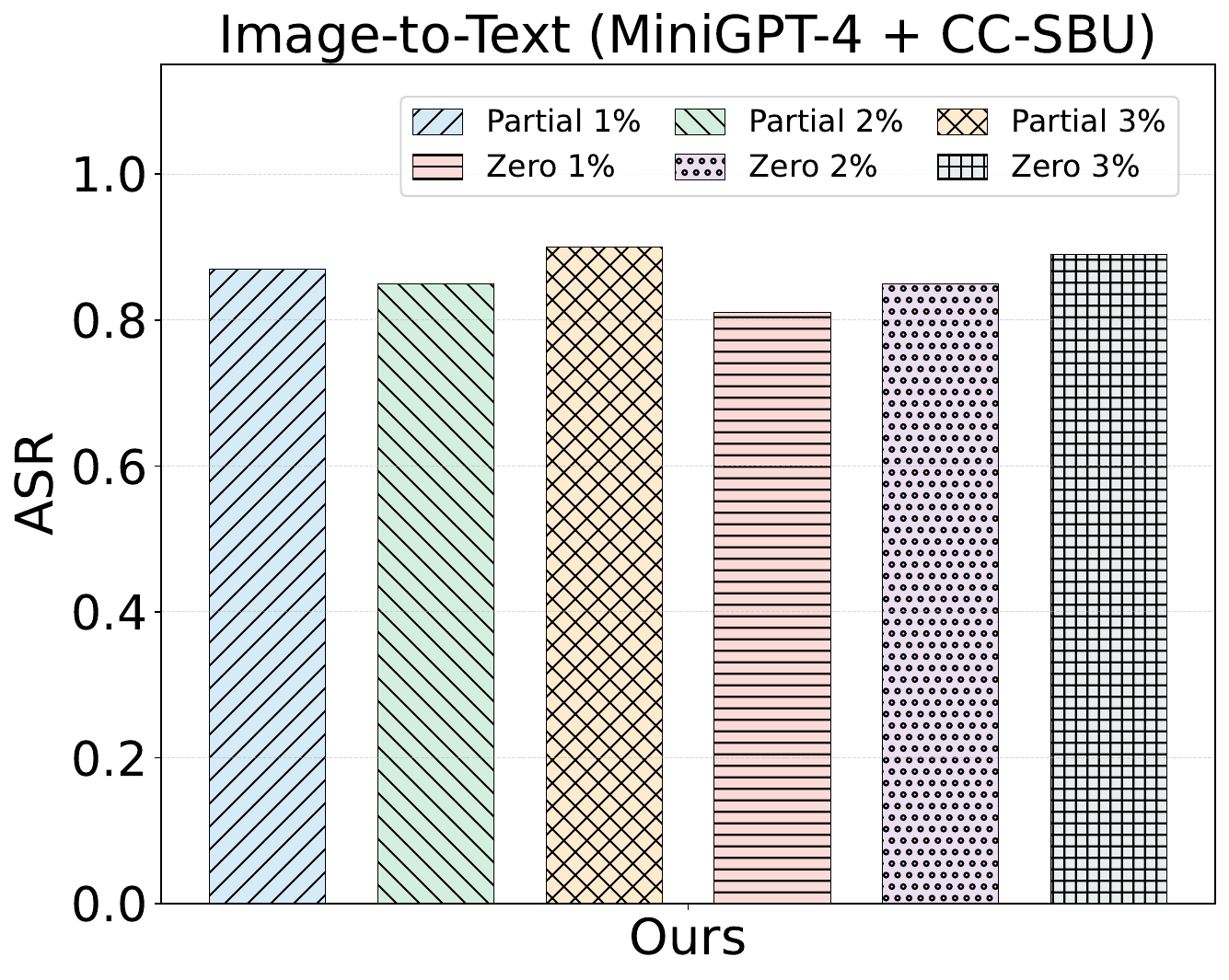}
			\label{fig:DataSizeI2TModel2ASR}}\\
    \end{minipage}
	\caption{ASR Performance of Our Method Under Different Dataset Sizes}
	\label{fig:DataSizeASR}
\end{figure}

\begin{figure}[ht!]
\centering
	\begin{minipage}{1\textwidth}
    \subfigure[\small{T2T GPT2+Wiki103}]{
    \includegraphics[scale=0.18]{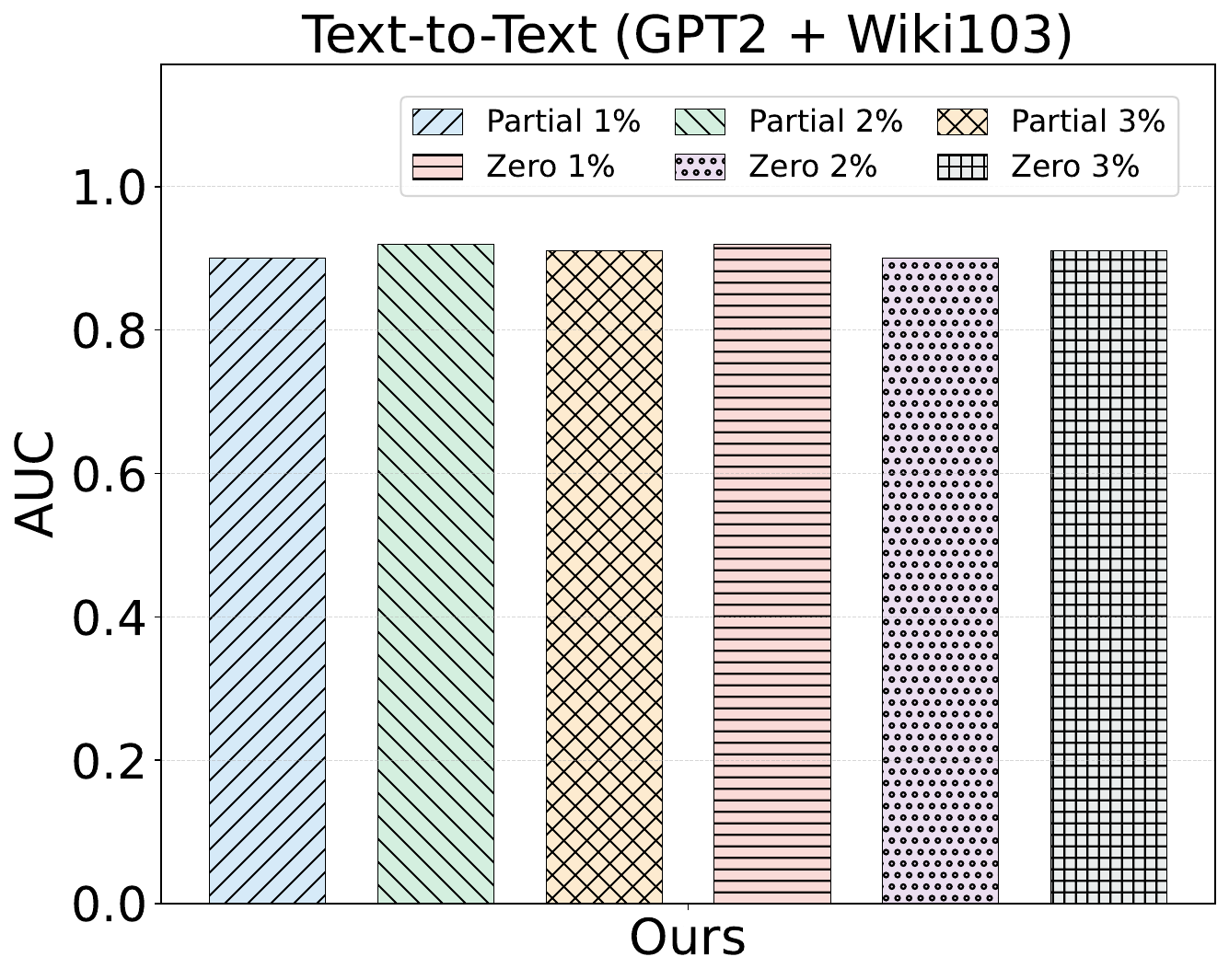}
			\label{fig:DataSizeT2TModel1AUC}}
	\subfigure[\small{T2T Falcon+XSum}]{
    \includegraphics[scale=0.18]{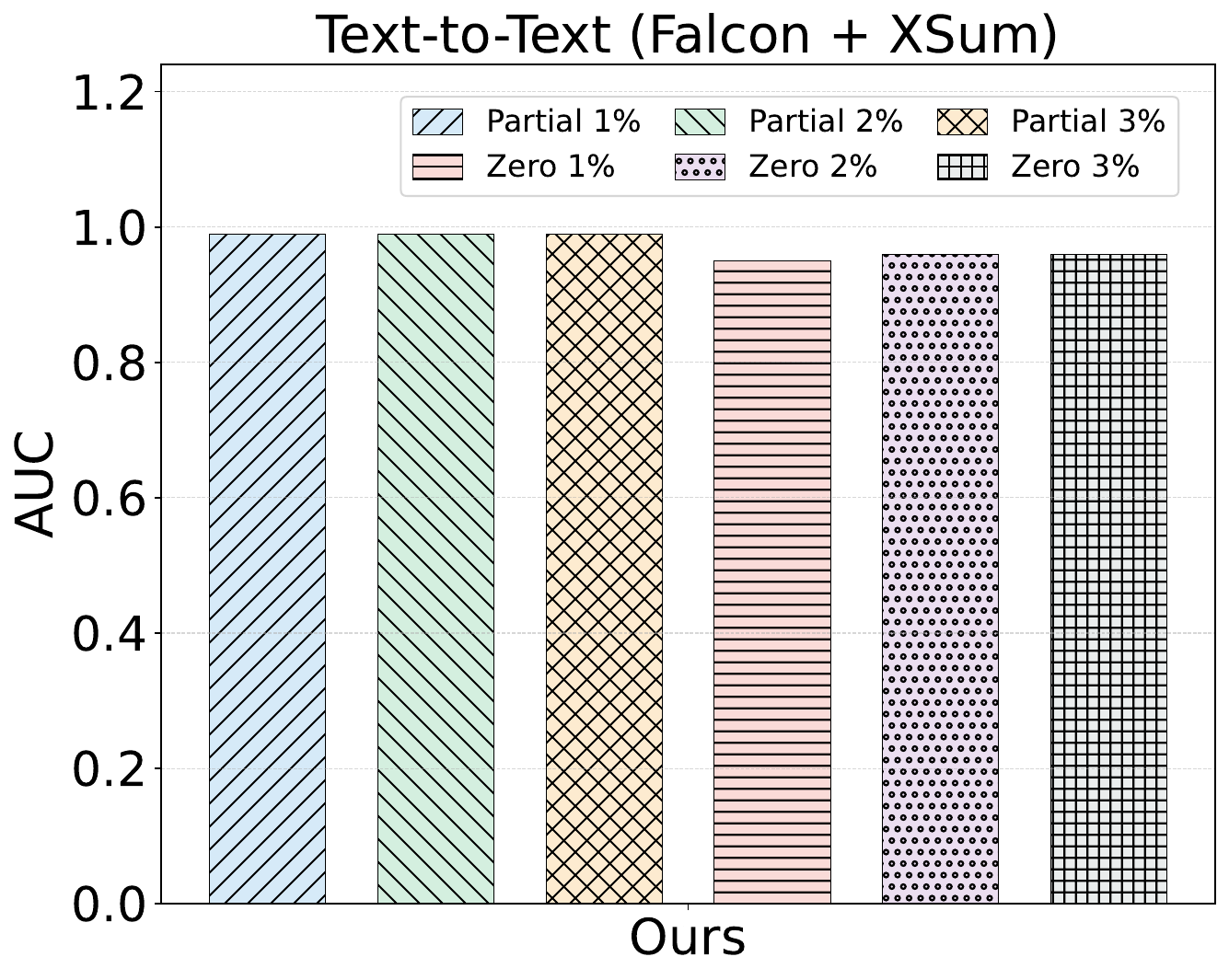}
			\label{fig:DataSizeT2TModel2AUC}}\\
    \subfigure[\small{T2I SD1.5+MSCOCO}]{
    \includegraphics[scale=0.18]{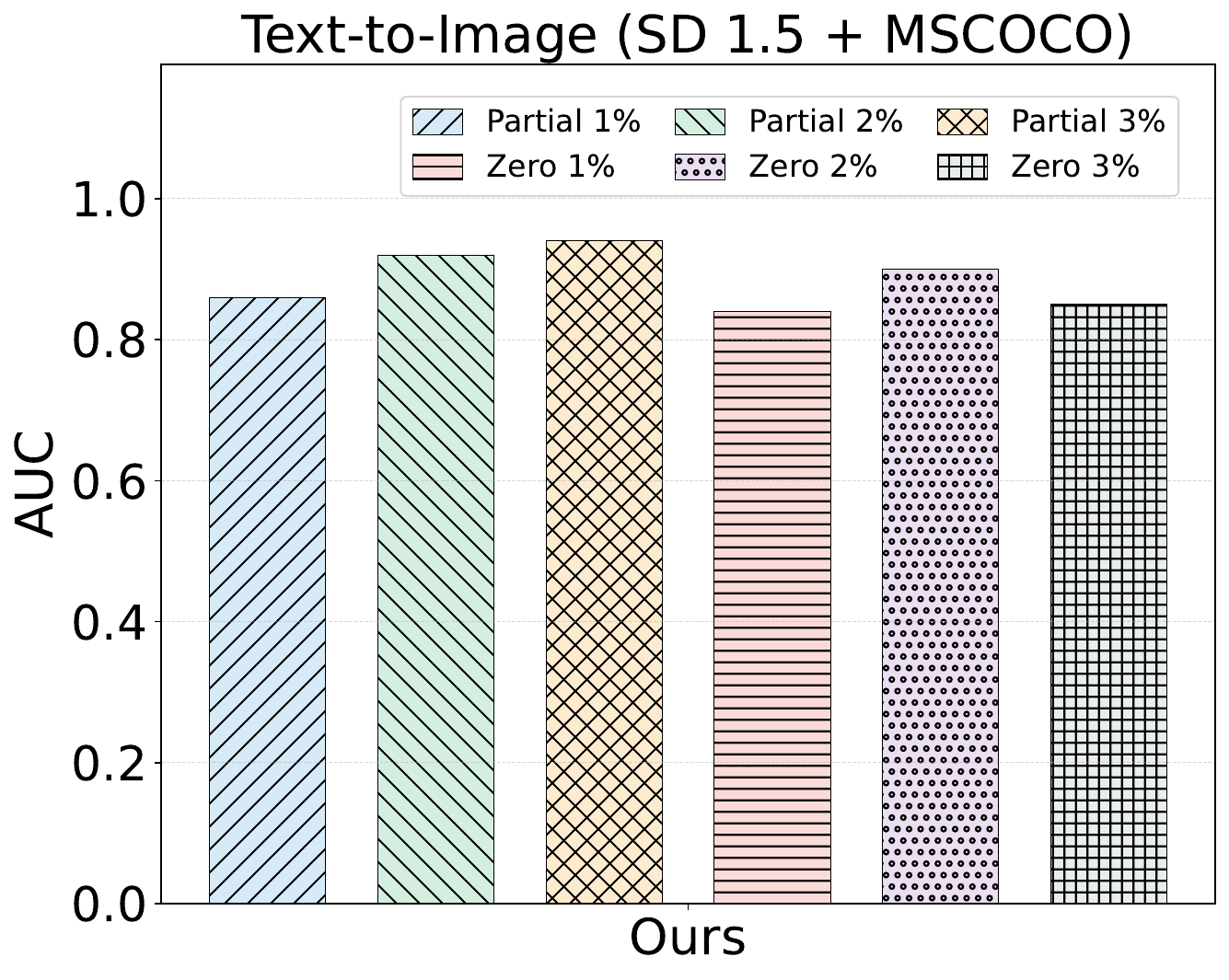}
			\label{fig:DataSizeT2IModel1AUC}}
	\subfigure[\small{T2I SD2.1+CelebA-D.}]{
    \includegraphics[scale=0.18]{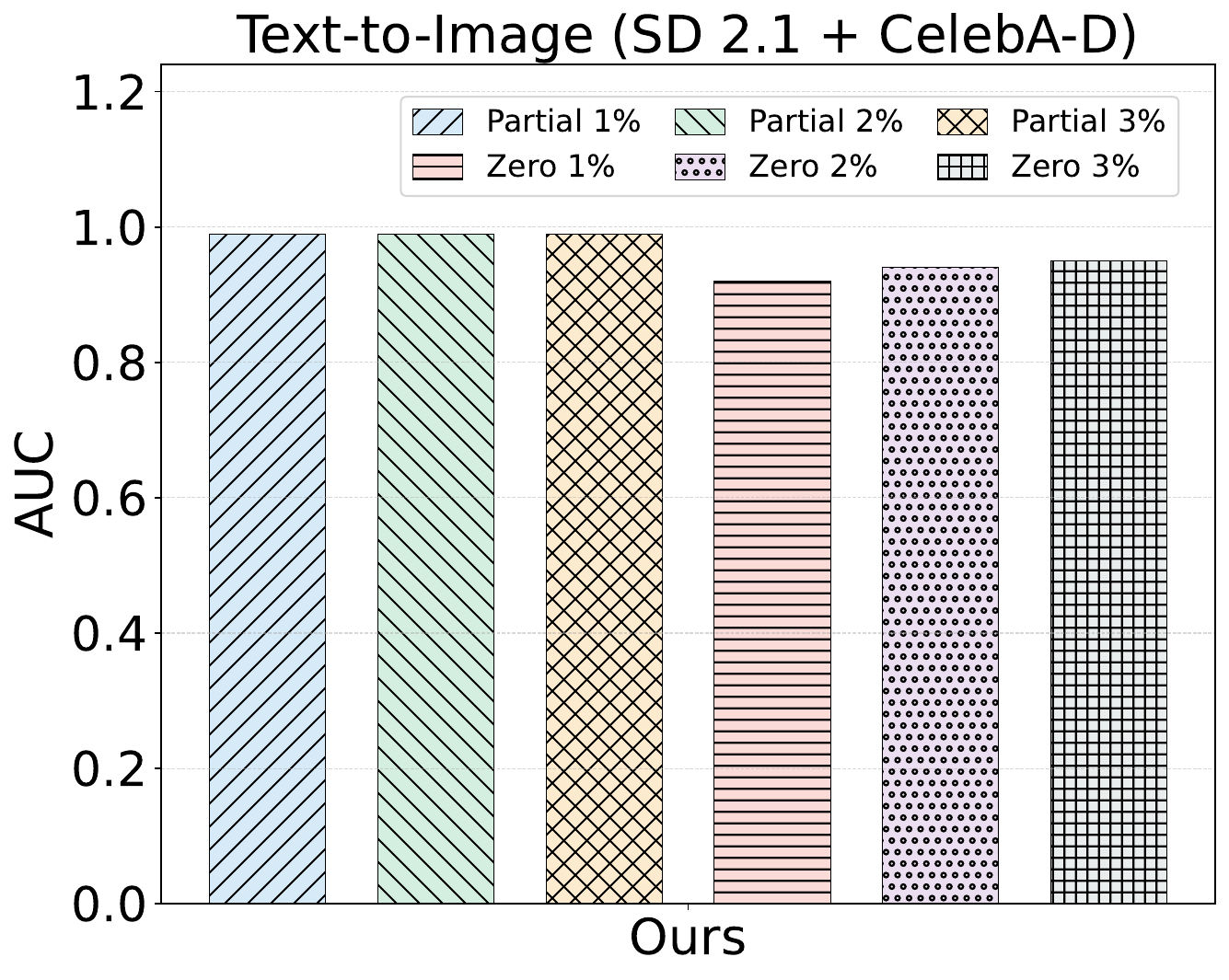}
			\label{fig:DataSizeT2IModel2AUC}}\\
    \subfigure[\small{I2T LLaVa+COCO}]{
    \includegraphics[scale=0.18]{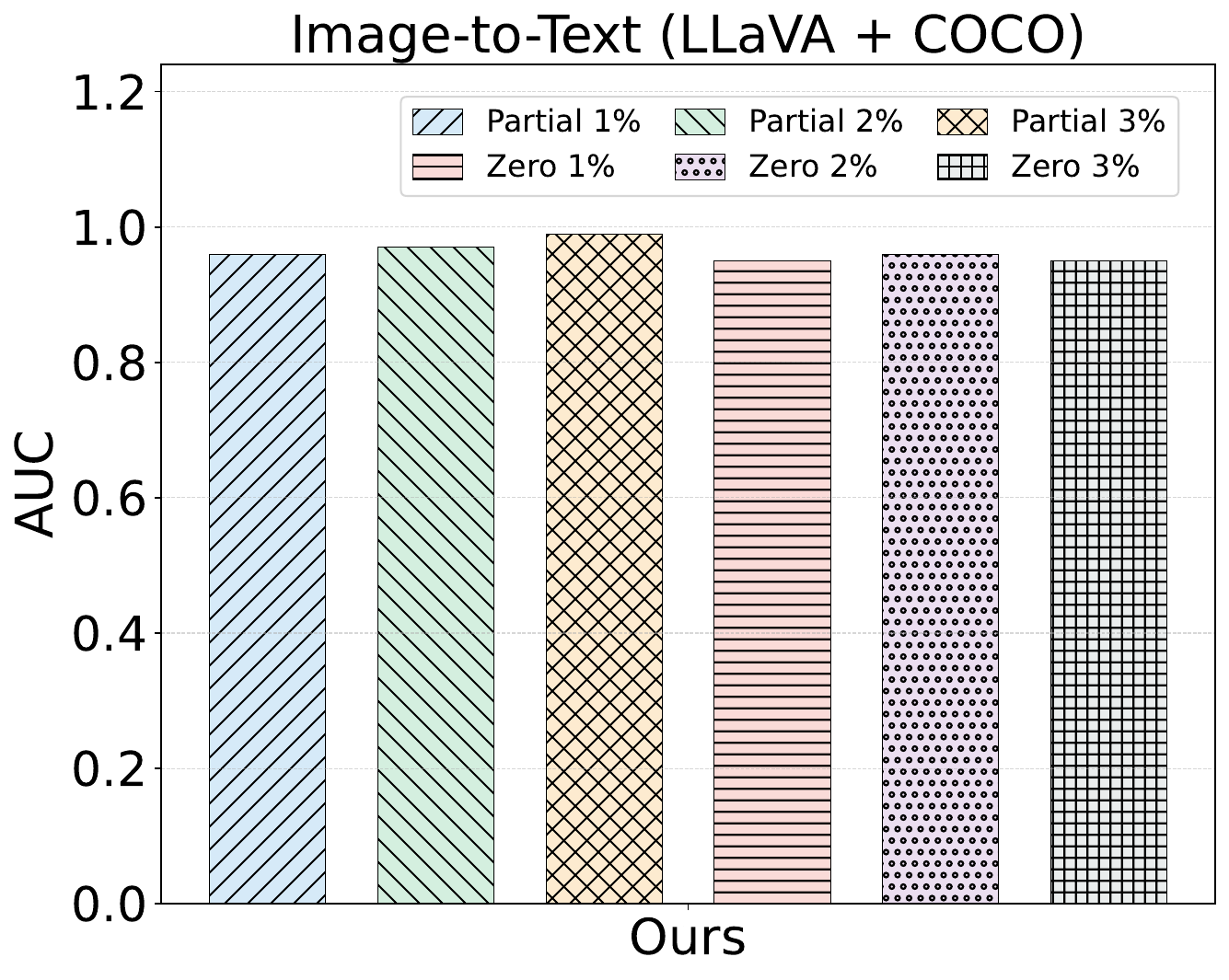}
			\label{fig:DataSizeI2TModel1AUC}}
	\subfigure[\small{I2T MiniGPT-4+CC\_SUB}]{
    \includegraphics[scale=0.18]{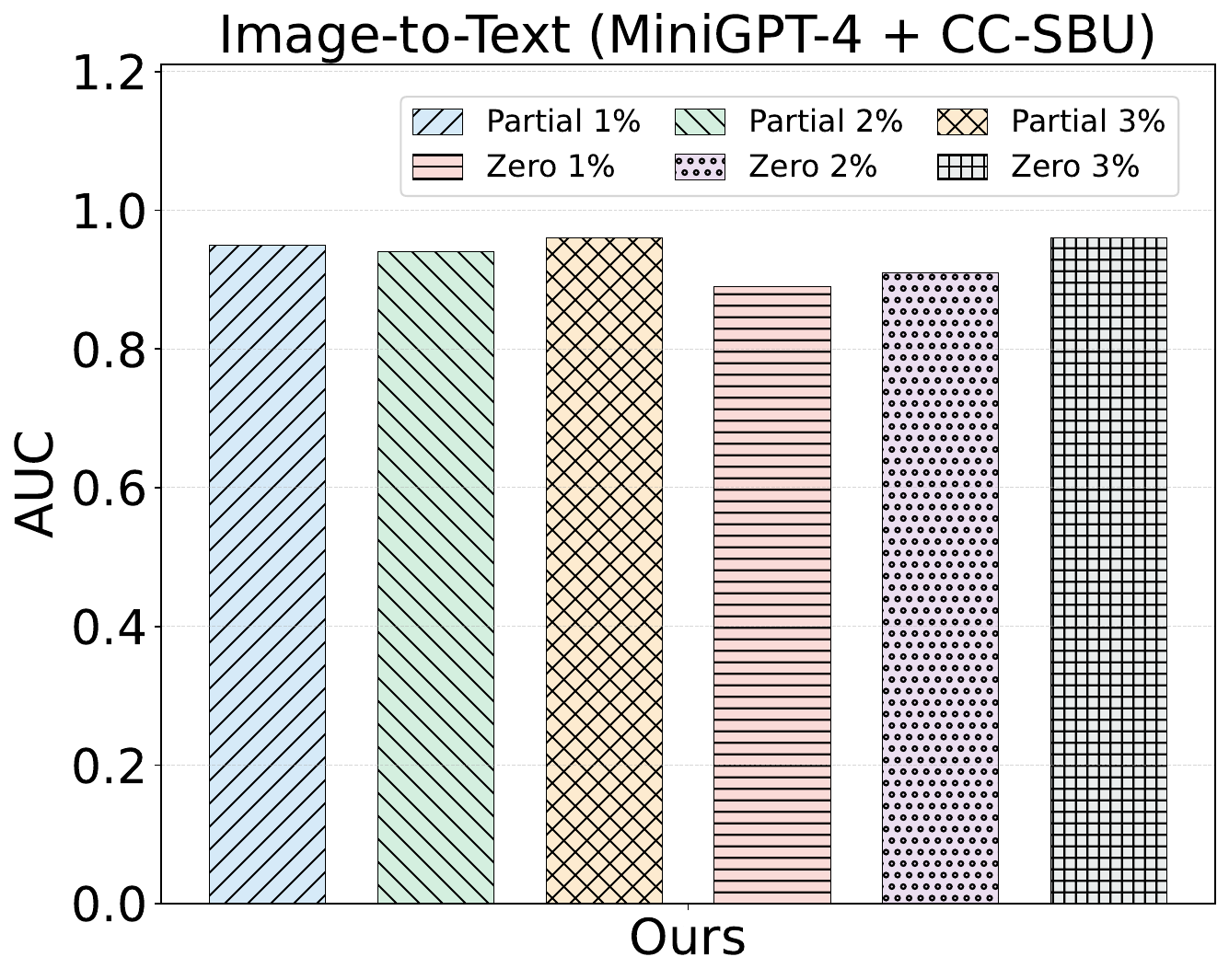}
			\label{fig:DataSizeI2TModel2AUC}}\\
    \end{minipage}
	\caption{AUC Performance of Our Method Under Different Dataset Sizes}
	\label{fig:DataSizeAUC}
\end{figure}

\begin{figure}[ht!]
\centering
	\begin{minipage}{1\textwidth}
    \subfigure[\small{T2T GPT2+Wiki103}]{
    \includegraphics[scale=0.18]{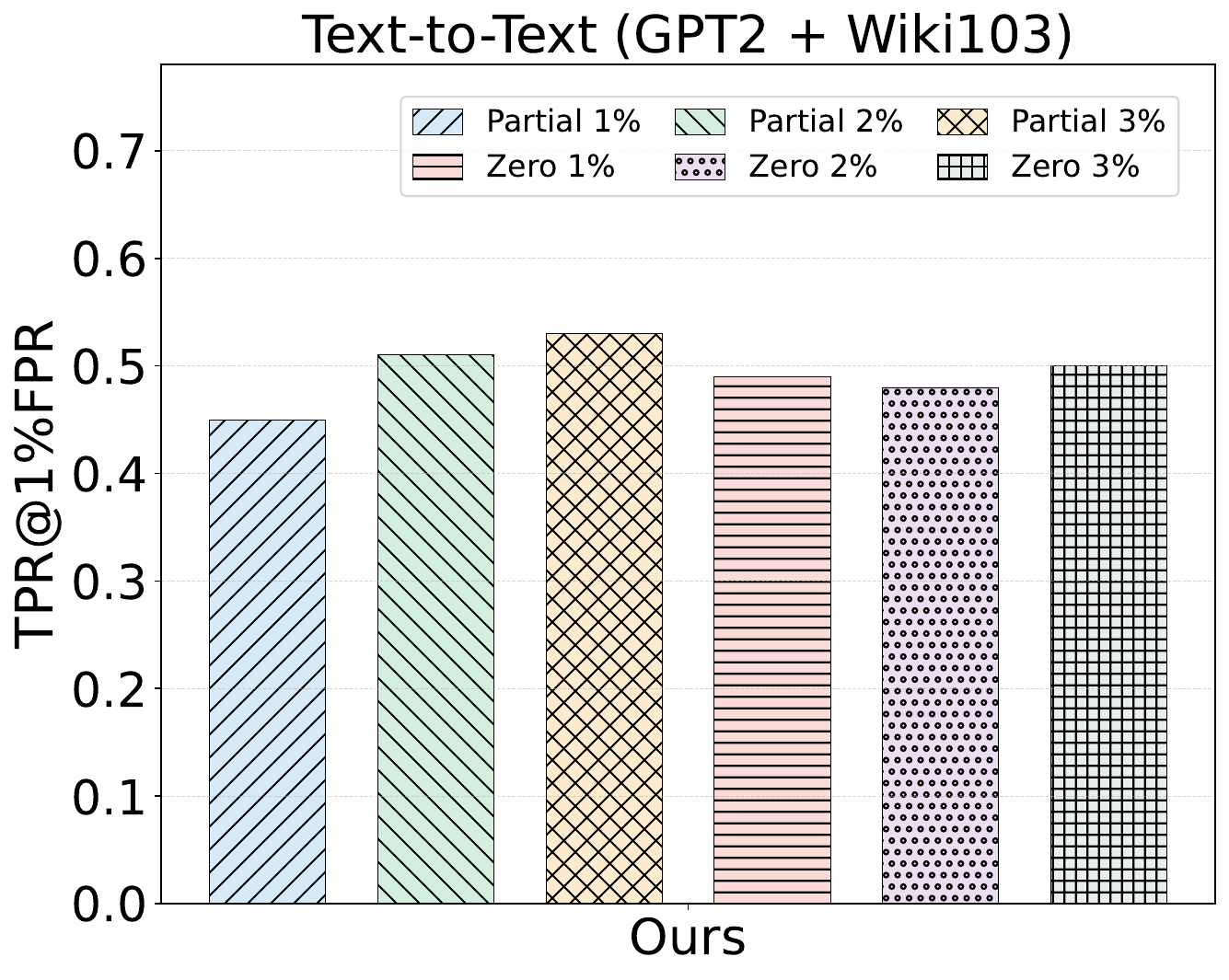}
			\label{fig:DataSizeT2TModel1TPR}}
	\subfigure[\small{T2T Falcon+XSum}]{
    \includegraphics[scale=0.18]{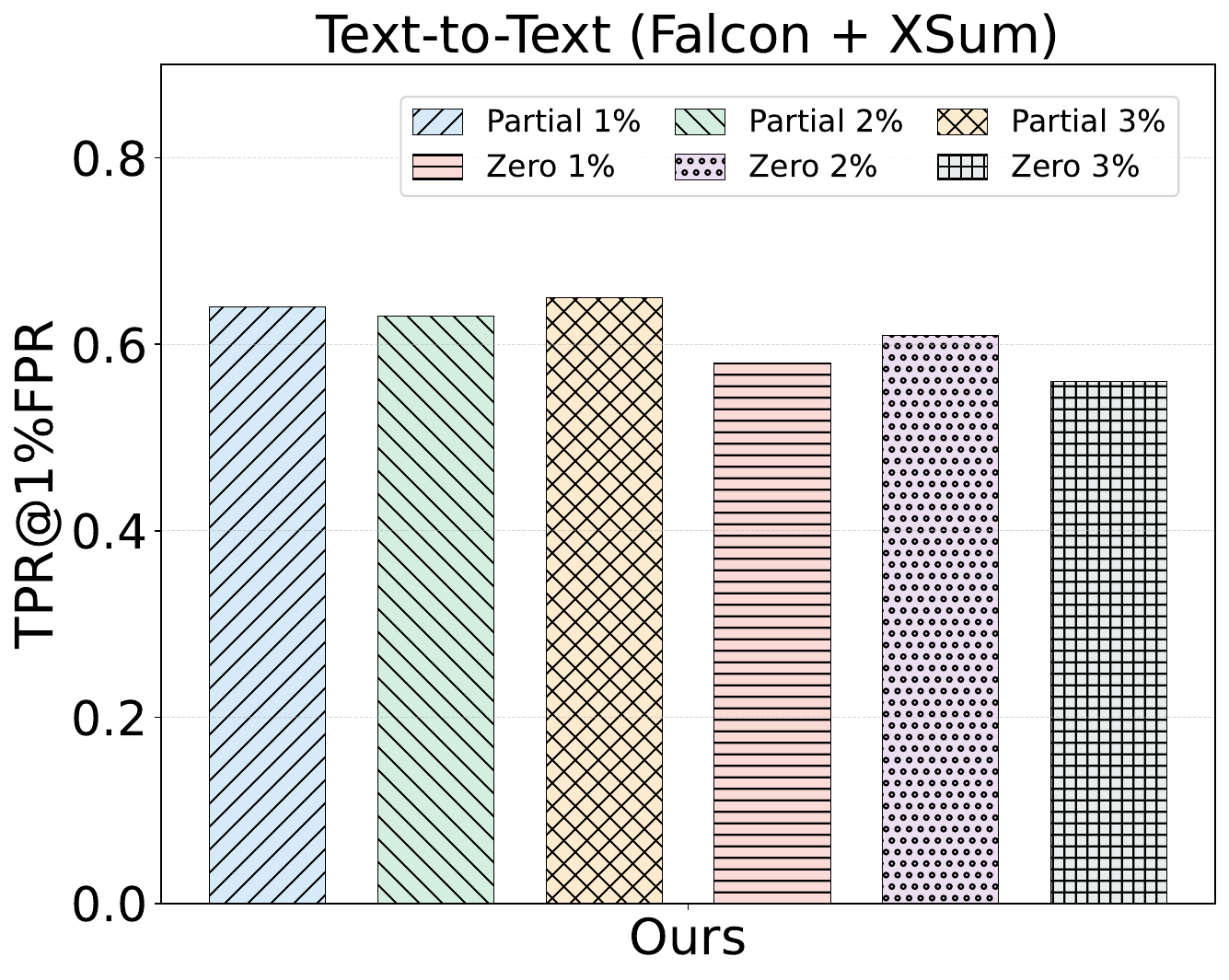}
			\label{fig:DataSizeT2TModel2TPR}}\\
    \subfigure[\small{T2I SD1.5+MSCOCO}]{
    \includegraphics[scale=0.18]{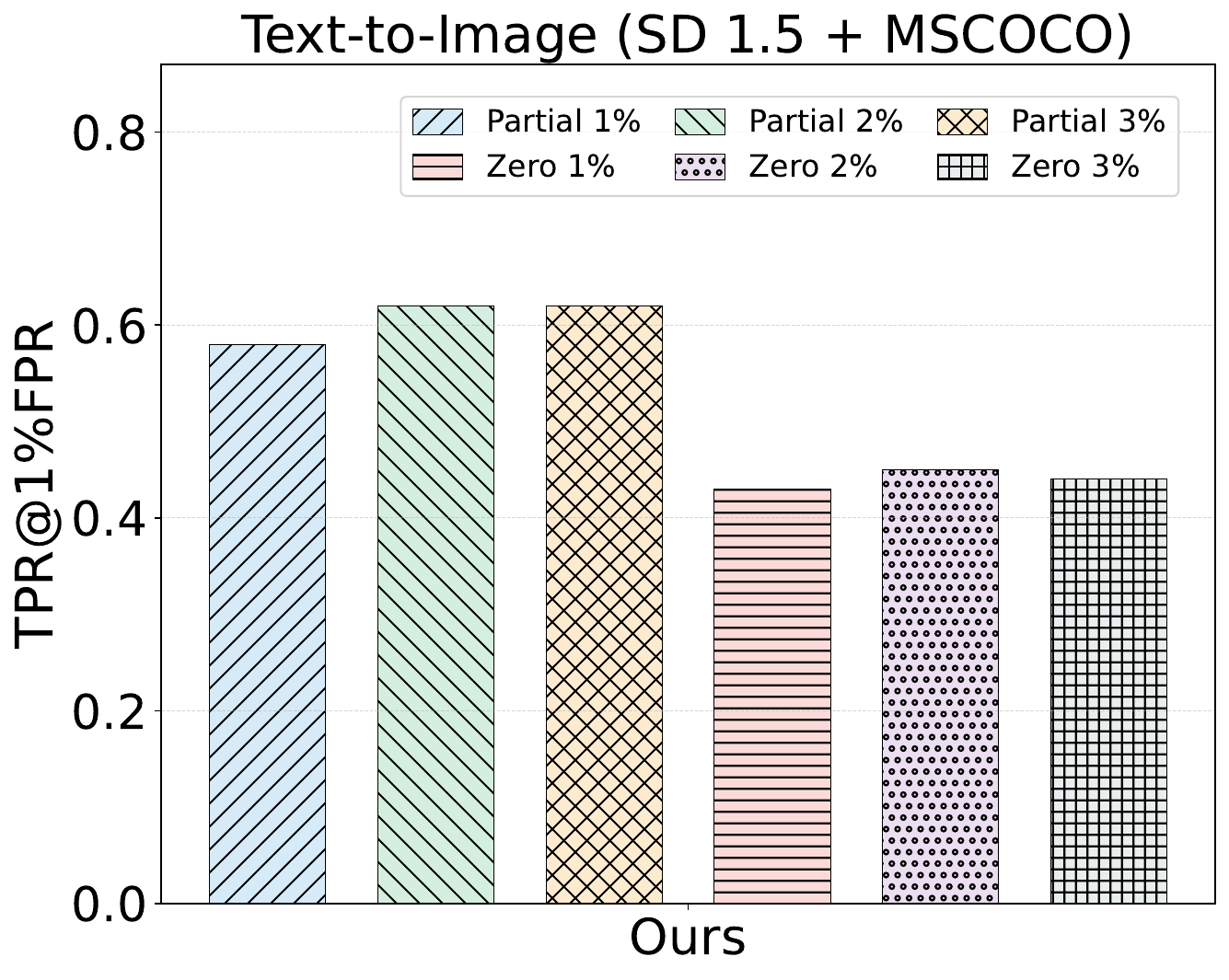}
			\label{fig:DataSizeT2IModel1TPR}}
	\subfigure[\small{T2I SD2.1+CelebA-D.}]{
    \includegraphics[scale=0.18]{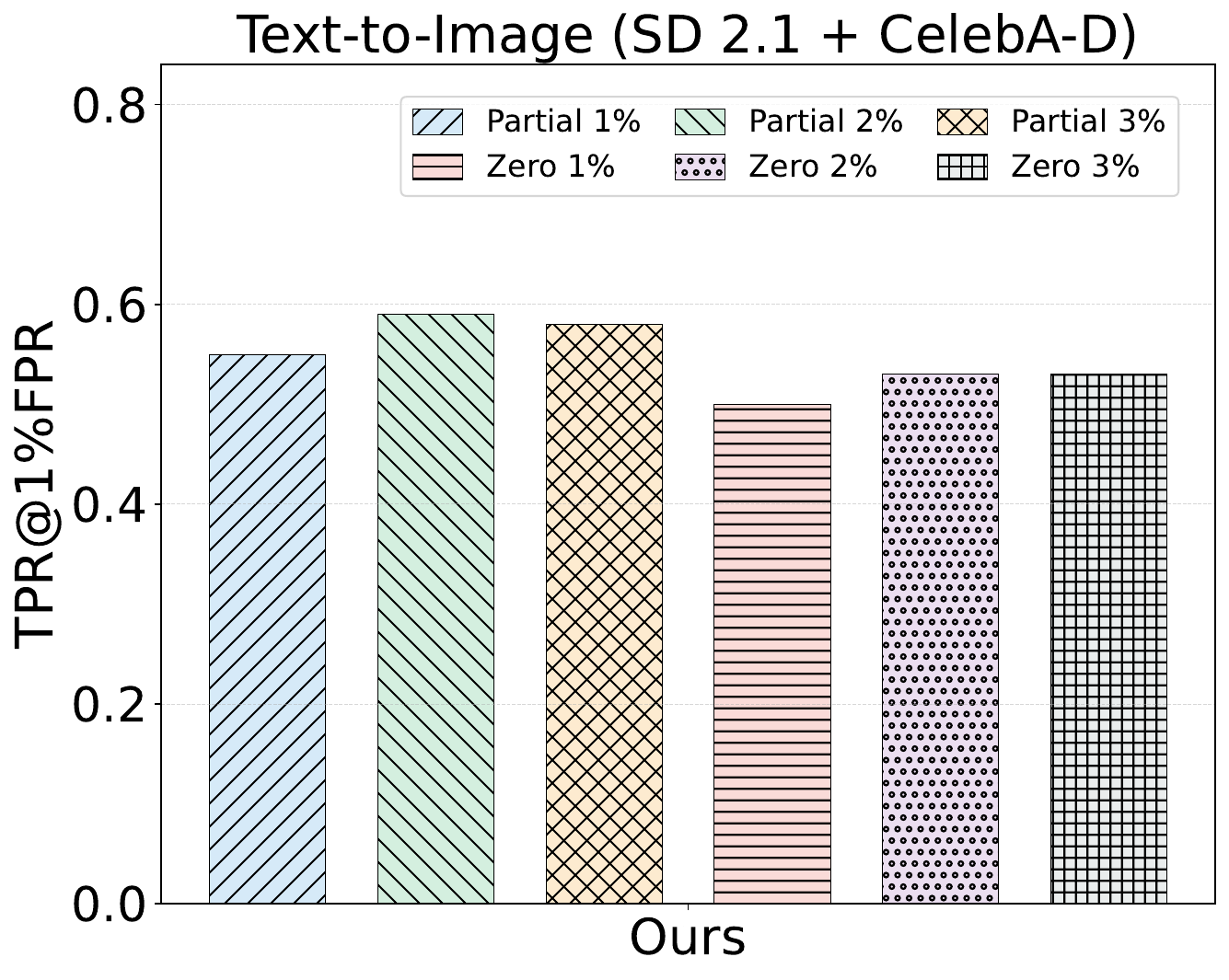}
			\label{fig:DataSizeT2IModel2TPR}}\\
    \subfigure[\small{I2T LLaVa+COCO}]{
    \includegraphics[scale=0.18]{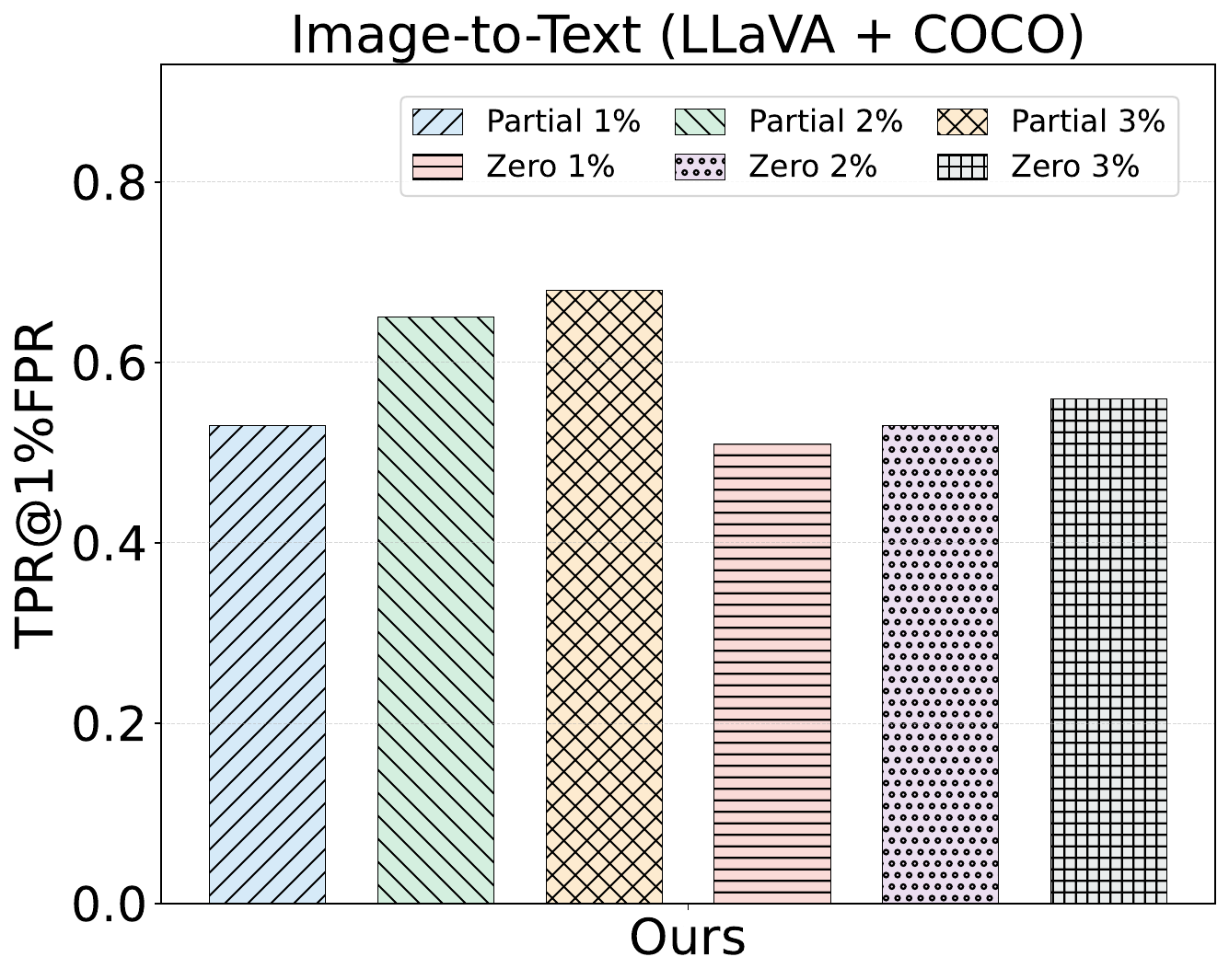}
			\label{fig:DataSizeI2TModel1TPR}}
	\subfigure[\small{I2T MiniGPT-4+CC\_SUB}]{
    \includegraphics[scale=0.18]{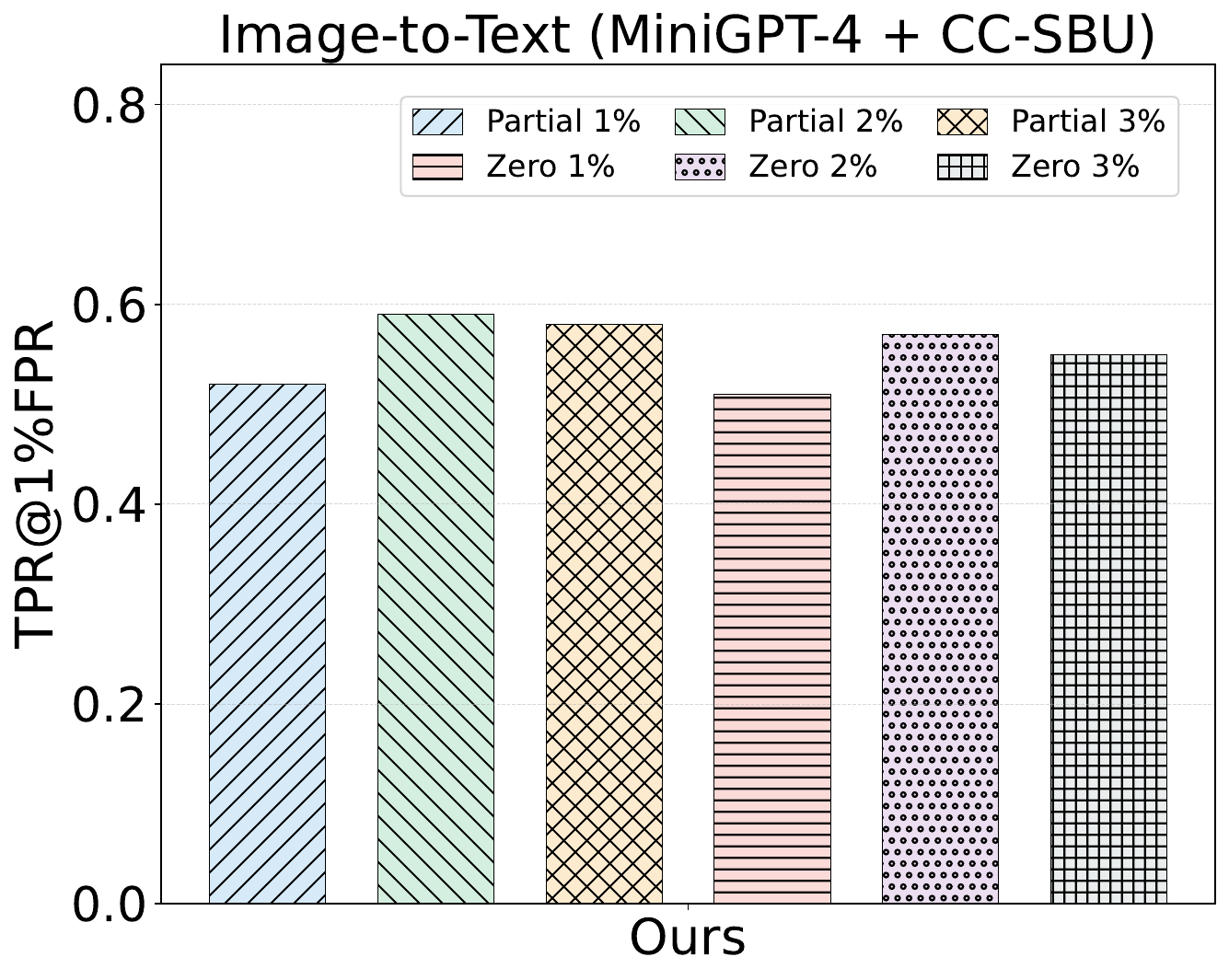}
			\label{fig:DataSizeI2TModel2TPR}}\\
    \end{minipage}
	\caption{TPR@1\%FPR of Our Method Under Different Dataset Sizes}
	\label{fig:DataSizeTPR}
\end{figure}

\vspace{1mm}
\noindent\textbf{Impact of Different Dataset Sizes.} 
Step 1 of our method focuses on data collection and synthesis. We investigate whether the amount of collected data has a significant impact on the performance of our method. 
Specifically, we vary the sizes of both $D_{\mathrm{syn}}$ and $D_{\mathrm{real}}$ ($D_{\mathrm{aux}}$) to $1\%$, $2\%$, and $3\%$ of the size of $D_{\mathrm{tune}}$. Figure \ref{fig:DataSizeASR}, \ref{fig:DataSizeAUC}, and \ref{fig:DataSizeTPR}, respectively, report the ASR, AUC, and TPR@1\%FPR results across the three types of generative models. 
We observe that the amount of data has a limited impact on the performance of our method under both the partial-knowledge and zero-knowledge settings, demonstrating strong scalability and robustness to data availability. This behavior is expected, as our method relies on distributional estimation in the embedding space, which stabilizes quickly even with a relatively small number of samples. 

\begin{table*}[ht!]\scriptsize
	\centering
 	\caption{Results of Our Method on Text-to-Text Models with Different Embedding Extractors}
\begin{tabular} {cccc|ccc|ccc|ccc}
\toprule
& \multicolumn{6}{c|}{Partial-Knowledge Setting} & \multicolumn{6}{c}{Zero-Knowledge Setting} \\\cline{2-13}
& \multicolumn{3}{c|}{GPT2+Wiki103}\centering & \multicolumn{3}{c|}{Falcon+XSum} & \multicolumn{3}{c|}{GPT2+Wiki103}\centering & \multicolumn{3}{c}{Falcon+XSum}\\\cline{2-13}
&  \textbf{Distil} & RoBERTa & MiniLM & \textbf{Distil} & RoBERTa & MiniLM &  \textbf{Distil} & RoBERTa & MiniLM & \textbf{Distil} & RoBERTa & MiniLM\\
\midrule
ASR $\uparrow$       & $0.88$ & $0.89$ & $0.87$ & $0.96$ & $0.96$ & $0.85$ & $0.83$ & $0.71$ & $0.73$ & $0.88$ & $0.68$ & $0.70$ \\%
AUC $\uparrow$       & $0.92$ & $0.94$ & $0.94$ & $0.99$ & $0.99$ & $0.92$ & $0.90$ & $0.91$ & $0.92$ & $0.96$ & $0.91$ & $0.92$ \\
TPR@1\%FPR $\uparrow$     & $0.51$ & $0.50$ & $0.44$ & $0.63$ & $0.61$ & $0.39$ & $0.48$ & $0.40$ & $0.42$ & $0.51$ & $0.32$ & $0.37$ \\
\bottomrule
\end{tabular}
	\label{tab:T2TExtractor}
\end{table*}

\begin{table*}[ht!]\scriptsize
	\centering
 	\caption{Results of Our Method on Text-to-Image Models with Different Embedding Extractors}
\begin{tabular} {cccc|ccc|ccc|ccc}
\toprule
& \multicolumn{6}{c|}{Partial-Knowledge Setting} & \multicolumn{6}{c}{Zero-Knowledge Setting} \\\cline{2-13}
& \multicolumn{3}{c|}{SD1.5+MSCOCO}\centering & \multicolumn{3}{c|}{SD2.1+CelebA-Dialog} & \multicolumn{3}{c|}{SD1.5+MSCOCO}\centering & \multicolumn{3}{c}{SD2.1+CelebA-Dialog}\\\cline{2-13}
& \textbf{BLIP} & ViT-Base & CLIP-ViT & \textbf{BLIP} & ViT-Base & CLIP-ViT & \textbf{BLIP} & ViT-Base & CLIP-ViT & \textbf{BLIP} & ViT-Base & CLIP-ViT  \\
\midrule
ASR $\uparrow$        & $0.84$ & $0.86$ & $0.73$ & $0.96$ & $0.98$ & $0.98$ & $0.78$ & $0.79$ & $0.66$ & $0.89$ & $0.70$ & $0.43$ \\%
AUC $\uparrow$        & $0.92$ & $0.93$ & $0.84$ & $0.99$ & $0.99$ & $0.99$ & $0.90$ & $0.87$ & $0.72$ & $0.94$ & $0.82$ & $0.32$ \\
TPR@1\%FPR $\uparrow$     & $0.62$ & $0.57$ & $0.23$ & $0.59$ & $0.54$ & $0.57$ & $0.45$ & $0.18$ & $0.19$ & $0.53$ & $0.13$ & $0.10$ \\
\bottomrule
\end{tabular}
	\label{tab:T2IExtractor}
\end{table*}

\begin{table*}[ht!]\scriptsize
	\centering
 	\caption{Results of Our Method on Image-to-Text Models with Different Embedding Extractors}
\begin{tabular} {cccc|ccc|ccc|ccc}
\toprule
& \multicolumn{6}{c|}{Partial-Knowledge Setting} & \multicolumn{6}{c}{Zero-Knowledge Setting} \\\cline{2-13}
& \multicolumn{3}{c|}{LLaVA+COCO}\centering & \multicolumn{3}{c|}{MiniGPT4+CC\_SBU} & \multicolumn{3}{c|}{LLaVA+COCO}\centering & \multicolumn{3}{c}{MiniGPT4+CC\_SBU}\\\cline{2-13}
& \textbf{MiniLM} & ALBERT & MPNet & \textbf{MiniLM} & ALBERT & MPNet & \textbf{MiniLM} & ALBERT & MPNet & \textbf{MiniLM} & ALBERT & MPNet \\
\midrule
ASR $\uparrow$       & $0.89$ & $0.88$ & $0.62$ & $0.85$ & $0.63$ & $0.68$ & $0.88$ & $0.89$ & $0.68$ & $0.85$ & $0.62$ & $0.68$ \\%
AUC $\uparrow$       & $0.97$ & $0.82$ & $0.66$ & $0.94$ & $0.67$ & $0.72$ & $0.96$ & $0.92$ & $0.73$ & $0.91$ & $0.65$ & $0.75$ \\
TPR@1\%FPR $\uparrow$     & $0.55$ & $0.50$ & $0.25$ & $0.59$ & $0.26$ & $0.28$ & $0.53$ & $0.58$ & $0.35$ & $0.57$ & $0.24$ & $0.25$ \\
\bottomrule
\end{tabular}
	\label{tab:I2TExtractor}
\end{table*}

\vspace{1mm}
\noindent\textbf{Impact of Different Embedding Extractors.} 
In Step~2 of our method, we employ embedding extractors to map data samples into a numerical embedding space for subsequent computation. We investigate whether the choice of embedding extractor has a significant impact on the performance of our method. Specifically, for text-to-text generative models, we additionally consider two widely used embedding extractors: RoBERTa-base \cite{roberta} and all-MiniLM-L6-v2 \cite{MiniLM}. For text-to-image generative models, we evaluate two additional visual embedding extractors: ViT-base \cite{VIT} and CLIP-ViT \cite{CLIP-VIT}. For image-to-text generative models, we further include ALBERT-large-v2 \cite{ALBERT} and all-MPNet-base-v2 \cite{MPNet} as alternatives. 

Tables~\ref{tab:T2TExtractor}, \ref{tab:T2IExtractor}, and \ref{tab:I2TExtractor} report the performance of our method across the three classes of generative models. The results show that, under the partial-knowledge setting, the choice of embedding extractor does affect the attack performance. For instance, in the image-to-text setting (LLaVA + COCO2017 in Table \ref{tab:I2TExtractor}), when using MPNet-base as the embedding extractor under the partial-knowledge scenario, all three evaluation metrics, ASR, AUC, and TPR@1\%FPR, decrease compared to using MiniLM as the extractor.
This performance degradation can be attributed to differences in semantic sensitivity across embedding extractors. In particular, MiniLM is optimized for sentence-level semantic similarity, which is more effective at capturing fine-grained distributional shifts between member and non-member samples in the embedding space. In contrast, MPNet-base prioritizes global contextual encoding, which may smooth out subtle membership-related signals and reduce distributional separability.

Similar trends are observed in the zero-knowledge setting. Notably, a closer inspection of Table~\ref{tab:T2IExtractor} reveals that for text-to-image models, the ViT-Base and CLIP-ViT embedding extractors perform well under the partial-knowledge setting but exhibit noticeably degraded performance in the zero-knowledge setting.
This discrepancy can be attributed to the fact that ViT-Base and CLIP-ViT rely more heavily on high-level semantic alignment learned from large-scale paired data, which is beneficial when some prior knowledge about the training distribution is available. In the zero-knowledge setting, however, the lack of training data information makes it more difficult for these extractors to capture subtle distributional differences between member and non-member samples. As a result, the induced embedding distributions become less separable, leading to reduced membership inference performance.

Additionally, as shown in Table~\ref{tab:I2TExtractor}, when using ALBERT-large-v2 as the embedding extractor, the proposed method achieves strong performance under the LLaVA + COCO2017 configuration, but exhibits noticeable performance degradation under the MiniGPT4 + CC\_SBU\_ALIGN configuration, in both partial-knowledge and zero-knowledge settings. This discrepancy may be caused by the differences in the semantic characteristics of the underlying datasets as well as the representational bias of the extractor. In particular, ALBERT is optimized for textual semantic compression and sentence-level similarity, which aligns well with the linguistic structure of COCO2017. However, CC\_SBU\_ALIGN contains more diverse web-scale captions, making it harder for ALBERT to capture fine-grained cross-modal correspondences. 

\vspace{1mm}
\noindent\textbf{Impact of Different Distribution Estimation Methods.} 
In Step~3 of our method, the embedding distribution is estimated by computing the average vector and the covariance matrix of the extracted embeddings. We investigate whether alternative strategies for estimating the embedding distribution affect the performance of our method. Specifically, instead of using the average vector, we consider the geometric median of the embeddings, defined as: $argmin_z \sum_i \lVert E(x^{(i)}) - z \rVert_2$, as a robust estimator of central tendency.

\begin{figure}[ht!]
\centering
	\begin{minipage}{1\textwidth}
    \subfigure[\small{Text-to-Text ASR}]{
    \includegraphics[scale=0.18]{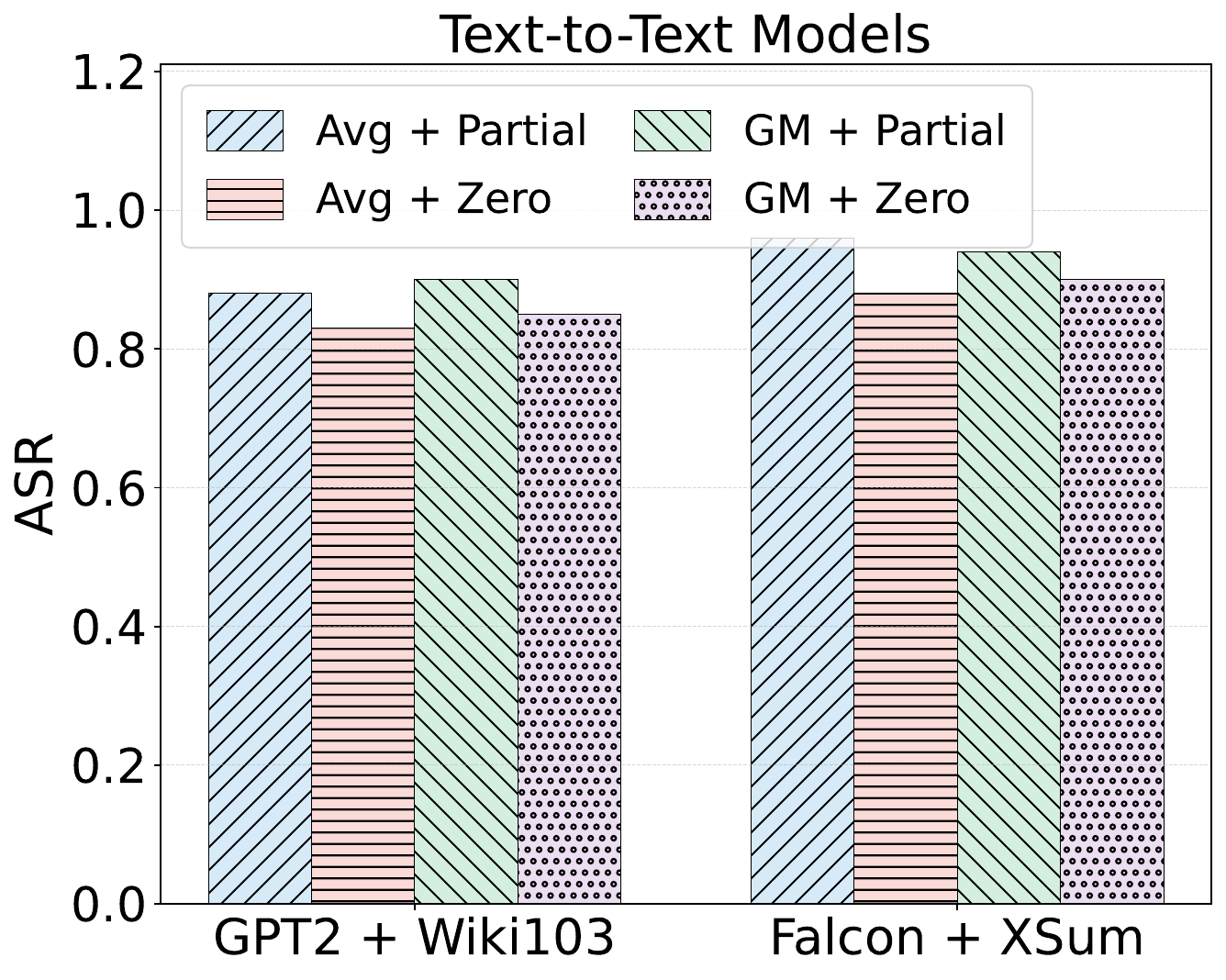}
			\label{fig:DistributeT2TASR}}
	\subfigure[\small{Text-to-Text AUC}]{
    \includegraphics[scale=0.18]{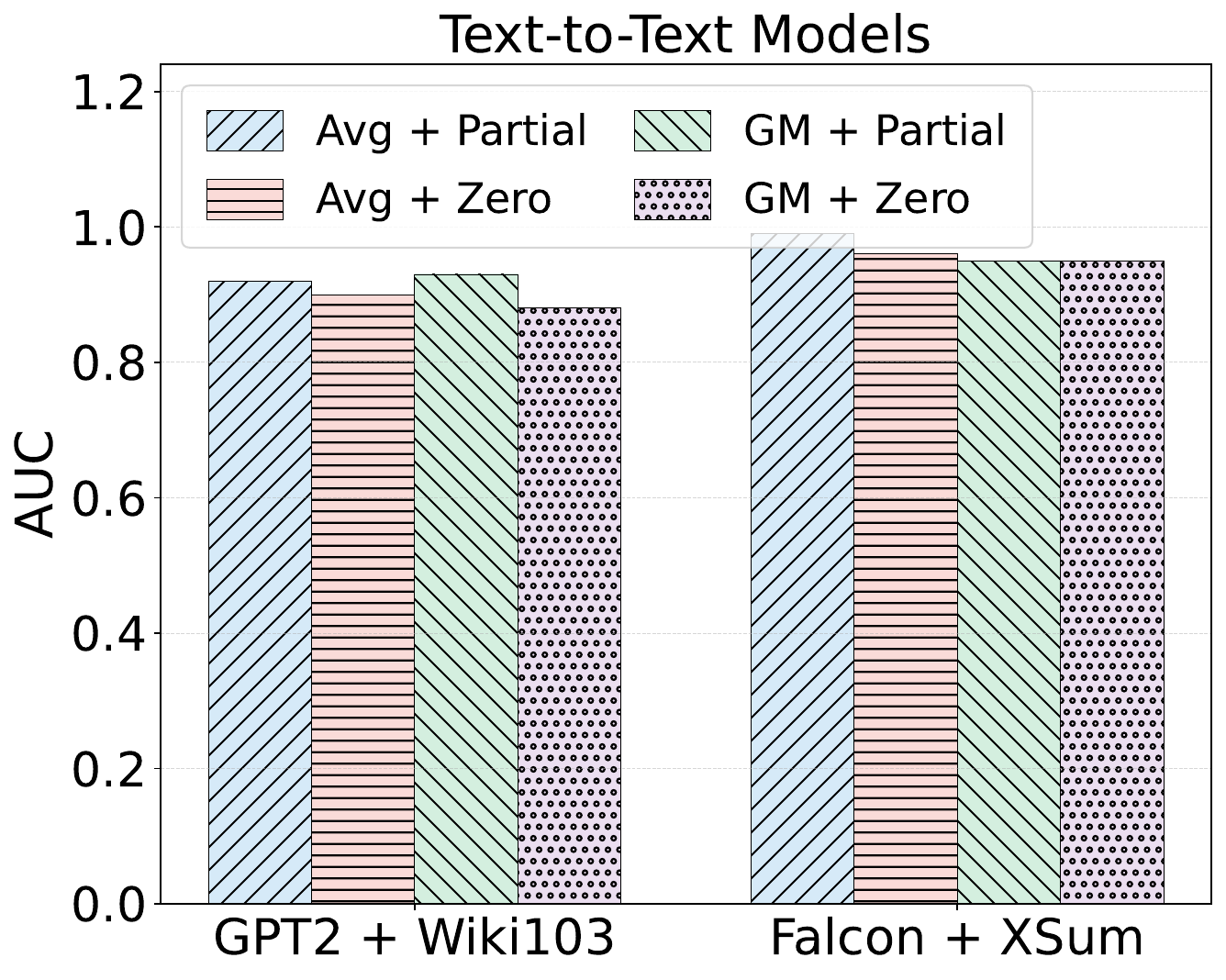}
			\label{fig:DistributeT2TAUC}}\\
    \subfigure[\small{Text-to-Image ASR}]{
    \includegraphics[scale=0.18]{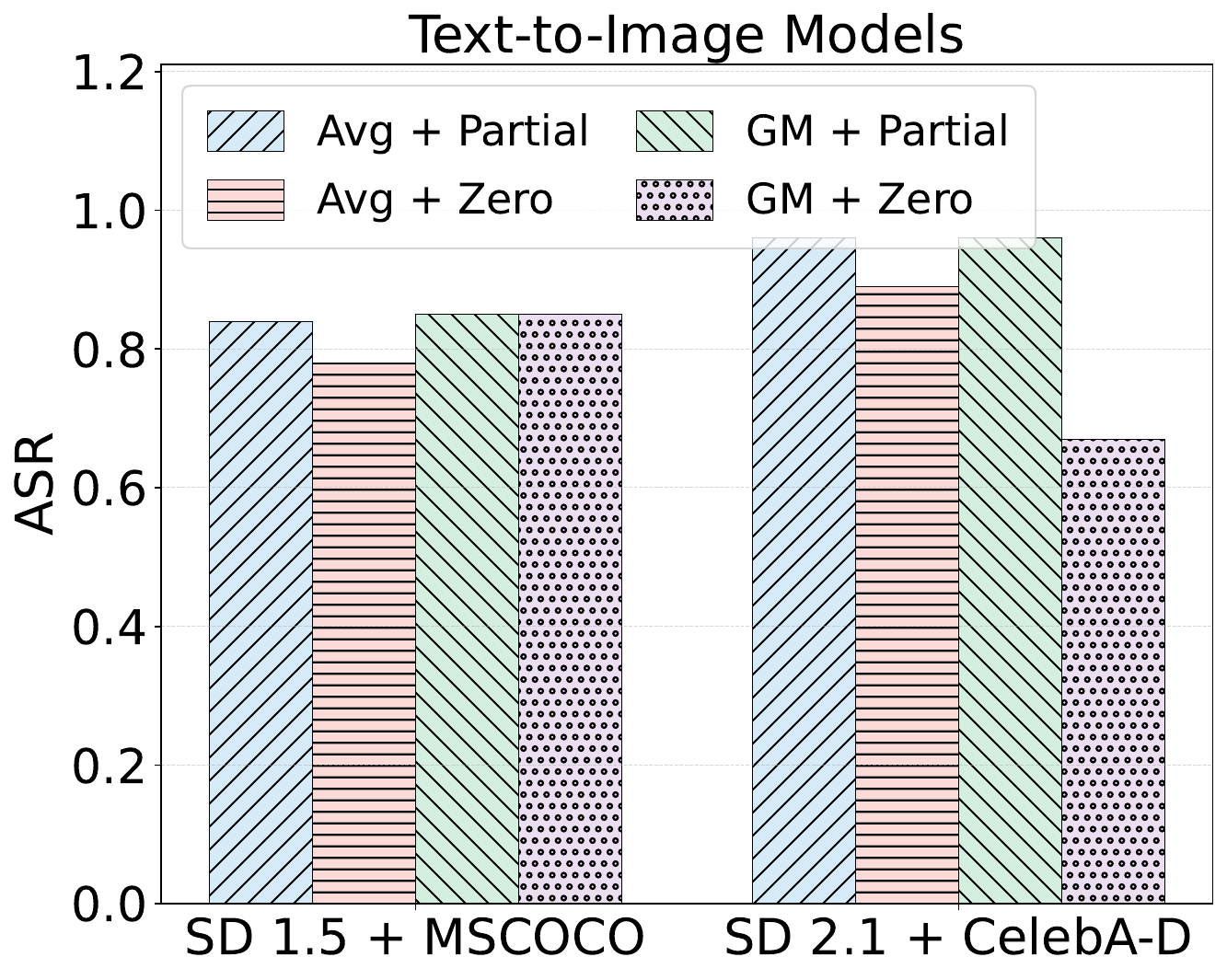}
			\label{fig:DistributeT2IASR}}
	\subfigure[\small{Text-to-Image AUC}]{
    \includegraphics[scale=0.18]{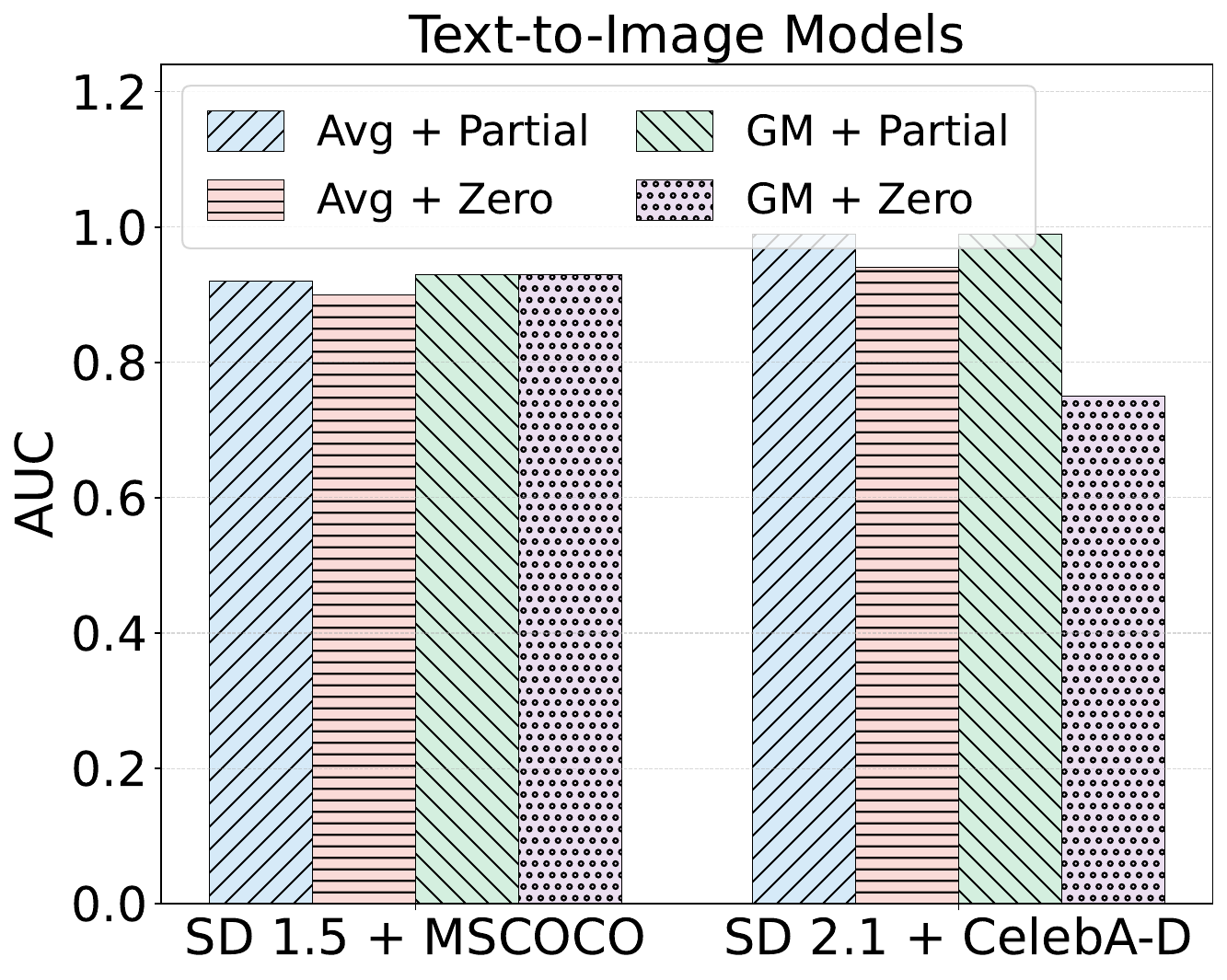}
			\label{fig:DistributeT2IAUC}}\\
    \subfigure[\small{Image-to-Text ASR}]{
    \includegraphics[scale=0.18]{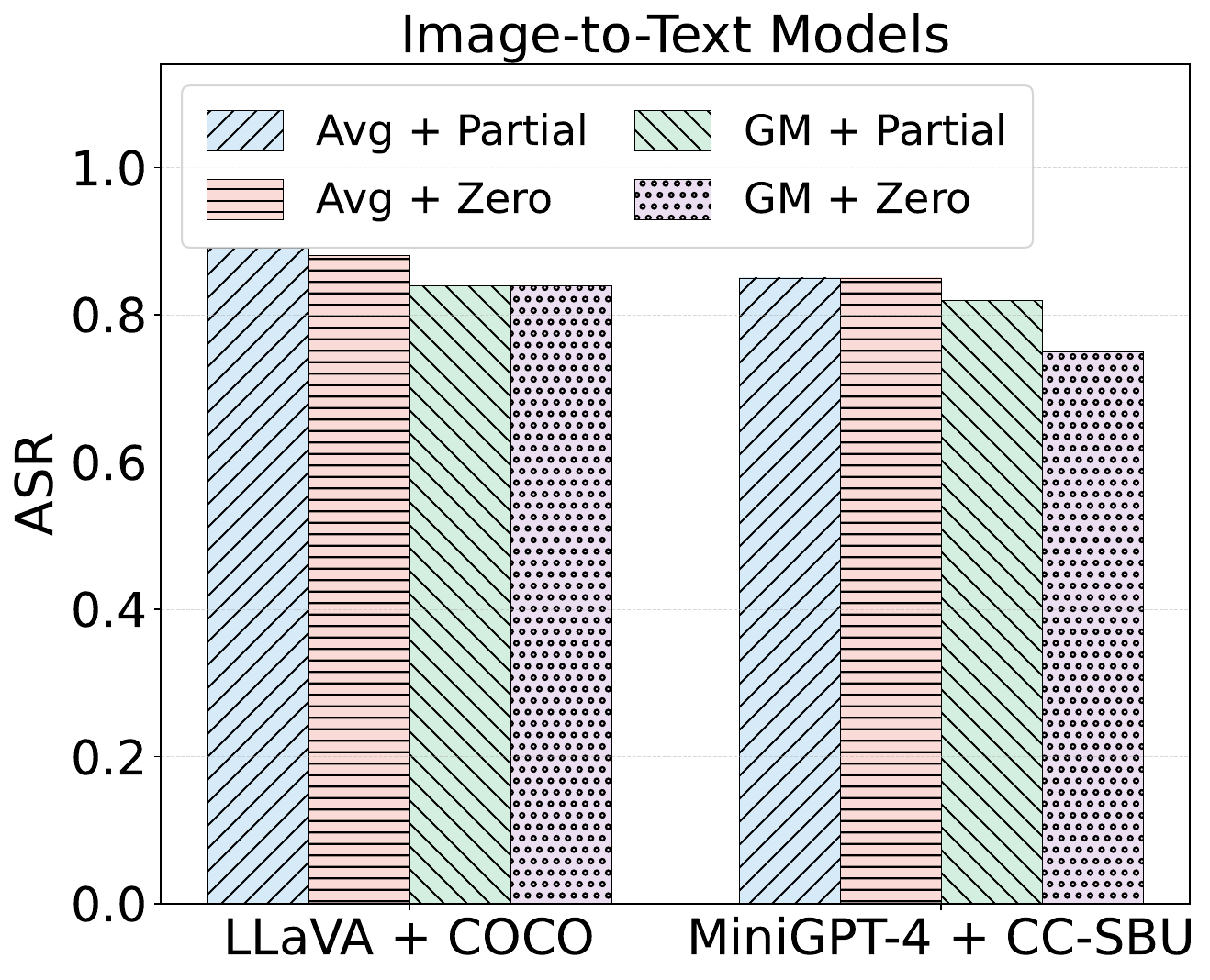}
			\label{fig:DistributeI2TASR}}
	\subfigure[\small{Image-to-Text AUC}]{
    \includegraphics[scale=0.18]{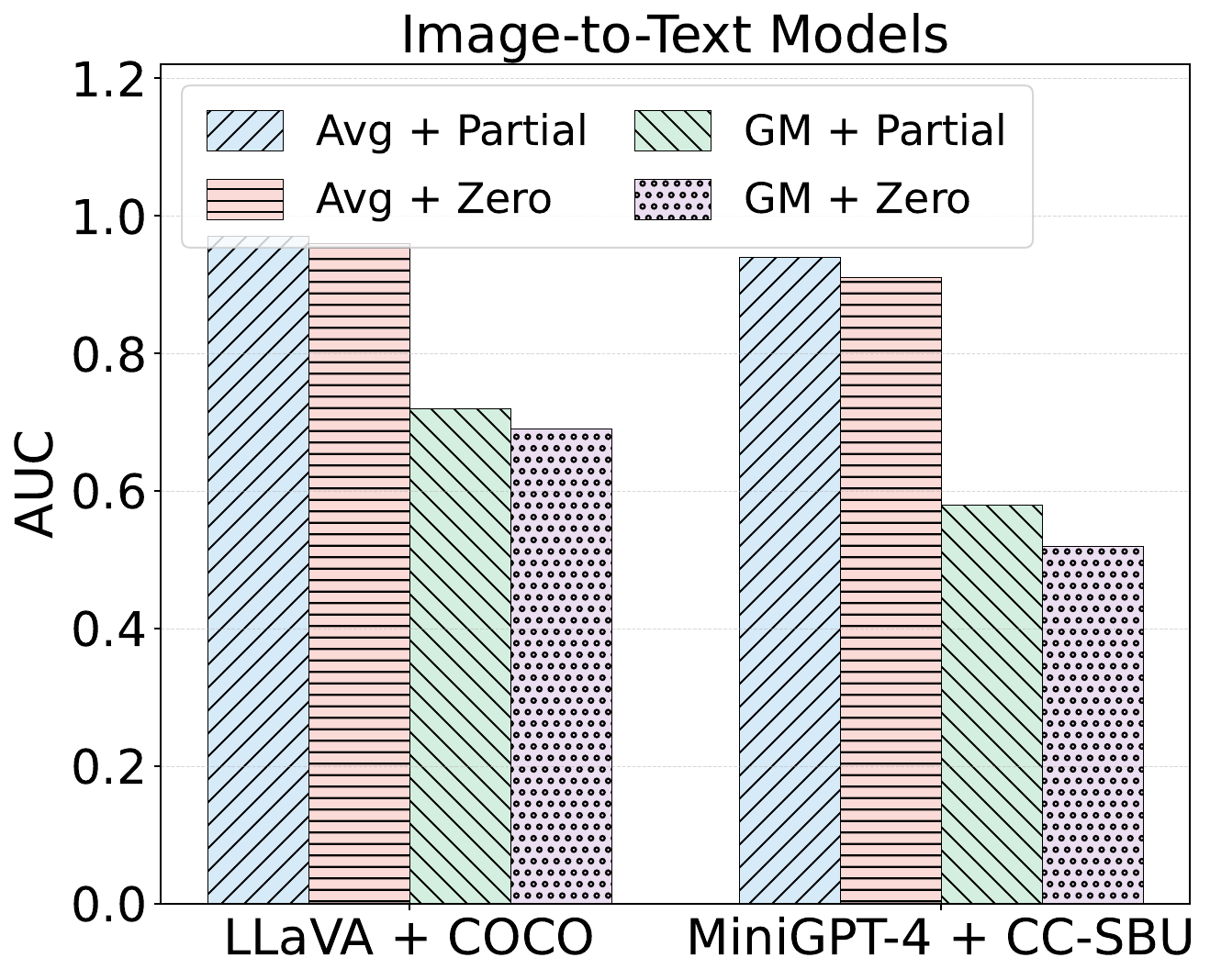}
			\label{fig:DistributeI2TAUC}}\\
    \end{minipage}
	\caption{ASR and AUC Performance of Our Method with Different Distribution Estimation Methods}
	\label{fig:DistributeASRAUC}
\end{figure}

\begin{figure}[ht!]
\centering
	\begin{minipage}{1\textwidth}
    \subfigure[\small{Text-to-Text TPR}]{
    \includegraphics[scale=0.18]{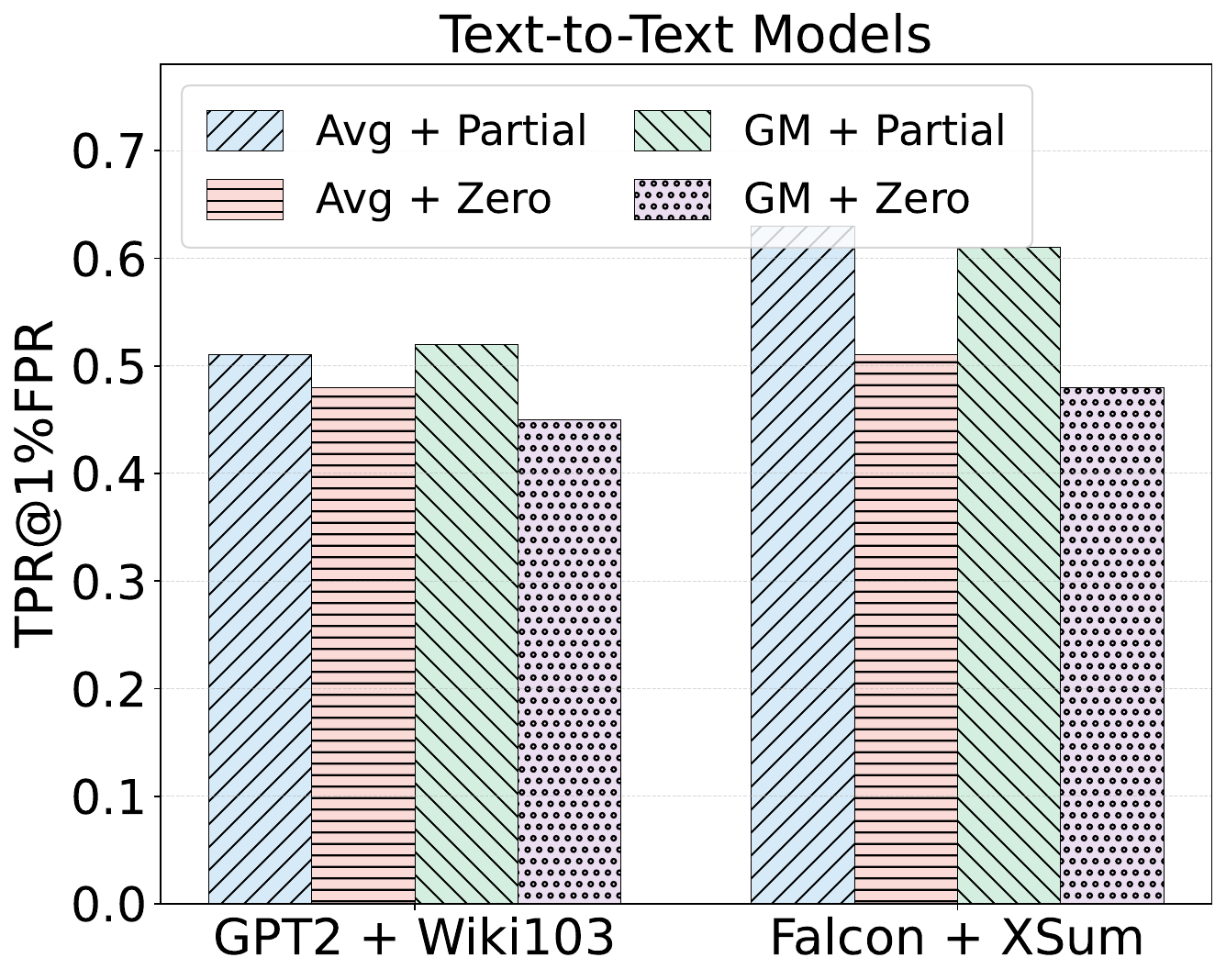}
			\label{fig:DistributeT2TTPR}}
    \subfigure[\small{Text-to-Image TPR}]{
    \includegraphics[scale=0.18]{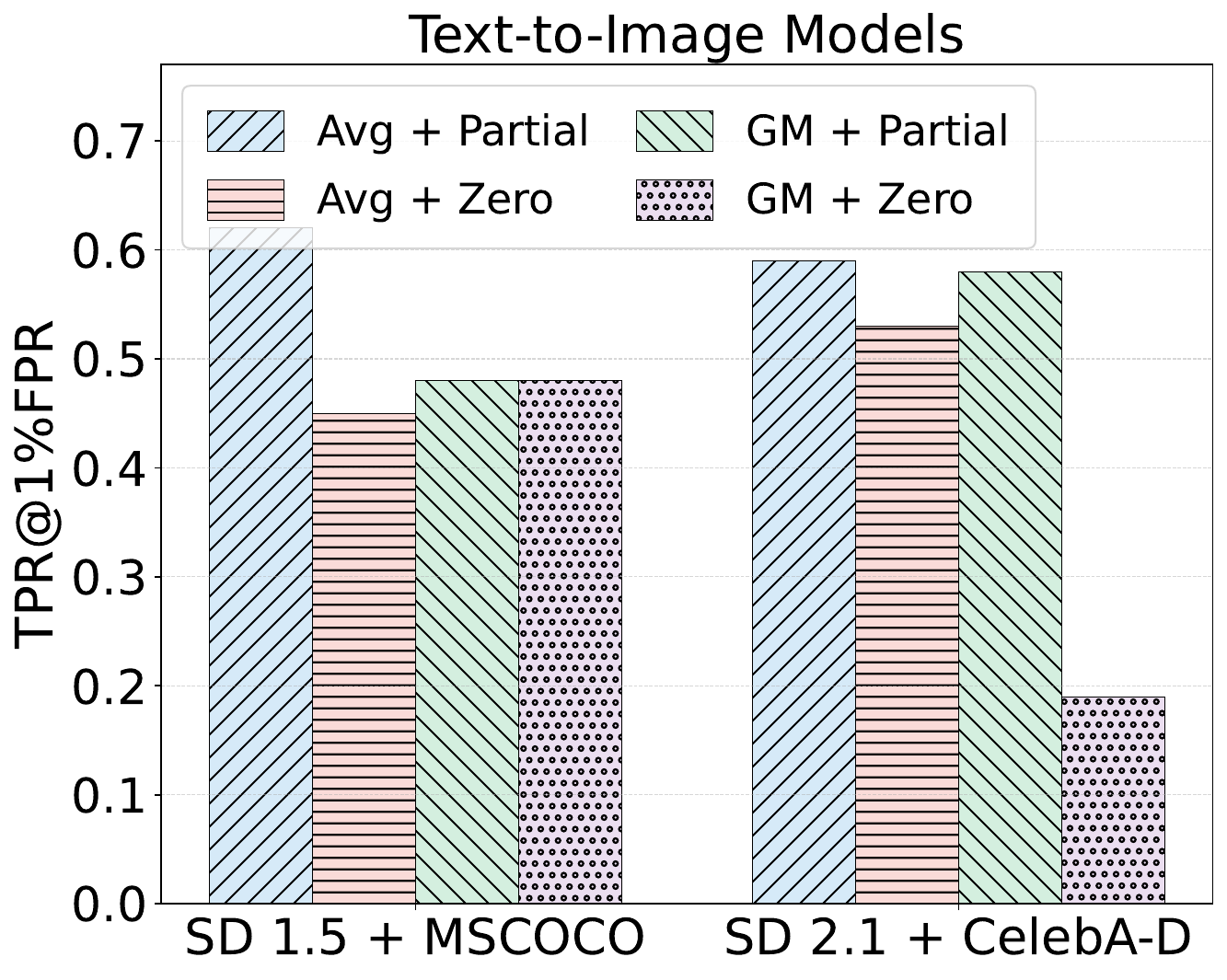}
			\label{fig:DistributeT2ITPR}}\\
    \subfigure[\small{Image-to-Text TPR}]{
    \includegraphics[scale=0.18]{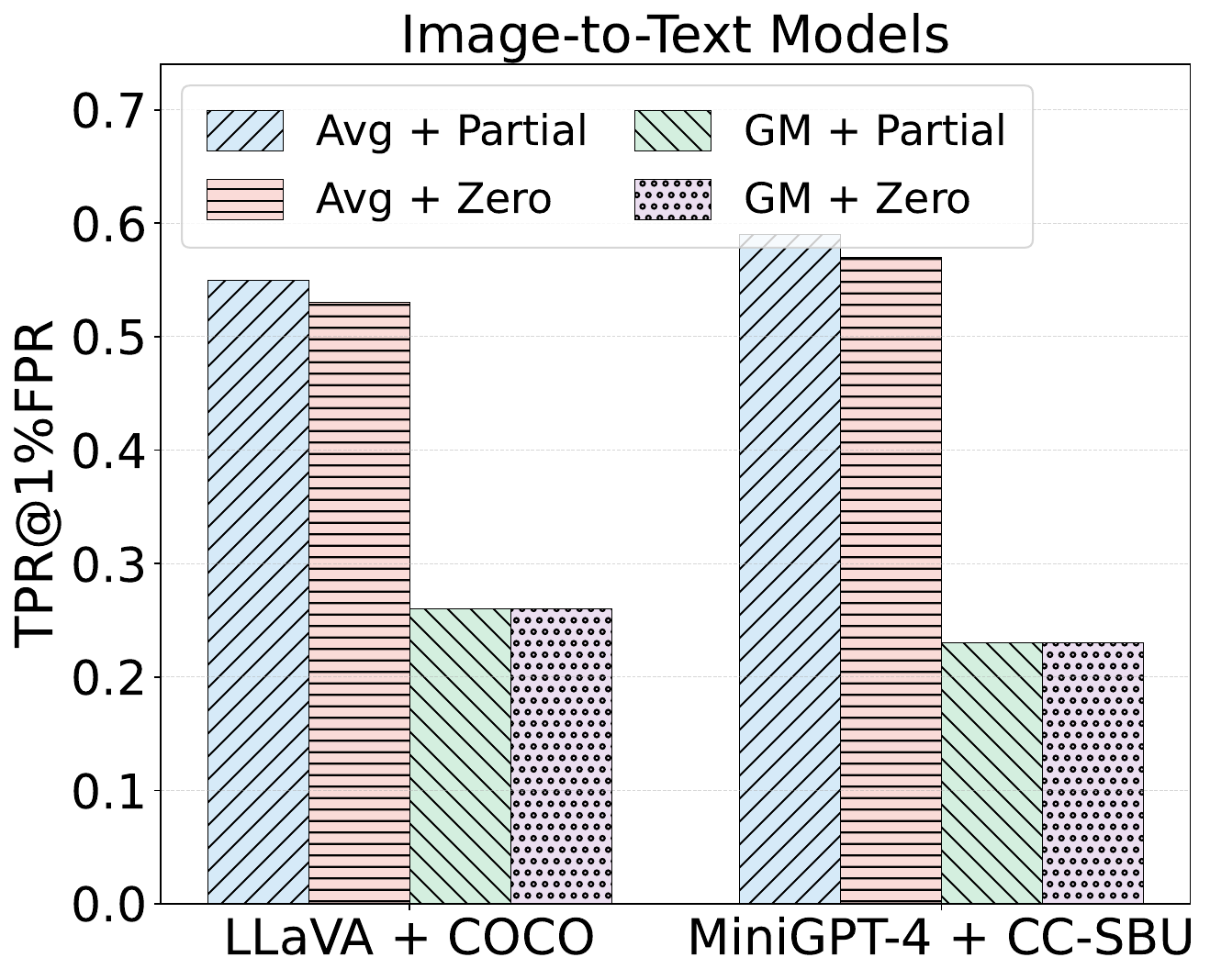}
			\label{fig:DistributeI2TTPR}}
    \end{minipage}
	\caption{TPR@1\%FPR Performance of Our Method with Different Distribution Estimation Methods}
	\label{fig:DistributeTPR}
\end{figure}

The results of ASR and AUC are shown in Figure~\ref{fig:DistributeASRAUC}, while the corresponding TPR@1\%FPR results are reported in Figure \ref{fig:DistributeTPR}. We observe that, in Figure \ref{fig:DistributeASRAUC}, using the geometric median to represent the embedding distribution yields performance comparable to using the average vector for text-to-text and text-to-image generative models. However, for image-to-text generative models, adopting the geometric median leads to a noticeable degradation in performance compared to the average-based estimator (Figure \ref{fig:DistributeI2TAUC}).
This performance drop is due to the fact that image-to-text embeddings tend to exhibit higher variance and more complex, multimodal structures. 
While the geometric median is robust to outliers, it captures only the central location of the distribution and ignores second-order statistics that are critical for distinguishing subtle membership-related distributional shifts.

Moreover, in Figure \ref{fig:DistributeTPR}, focusing on the image-to-text setting (Figure~\ref{fig:DistributeI2TTPR}), we observe that using the geometric median to estimate the embedding distribution leads to substantially worse performance than the average-based estimation adopted in our method.
This degradation arises because image-to-text embeddings often exhibit high variance and complex multimodal distributions. While the geometric median is robust to outliers, it captures only the central tendency of the distribution and discards important second-order information, thus providing an insufficient representation for likelihood-based membership inference.

\vspace{-0mm}

\vspace{1mm}
\noindent\textbf{Impact of Replacing Likelihood Scores with Distance-Based Metrics.} 
In Step~4 of our method, we determine the membership status of a target sample based on the log-likelihood score of its embedding (Eq. \ref{eq:score}). We now evaluate the importance of using the log-likelihood score by considering an alternative, distance-based decision rule. Specifically, instead of computing log-likelihoods, we directly compare the embedding of the target sample, $E(x^*)$, with the average embeddings of synthetic data, $\mu_{x_{\mathrm{syn}}}$, and real (auxiliary) data, $\mu_{x_{\mathrm{real}}}$ (or $\mu_{x_{\mathrm{aux}}}$). If $\mathrm{dist}(E(x^*), \mu_{x_{\mathrm{syn}}}) < \mathrm{dist}(E(x^*), \mu_{x_{\mathrm{real}}})$ (or $\mathrm{dist}(E(x^*), \mu_{x_{\mathrm{syn}}}) < \mathrm{dist}(E(x^*), \mu_{x_{\mathrm{aux}}})$), that is, if the target sample $x^*$ is closer to the synthetic data than to the real (auxiliary) data in the embedding space, it is inferred to be a member.
We evaluate this distance-based variant using two commonly adopted distance metrics, namely cosine similarity and Wasserstein distance. 

The results are reported in Tables~\ref{tab:T2TDistance}, \ref{tab:T2IDistance}, and \ref{tab:I2TDistance}. We observe that replacing the log-likelihood score with a distance-based decision rule leads to consistently degraded performance in most settings, regardless of whether cosine similarity or Wasserstein distance is used, and under both partial-knowledge and zero-knowledge threat models. This performance drop highlights the importance of the log-likelihood score, which explicitly incorporates both first-order and second-order statistics of the embedding distribution, rather than relying solely on pointwise proximity. An interesting phenomenon arises for text-to-text generative models (Table~\ref{tab:T2TDistance}), where using the Wasserstein distance yields relatively strong performance in the partial-knowledge setting but performs poorly in the zero-knowledge setting. This behavior can be attributed to the fact that, under partial knowledge, the adversary has access to auxiliary samples that partially reflect the true training distribution, allowing Wasserstein distance to capture coarse distributional differences. In contrast, in the zero-knowledge setting, the lack of reliable auxiliary information makes distance-based comparisons highly sensitive to noise and distributional mismatch, causing the Wasserstein distance to lose discriminative power.

\begin{table*}[ht!]\scriptsize
	\centering
 	\caption{Results of Our Method on Text-to-Text Models with Distance-based Metrics}
\begin{tabular} {cccc|ccc|ccc|ccc}
\toprule
& \multicolumn{6}{c|}{Partial-Knowledge Setting} & \multicolumn{6}{c}{Zero-Knowledge Setting} \\\cline{2-13}
& \multicolumn{3}{c|}{GPT2+Wiki103}\centering & \multicolumn{3}{c|}{Falcon+XSum} & \multicolumn{3}{c|}{GPT2+Wiki103}\centering & \multicolumn{3}{c}{Falcon+XSum}\\\cline{2-13}
&  \textbf{Likelihood} & Wassers. & Cosine  & \textbf{Likelihood} & Wassers. & Cosine & \textbf{Likelihood} & Wassers. & Cosine & \textbf{Likelihood} & Wassers. & Cosine \\
\midrule
ASR $\uparrow$       & $0.93$ & $0.94$ & $0.49$ & $0.98$ & $0.98$ & $0.46$ & $0.83$ & $0.51$ & $0.73$ & $0.88$ & $0.45$ & $0.81$ \\%
AUC $\uparrow$       & $0.97$ & $0.97$ & $0.46$ & $0.99$ & $0.99$ & $0.44$ & $0.90$ & $0.49$ & $0.82$ & $0.96$ & $0.44$ & $0.89$ \\
T@1\%F $\uparrow$     & $0.51$ & $0.49$ & $0.10$ & $0.58$ & $0.56$ & $0.04$ & $0.48$ & $0.11$ & $0.14$ & $0.51$ & $0.04$ & $0.34$ \\
\bottomrule
\end{tabular}
	\label{tab:T2TDistance}
\end{table*}

\begin{table*}[ht!]\scriptsize
	\centering
 	\caption{Results of Our Method on Text-to-Image Models with Distance-based Metrics}
\begin{tabular} {cccc|ccc|ccc|ccc}
\toprule
& \multicolumn{6}{c|}{Partial-Knowledge Setting} & \multicolumn{6}{c}{Zero-Knowledge Setting} \\\cline{2-13}
& \multicolumn{3}{c|}{SD1.5+MSCOCO}\centering & \multicolumn{3}{c|}{SD2.1+CelebA-Dialog} & \multicolumn{3}{c|}{SD1.5+MSCOCO}\centering & \multicolumn{3}{c}{SD2.1+CelebA-Dialog}\\\cline{2-13}
& \textbf{Likelihood} & Wassers. & Cosine  & \textbf{Likelihood} & Wassers. & Cosine & \textbf{Likelihood} & Wassers. & Cosine & \textbf{Likelihood} & Wassers. & Cosine  \\
\midrule
ASR $\uparrow$        & $0.84$ & $0.59$ & $0.64$ & $0.96$ & $0.80$ & $0.97$ & $0.78$ & $0.59$ & $0.75$ & $0.89$ & $0.80$ & $0.87$ \\%
AUC $\uparrow$        & $0.92$ & $0.77$ & $0.71$ & $0.99$ & $0.97$ & $0.99$ & $0.90$ & $0.76$ & $0.83$ & $0.94$ & $0.95$ & $0.93$ \\
T@1\%F $\uparrow$     & $0.62$ & $0.21$ & $0.25$ & $0.59$ & $0.46$ & $0.65$ & $0.45$ & $0.15$ & $0.19$ & $0.53$ & $0.46$ & $0.68$ \\
\bottomrule
\end{tabular}
	\label{tab:T2IDistance}
\end{table*}

\begin{table*}[ht!]\scriptsize
	\centering
 	\caption{Results of Our Method on Image-to-Text Models with Distance-based Metrics}
\begin{tabular} {cccc|ccc|ccc|ccc}
\toprule
& \multicolumn{6}{c|}{Partial-Knowledge Setting} & \multicolumn{6}{c}{Zero-Knowledge Setting} \\\cline{2-13}
& \multicolumn{3}{c|}{LLaVA+COCO}\centering & \multicolumn{3}{c|}{MiniGPT4+CC\_SBU} & \multicolumn{3}{c|}{LLaVA+COCO}\centering & \multicolumn{3}{c}{MiniGPT4+CC\_SBU}\\\cline{2-13}
& \textbf{Likelihood} & Wassers. & Cosine & \textbf{Likelihood} & Wassers. & Cosine & \textbf{Likelihood} & Wassers. & Cosine & \textbf{Likelihood} & Wassers. & Cosine  \\
\midrule
ASR $\uparrow$       & $0.89$ & $0.70$ & $0.69$ & $0.85$ & $0.56$ & $0.68$ & $0.88$ & $0.69$ & $0.69$ & $0.85$ & $0.56$ & $0.57$ \\%
AUC $\uparrow$       & $0.97$ & $0.68$ & $0.80$ & $0.94$ & $0.53$ & $0.65$ & $0.96$ & $0.80$ & $0.70$ & $0.91$ & $0.52$ & $0.56$ \\
T@1\%F $\uparrow$     & $0.55$ & $0.19$ & $0.35$ & $0.59$ & $0.10$ & $0.21$ & $0.53$ & $0.35$ & $0.26$ & $0.57$ & $0.09$ & $0.12$ \\
\bottomrule
\end{tabular}
	\label{tab:I2TDistance}
\end{table*}

\begin{figure}[ht!]
\centering
	\begin{minipage}{1\textwidth}
    \subfigure[\small{Text-to-Text ASR}]{
    \includegraphics[scale=0.155]{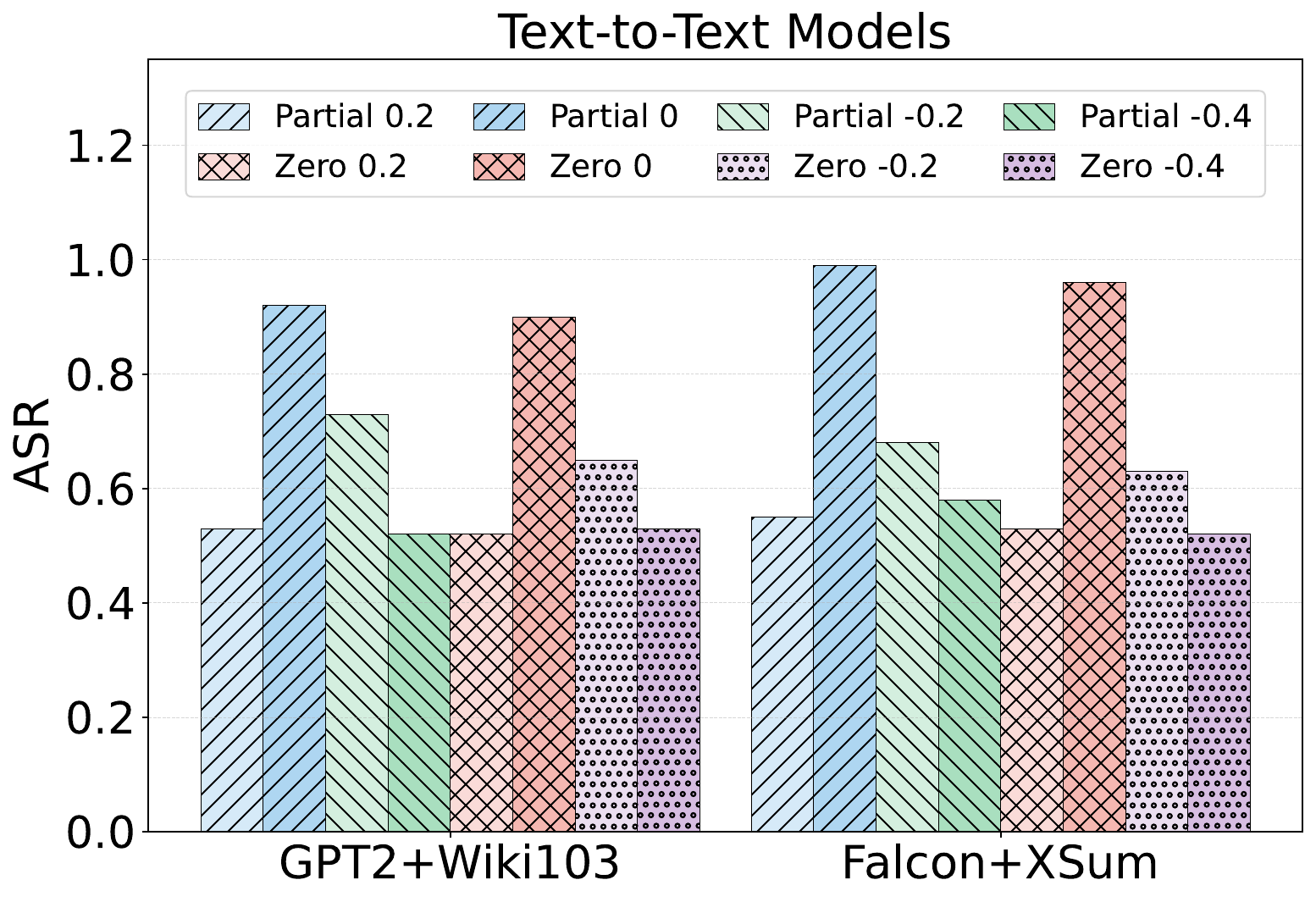}
			\label{fig:ThresholdT2TASR}}
	\subfigure[\small{Text-to-Text AUC}]{
    \includegraphics[scale=0.16]{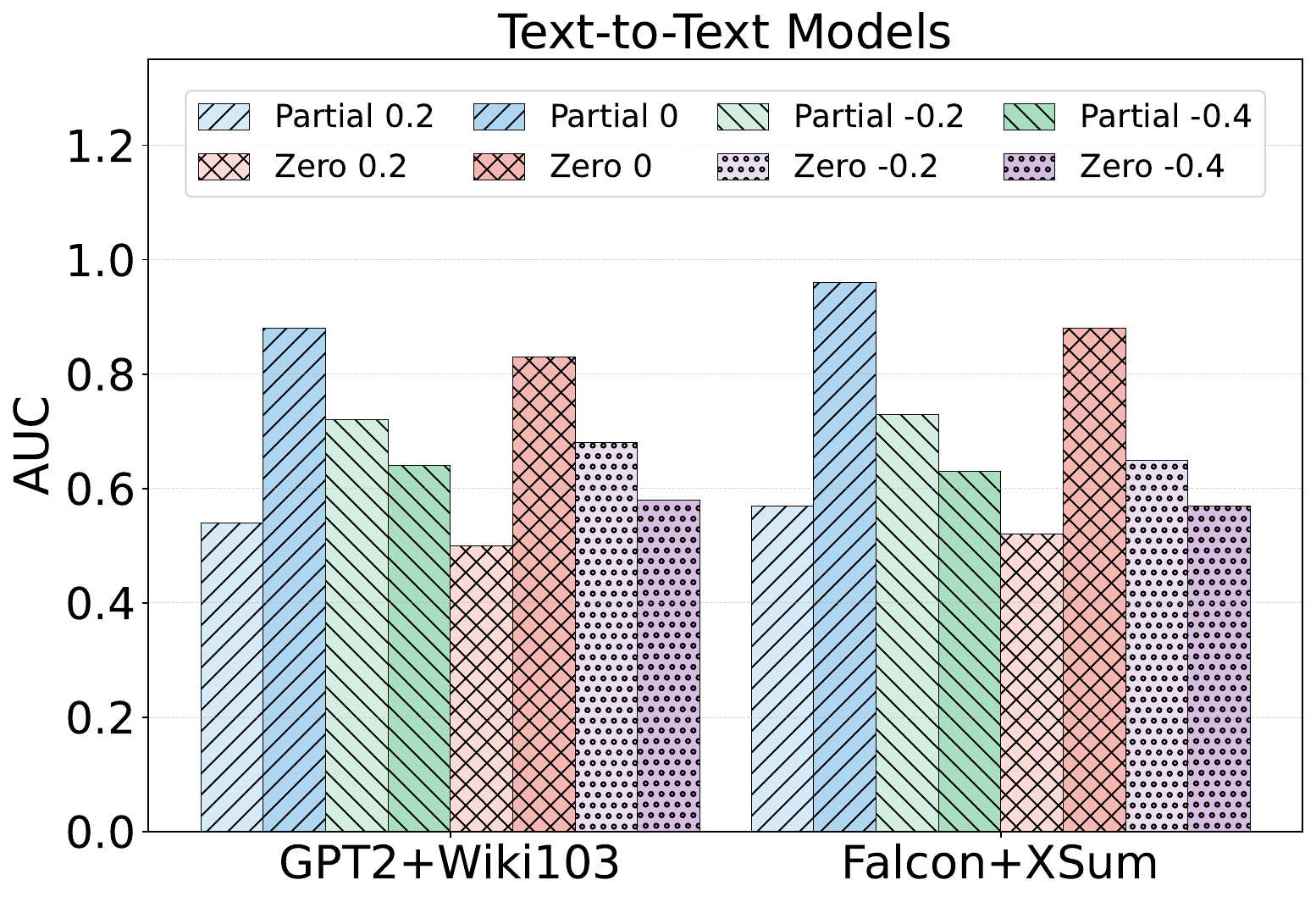}
			\label{fig:ThresholdT2TAUC}}\\
    \subfigure[\small{Text-to-Image ASR}]{
    \includegraphics[scale=0.155]{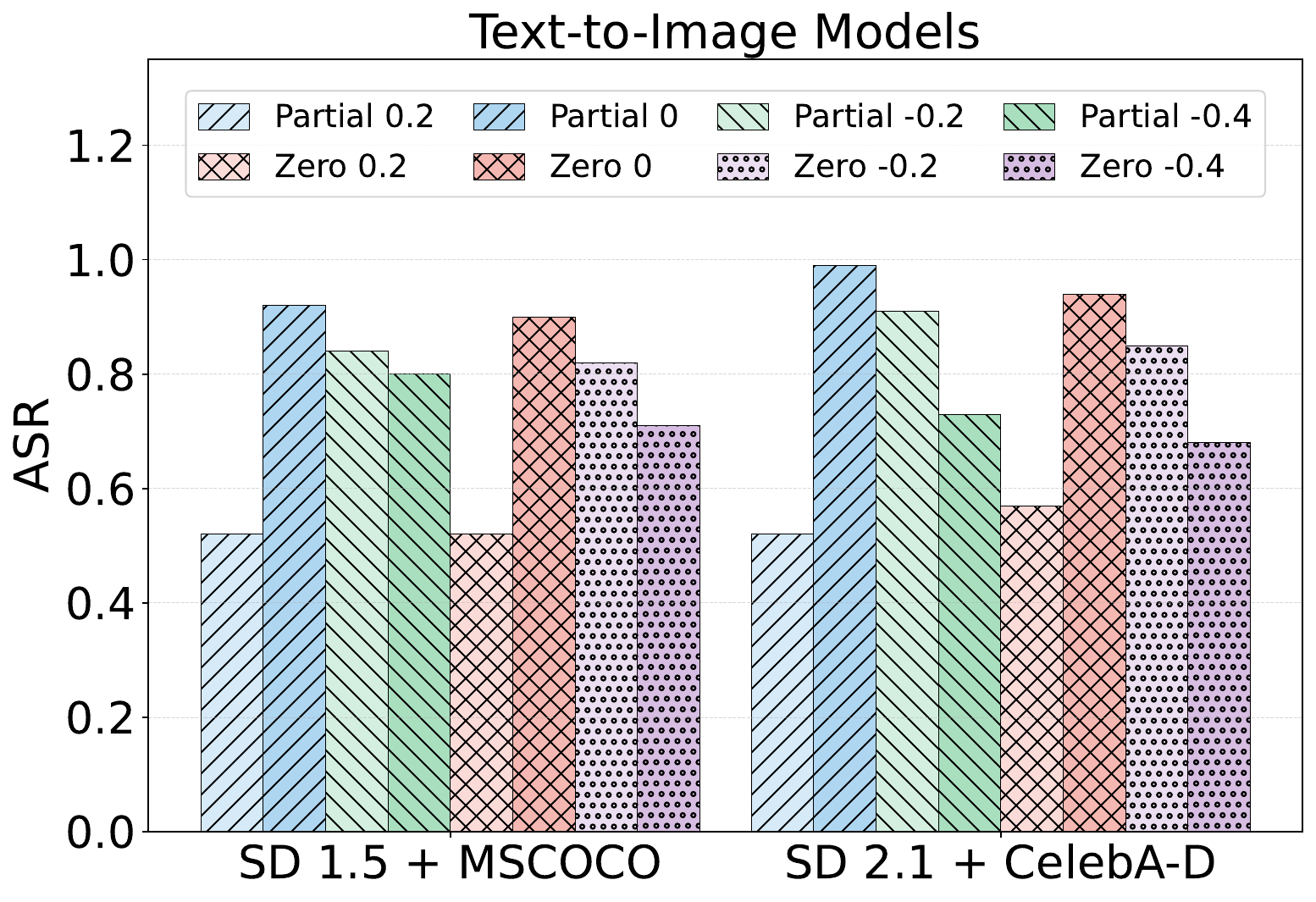}
			\label{fig:ThresholdT2IASR}}
	\subfigure[\small{Text-to-Image AUC}]{
    \includegraphics[scale=0.155]{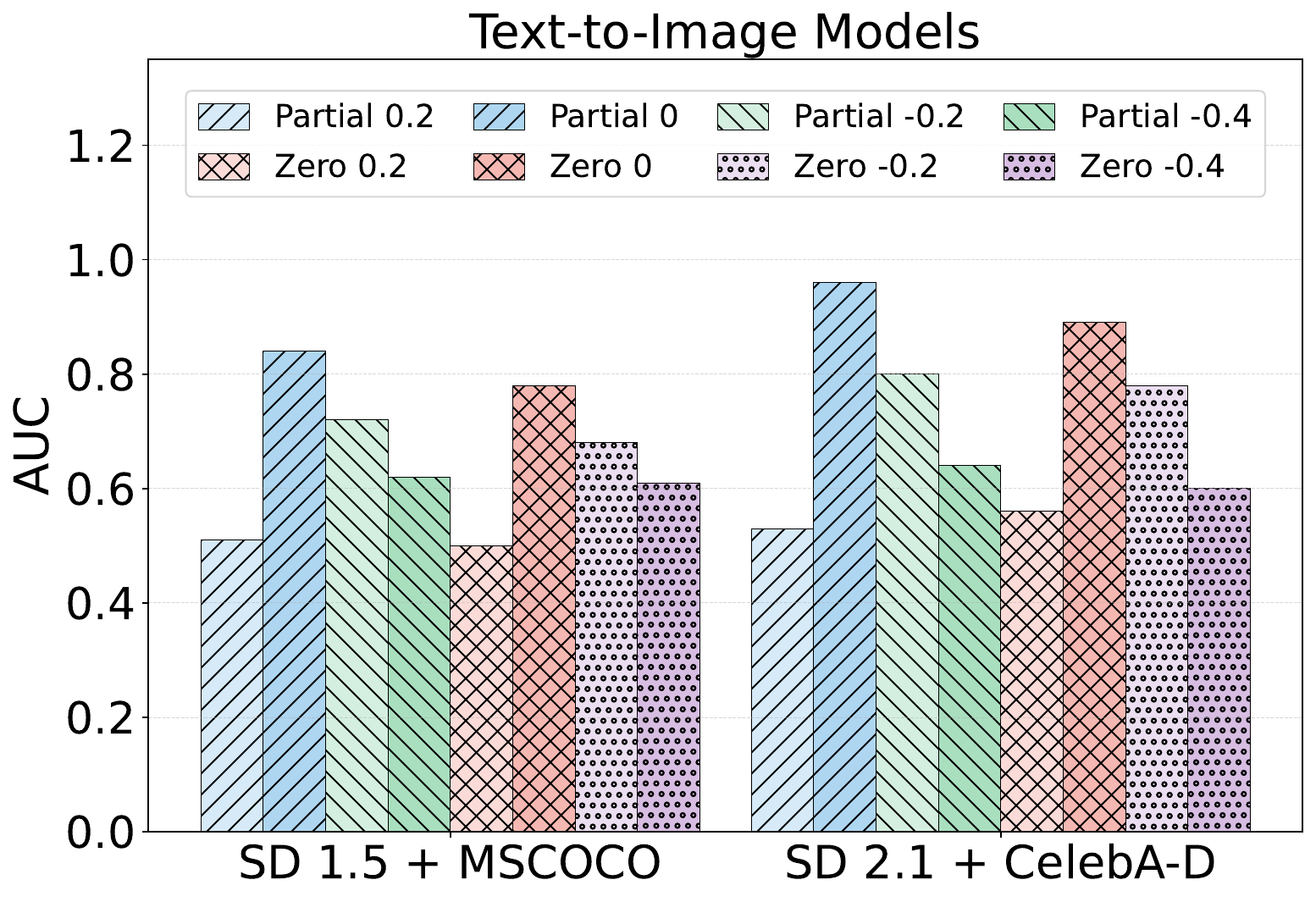}
			\label{fig:ThresholdT2IAUC}}\\
    \subfigure[\small{Image-to-Text ASR}]{
    \includegraphics[scale=0.155]{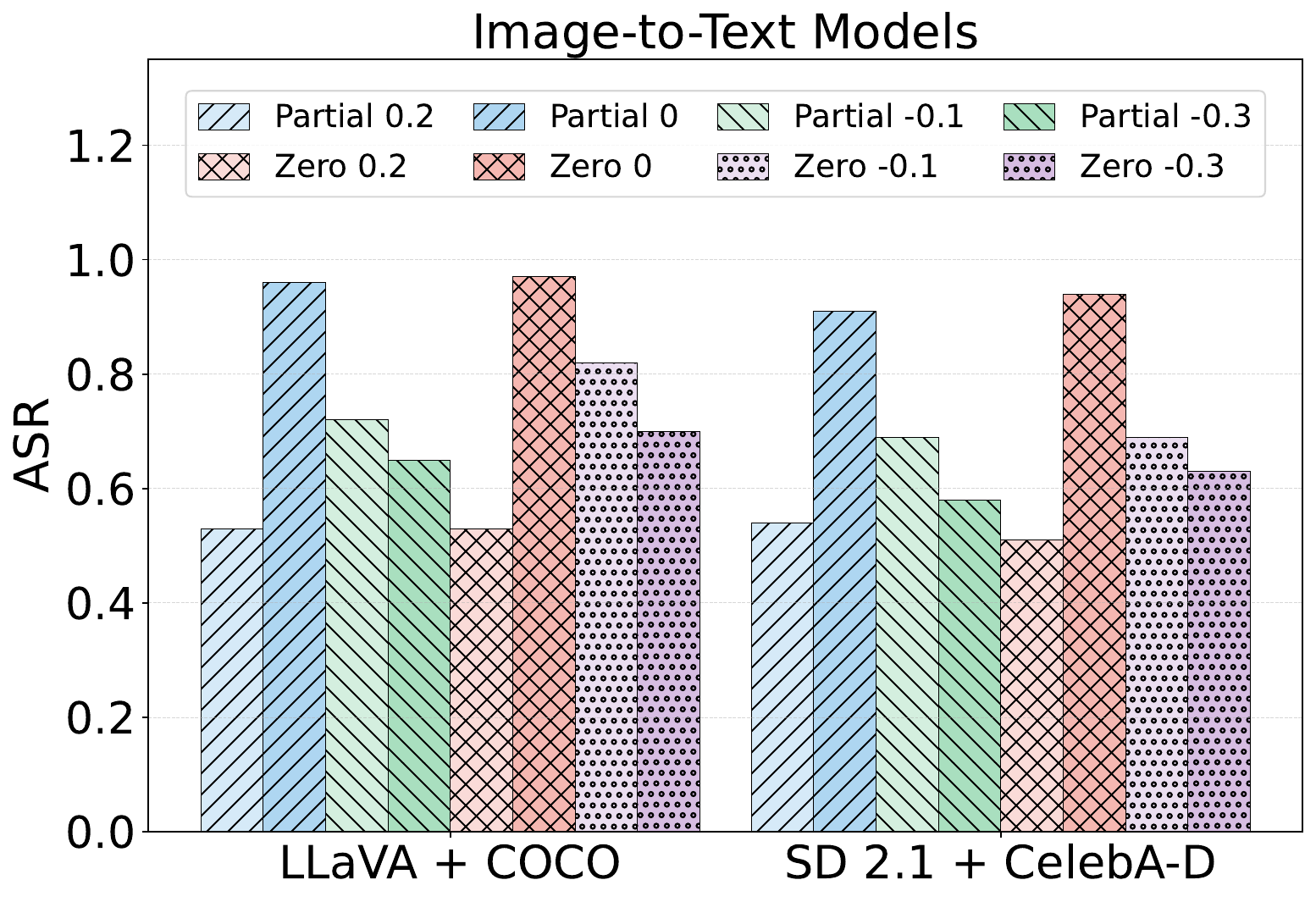}
			\label{fig:ThresholdI2TASR}}
	\subfigure[\small{Image-to-Text AUC}]{
    \includegraphics[scale=0.155]{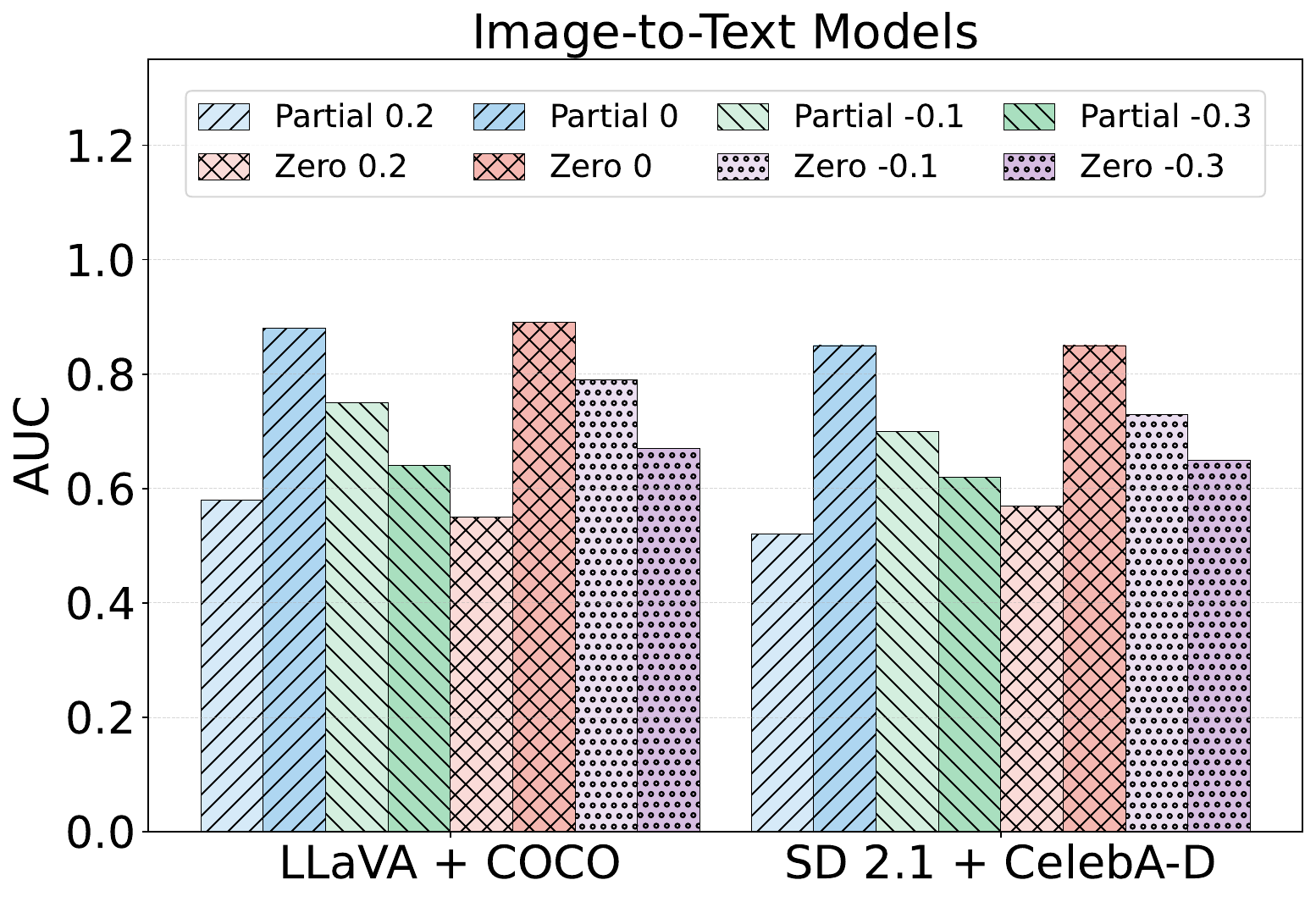}
			\label{fig:ThresholdI2TAUC}}\\
    \end{minipage}
	\caption{ASR and AUC of Our Method with Different Threshold Values}
    \vspace{2mm}
	\label{fig:ThresholdASRAUC}
\end{figure}

\begin{figure}[ht]
\centering
	\begin{minipage}{1\textwidth}
    \subfigure[\small{Text-to-Text TPR}]{
    \includegraphics[scale=0.155]{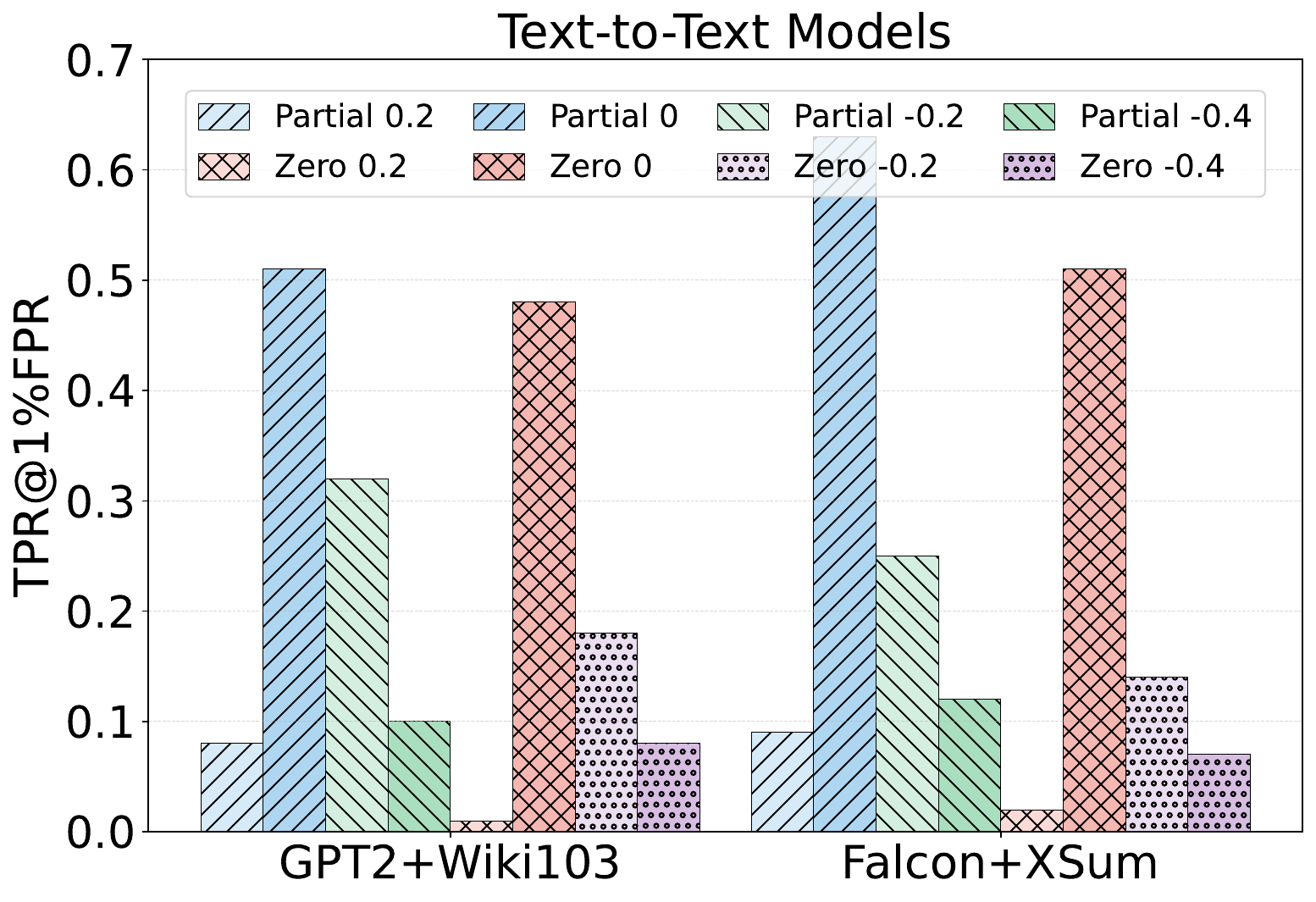}
			\label{fig:ThresholdT2TTPR}}
    \subfigure[\small{Text-to-Image TPR}]{
    \includegraphics[scale=0.155]{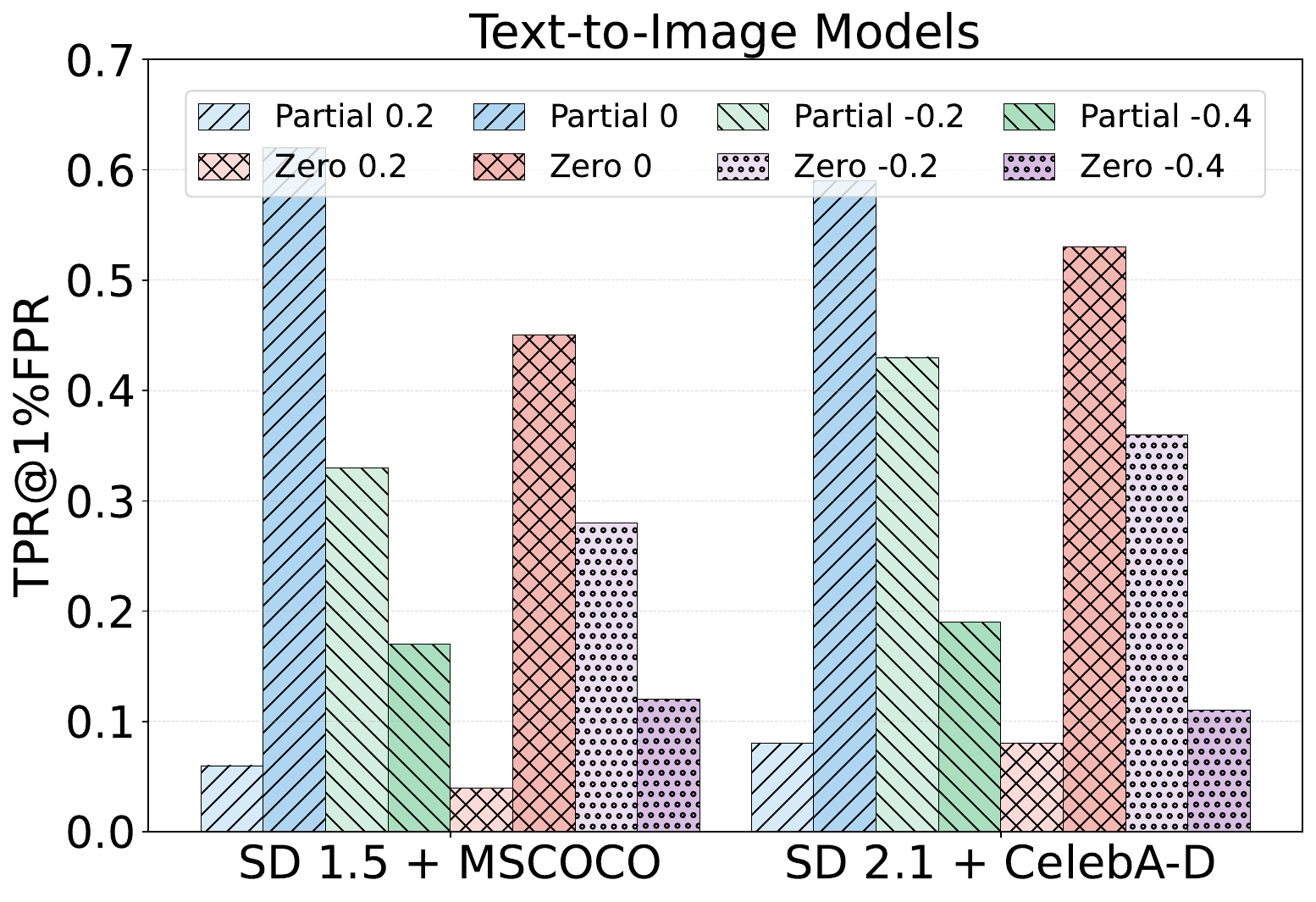}
			\label{fig:ThresholdT2ITPR}}\\
    \subfigure[\small{Image-to-Text TPR}]{
    \includegraphics[scale=0.155]{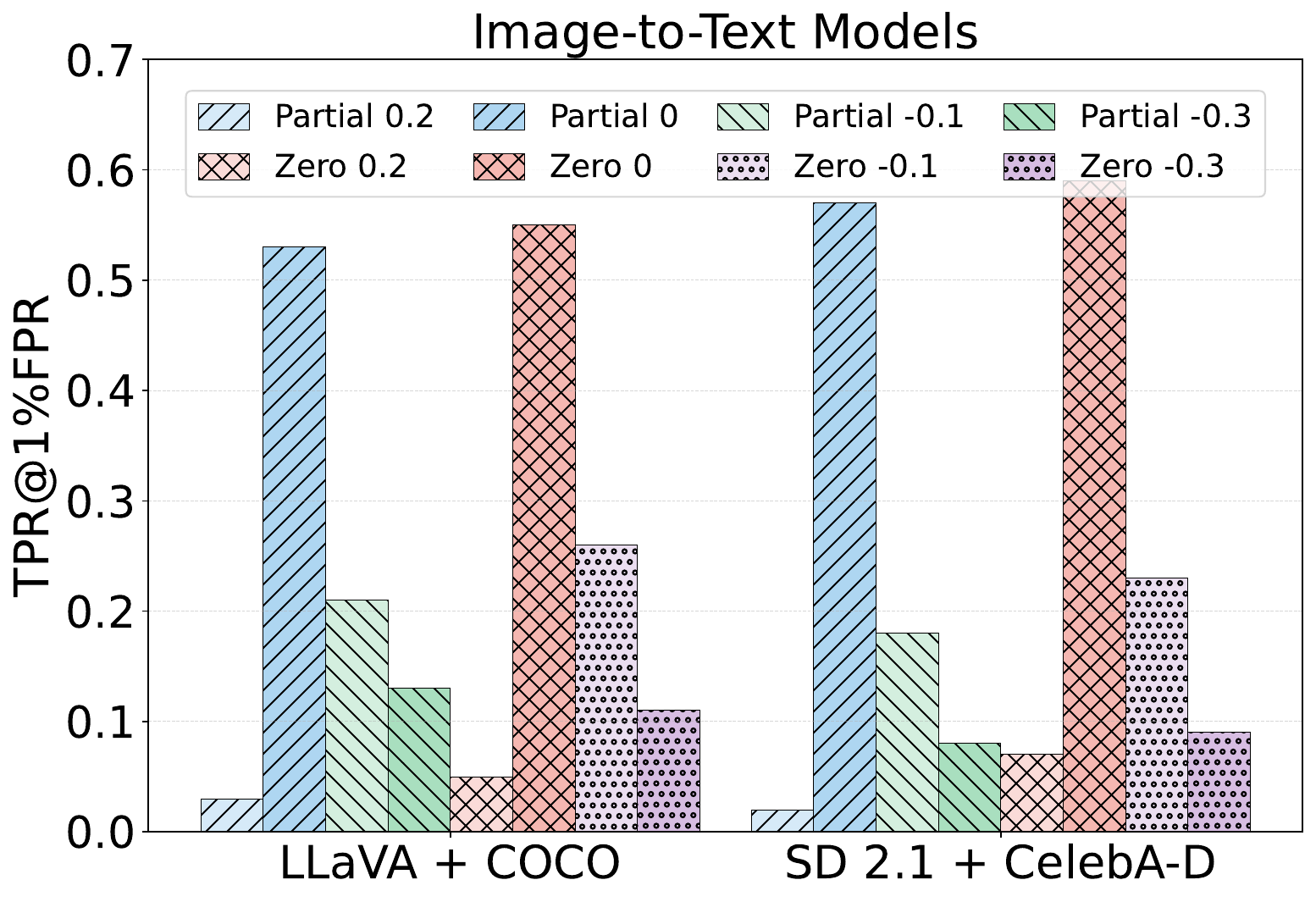}
			\label{fig:ThresholdI2TTPR}}
    \end{minipage}
	\caption{TPR@1\%FPR of Our Method with Different Threshold Values}
    \vspace{2mm}
	\label{fig:ThresholdTPR}
\end{figure}

\vspace{1mm}
\noindent\textbf{Impact of Threshold Selection on Membership Inference.} 
In Step~4 of our method, the inference threshold is implicitly set to $0$, i.e., $s_{\mathrm{real}}(E(x^*))-s_{\mathrm{syn}}(E(x^*))<0$. We investigate if varying the threshold value affects the performance of our method. The results of ASR and AUC are shown in Figure~\ref{fig:ThresholdASRAUC}, while the results of TPR@1\%FPR are presented in Figure \ref{fig:ThresholdTPR}. 
We observe that setting the threshold to $0$ consistently yields the best performance across both partial-knowledge and zero-knowledge settings under all three evaluation metrics. Deviating from this value leads to performance degradation, which aligns well with our theoretical analysis.
This behavior arises because thresholding the log-likelihood ratio at $0$ corresponds to the Bayes-optimal decision rule under equal prior probabilities. Any deviation from this optimal threshold introduces a systematic bias toward either false positives or false negatives, increasing the overall classification error.

In particular, setting the threshold to a positive value (e.g., $0.2$ in our experiments) results in substantial performance degradation. A positive threshold indicates that the target sample is closer to the real (auxiliary) data than to the synthetic data in the embedding space, which strongly suggests non-membership. Consequently, positive thresholds bias the decision rule toward predicting non-membership, leading to a significant reduction in membership inference effectiveness.

\begin{table*}[ht!]\scriptsize
	\centering
 	\caption{Results of Our Method Using the Comparison Rule vs. a Classifier Across Three Categories of Generative Models in the Partial Knowledge Setting}
\begin{tabular} {ccccc|cccc|cccc}
\toprule
& \multicolumn{4}{c|}{\textbf{Text-to-Text}}\centering & \multicolumn{4}{c|}{\textbf{Text-to-Image}}\centering & \multicolumn{4}{c}{\textbf{Image-to-Text}}\\\cline{2-13}
& \multicolumn{2}{c}{GPT2+Wiki103} & \multicolumn{2}{c|}{Falcon+XSum} & \multicolumn{2}{c}{SD1.5+MSCOCO} & \multicolumn{2}{c|}{SD2.1+CelebA-D.} & \multicolumn{2}{c}{LLaVA+COCO} & \multicolumn{2}{c}{MiniGPT4+CC\_SBU} \\\cline{2-13}
& \textbf{Compa.} & Classifier & \textbf{Compa.} & Classifier & \textbf{Compa.} & Classifier & \textbf{Compa.} & Classifier & \textbf{Compa.} & Classifier & \textbf{Compa.} & Classifier  \\
\midrule
ASR $\uparrow$        & $0.88$ & $0.90$ & $0.96$ & $0.97$ & $0.84$ & $0.61$ & $0.96$ & $0.58$ & $0.89$ & $0.91$ & $0.85$ & $0.86$ \\%
AUC $\uparrow$        & $0.92$ & $0.93$ & $0.99$ & $0.99$ & $0.92$ & $0.57$ & $0.99$ & $0.55$ & $0.97$ & $0.94$ & $0.94$ & $0.90$ \\
TPR@1\%FPR $\uparrow$ & $0.51$ & $0.53$ & $0.63$ & $0.65$ & $0.62$ & $0.11$ & $0.59$ & $0.04$ & $0.55$ & $0.55$ & $0.59$ & $0.56$ \\
\bottomrule
\end{tabular}
	\label{tab:ClassifierPartial}
\end{table*}

\begin{table*}[ht!]\scriptsize
	\centering
 	\caption{Results of Our Method Using the Comparison Rule vs. a Classifier Across Three Categories of Generative Models in the Zero Knowledge Setting}
\begin{tabular} {ccccc|cccc|cccc}
\toprule
& \multicolumn{4}{c|}{\textbf{Text-to-Text}}\centering & \multicolumn{4}{c|}{\textbf{Text-to-Image}}\centering & \multicolumn{4}{c}{\textbf{Image-to-Text}}\\\cline{2-13}
& \multicolumn{2}{c}{GPT2+Wiki103} & \multicolumn{2}{c|}{Falcon+XSum} & \multicolumn{2}{c}{SD1.5+MSCOCO} & \multicolumn{2}{c|}{SD2.1+CelebA-D.} & \multicolumn{2}{c}{LLaVA+COCO} & \multicolumn{2}{c}{MiniGPT4+CC\_SBU} \\\cline{2-13}
& \textbf{Compa.} & Classifier & \textbf{Compa.} & Classifier & \textbf{Compa.} & Classifier & \textbf{Compa.} & Classifier & \textbf{Compa.} & Classifier & \textbf{Compa.} & Classifier  \\
\midrule
ASR $\uparrow$        & $0.83$ & $0.85$ & $0.88$ & $0.90$ & $0.78$ & $0.55$ & $0.89$ & $0.53$ & $0.88$ & $0.89$ & $0.85$ & $0.86$ \\%
AUC $\uparrow$        & $0.90$ & $0.89$ & $0.96$ & $0.93$ & $0.90$ & $0.56$ & $0.94$ & $0.51$ & $0.96$ & $0.90$ & $0.91$ & $0.89$ \\
TPR@1\%FPR $\uparrow$ & $0.48$ & $0.45$ & $0.61$ & $0.58$ & $0.45$ & $0.08$ & $0.53$ & $0.07$ & $0.53$ & $0.48$ & $0.57$ & $0.51$ \\
\bottomrule
\end{tabular}
	\label{tab:ClassifierZero}
\end{table*}

\vspace{1mm}
\noindent\textbf{Impact of Using Neural Network-based Binary Classifier to Decide Membership Status.} In Step~4, our method determines the membership status of a target sample using a comparison-based decision rule (Eq.~\ref{eq:rule}). We investigate whether replacing this rule with a binary classifier affects the attack performance. Specifically, we train a neural network–based binary classifier consisting of three layers: an input layer with $d$ neurons corresponding to the dimensionality of the embedding vector, a hidden layer with $256$ neurons, and an output layer with two neurons.
The classifier is trained on a combined dataset composed of the synthetic dataset and the auxiliary dataset, where synthetic samples are labeled as members and auxiliary samples as non-members. We also experimented with deeper network architectures. However, increasing the number of layers does not lead to significant changes in performance.

The results are reported in Tables~\ref{tab:ClassifierPartial} and \ref{tab:ClassifierZero}. We observe that, for text-to-text and image-to-text generative models, the classifier-based approach achieves performance comparable to that of the comparison-based decision rule under both partial-knowledge and zero-knowledge settings. This suggests that, in these modalities, the embedding distributions of synthetic and auxiliary data are sufficiently separable for a supervised classifier to learn an effective decision boundary.
In contrast, for text-to-image generative models, the classifier-based approach exhibits significantly poorer performance than the comparison-based decision rule. This may be due to the higher complexity and variability of image embeddings, which often exhibit substantial intra-class variance and weaker separability. In such cases, the likelihood-based comparison rule explicitly models the underlying embedding distributions, making it more robust to distributional noise than the classifier.

\vspace{1mm}
\noindent\textbf{Summary.} Overall, our method achieves performance comparable to existing baselines under the partial-knowledge setting, while consistently outperforming them in most scenarios under the more challenging zero-knowledge setting. We observe that the amount of collected data has a limited impact on the effectiveness of our method, demonstrating strong scalability. In contrast, the choice of embedding extractor, the strategy used for embedding distribution estimation, the selection of the decision rule and threshold values, and the use of an alternative binary classifier for decision making all have a noticeable impact on performance.

\section{Extension Study}
The above experiments focus on membership inference against fine-tuning data. We extend our study to pre-training data. Although pre-training datasets are often considered less sensitive than fine-tuning datasets, data owners may still wish to revoke their data and thus need to verify whether their data were used during pre-training. Therefore, determining membership in pre-training data is also a practical problem.

To conduct this evaluation, we select one representative model for each class of generative models: GPT-2 (1.5B) \cite{GPT2} for text-to-text, Guided-Diffusion \cite{guided-diffusion} for text-to-image, and MiniGPT-4 \cite{Zhu24ICLR} for image-to-text, all of which explicitly disclose their pre-training data sources.
The results are reported in Table~\ref{tab:Pre-trained MIA}. We observe that our method remains effective in inferring membership with respect to pre-training data. This is because likelihood-based pre-training, similar to fine-tuning, drives the model’s output distribution to approximate the underlying pre-training data distribution. Thus, samples drawn from the pre-training set induce embedding distributions that are statistically closer to those of model-generated outputs than non-member samples. 

\begin{table}[H]\scriptsize
	\centering
 	\caption{Results of Our Method on Pre-trained Models}
    \resizebox{\columnwidth}{!}{
\begin{tabular} {ccc|cc|cc}
\toprule
& \multicolumn{2}{c|}{GPT-2}\centering & \multicolumn{2}{c|}{Guided-diffu.}\centering & \multicolumn{2}{c}{LLaVa} \\\cline{2-7}
& \makecell[c]{Partial\\Knowl.} & \makecell[c]{Zero\\Knowl.} & \makecell[c]{Partial\\Knowl.} & \makecell[c]{Zero\\Knowl.} & \makecell[c]{Partial\\Knowl.} & \makecell[c]{Zero\\Knowl.} \\
\midrule
ASR        & $0.86$ & $0.84$ & $0.96$ & $0.89$ & $0.85$ & $0.83$ \\%
AUC        & $0.91$ & $0.88$ & $0.99$ & $0.85$ & $0.88$ & $0.85$ \\
T@1\%F     & $0.45$ & $0.42$ & $0.69$ & $0.32$ & $0.40$ & $0.36$ \\
\bottomrule
\end{tabular}}
	\label{tab:Pre-trained MIA}
\end{table}

\section{Defense}
To evaluate the robustness of our method, we examine its performance in the presence of a defense mechanism. We adopt DP-SGD (Differentially Private Stochastic Gradient Descent) \cite{Abadi16CCS}, a widely used modality-agnostic defense against privacy leakage. By injecting calibrated noise into gradients during training, DP-SGD bounds the influence of any individual training record on the learned model parameters. Thus, the information that the model can reveal about a specific data point is formally limited, and the resulting privacy leakage is guaranteed not to exceed a predefined upper bound, regardless of how many outputs the adversary can query from the target model.
In our evaluation, the hyperparameter settings for DP-SGD are summarized in Table~\ref{tab:DefenseParameters}, and the corresponding experimental results are reported in Table~\ref{tab:Defense}. 

\begin{table}[ht!]\scriptsize
	\centering
 	\caption{Hyperparameter Settings for DP-SGD}
\begin{tabular} {ccccc}
\toprule
& \makecell[c]{Clipping\\ norm $C$} & \makecell[c]{Sampling\\rate $L/N$} & \makecell[c]{Noise\\scale $\sigma$} & \makecell[c]{Privacy\\budget $\epsilon$} \\
\midrule
Text-to-Text        & $1.0$ & $64/20000$ & $1.0$ & $3.0$ \\%
Text-to-Image        & $0.5$ & $4/59143$ & $2.0$ & $1.0$ \\
Image-to-Text     & $1.0$ & $32/60000$ & $1.0$ & $3.0$ \\
\bottomrule
\end{tabular}
	\label{tab:DefenseParameters}
\end{table}

\begin{table}[ht!]\scriptsize
	\centering
 	\caption{Results of Our Method with and without Defense}
    \resizebox{\columnwidth}{!}{
\begin{tabular} {ccc|cc|cc}
\toprule
& \multicolumn{2}{c|}{GPT2+Wiki103}\centering & \multicolumn{2}{c|}{SD1.5+MSCOCO}\centering & \multicolumn{2}{c}{LLaVa+COCO} \\\cline{2-7}
& \makecell[c]{Partial\\Knowl.} & \makecell[c]{Zero\\Knowl.} & \makecell[c]{Partial\\Knowl.} & \makecell[c]{Zero\\Knowl.} & \makecell[c]{Partial\\Knowl.} & \makecell[c]{Zero\\Knowl.} \\
\midrule
ASR        & $0.62/0.88$ & $0.57/0.83$ & $0.77/0.84$ & $0.62/0.78$ & $0.63/0.89$ & $0.56/0.88$ \\%
AUC        & $0.65/0.92$ & $0.55/0.90$ & $0.84/0.92$ & $0.71/0.90$ & $0.67/0.97$ & $0.60/0.96$ \\
T@1\%F     & $0.20/0.51$ & $0.09/0.48$ & $0.38/0.62$ & $0.19/0.45$ & $0.24/0.65$ & $0.22/0.93$ \\
\bottomrule
\end{tabular}}
	\label{tab:Defense}
\end{table}

\begin{figure}[ht]
\centering
	\includegraphics[scale=0.5]{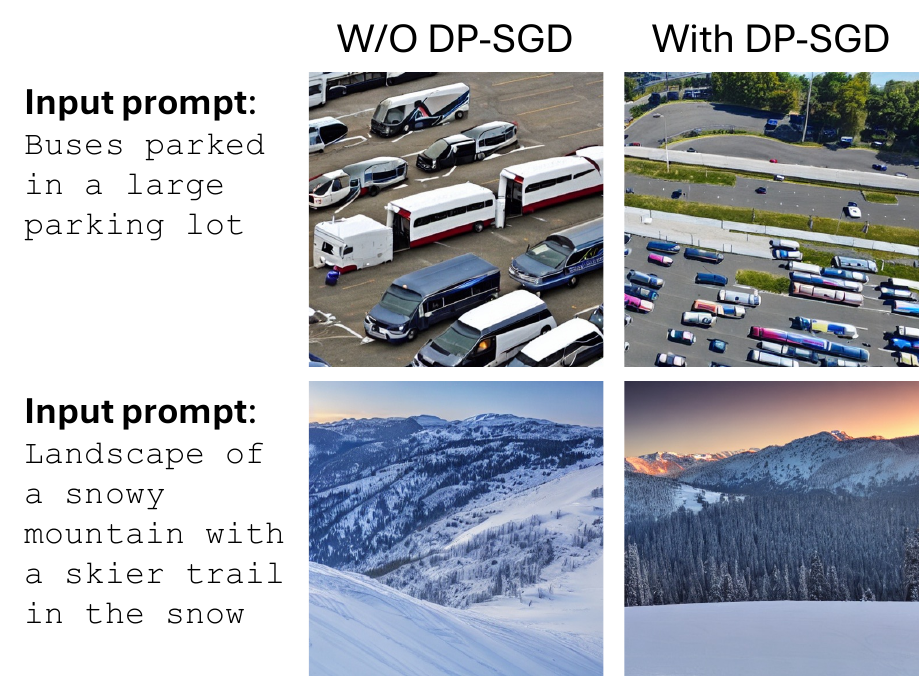}
	\caption{Examples of images generated by a text-to-image model trained with and without DP-SGD.}
	\label{fig:DPExample}
\end{figure}

As shown in Table~\ref{tab:Defense}, introducing DP-SGD during the fine-tuning process can effectively mitigate membership inference attacks. This is because the injected noise limits the influence of individual training samples on the learned model parameters, thereby reducing the distinguishability between member and non-member samples in the model’s output distribution.
However, a closer inspection of the qualitative examples in Figure~\ref{fig:DPExample} reveals that DP-SGD also degrades the overall generative quality of the models. For instance, in the first row of Figure~\ref{fig:DPExample}, the model trained without DP-SGD generates an image that closely aligns with the input prompt by producing an appropriate large parking lot. In contrast, when DP-SGD is applied, the generated image resembles a highway rather than a parking lot, indicating a loss of semantic fidelity.

In addition to DP-SGD, several defense mechanisms have also been proposed recently, such as \cite{Tran25ACL} for LLMs and \cite{Tran25ICLR} for diffusion models. However, these approaches are designed to specific modalities and therefore are not directly applicable to our modality-agnostic setting.

\section{Related Work}
Membership inference was first introduced by Shokri et al. \cite{Shokri17Oakland, Nasr19Oakland, Ye22CCS}. A common attack strategy trains a binary classifier using behavioral features, such as the target model’s output probabilities, collected from shadow models, and then predicts whether a given sample is within the model’s training set \cite{Hu22CSUR}. Subsequent research extended membership inference to a more practical label-only setting, where the adversary only observes hard-label outputs \cite{Choo21ICML,Li21CCS,Wu24ICLR,Li25USENIX}. In this setting, attacks typically estimate the distance between the input and the target model’s decision boundary: samples farther from the boundary are more likely to be members. However, applying membership inference to generative models is substantially harder because these models lack explicit decision boundaries and produce high-dimensional stochastic outputs.

Membership inference against generative models can be categorized based on whether the target data belong to the pre-training corpus or the fine-tuning dataset. Inferring membership in pre-training data is technically difficult because the pre-training corpus is extremely large, making leakage signals weak and noisy \cite{Duan24COLM,Puerto25NAACL}. In contrast, inferring membership in fine-tuning data is practically more important, since fine-tuning datasets are typically private or proprietary, whereas pre-training data are often publicly collected \cite{Hu25USENIX}. More recently, membership inference has also been extended to Retrieval-Augmented Generation (RAG) datastores \cite{Gao25CCS,Naseh25CCS}, but these studies mainly focus on LLMs and do not consider other classes of generative models. 

Existing work on fine-tuned generative models spans multiple modalities, including LLMs \cite{Fu24NeurIPS}, diffusion models \cite{Zhang24WACV,Zhai24NeurIPS, Pang25NDSS}, and vision-language or captioning models \cite{Samira25CVPR, Hu25USENIX}. These methods share a common principle: membership is inferred by comparing a target sample (or its output distribution) with a reference set of known non-members and measuring their distance. For example, Fu et al. \cite{Fu24NeurIPS} compare variations in probabilistic representations measured on the target model and a self-prompt reference model; Pang et al. \cite{Pang25NDSS} measure similarity between generated and target images; and Hu et al. \cite{Hu25USENIX} compare similarity score distributions between target and reference samples. 

In contrast, membership inference against pre-trained generative models explores different signals and designs. For example, Zhang et al. \cite{Zhang25ICLR} formulate membership detection as identifying local maxima under maximum-likelihood training. Other approaches rely on token-level semantic properties, such as lower perplexity for member sequences \cite{He25USENIX,Chang25EMNLP} or higher next-token probabilities for member tokens \cite{Li24NeurIPS}. 

\vspace{1mm}
\noindent\textbf{Summary.} Although existing membership inference methods are effective within their respective model types, there is currently no unified framework that can generalize across diverse generative models. 
This paper bridges this gap by proposing a unified framework, based on likelihood ratio testing, that applies to various types of generative models under a common inference paradigm.



\section{Conclusion}\label{sec:conclusion}
In this paper, we presented a unified study of membership inference attacks against generative models spanning text-to-text, text-to-image, and image-to-text modalities. We showed that likelihood-based training drives the output distribution of a generative model to approximate its training data distribution, and leveraged this modality-agnostic property to design a unified membership inference framework. Extensive experiments across diverse generative models and datasets demonstrated that our method achieves performance superior to state-of-the-art baselines tailored to individual modalities, while offering broader applicability and robustness. As future work, we plan to extend our study to additional modalities, such as text-to-video and text-to-audio generative models.

\bibliographystyle{IEEEtran}
\bibliography{references}

\setcounter{section}{0}
\renewcommand{\appendixname}{Appendix~\Alph{section}}


\section{Descriptions of Datasets}
\noindent\textbf{Wikitext-103} is a large-scale corpus containing over 100 million tokens extracted from verified Good and Featured Wikipedia articles, comprising 28,475 entries. It is widely used for training text generation models and for fine-tuning models on sequence completion and language modeling tasks.

\vspace{1mm}
\noindent\textbf{XSum}, short for Extreme Summarization, consists of 226,711 BBC news articles archived via Wayback from 2010 to 2017, covering a broad range of domains, including News, Politics, Sports, Weather, Business, Technology, Science, Health, Family, Education, and Entertainment. Each article is paired with a single-sentence summary, making the dataset particularly suitable for abstractive summarization tasks.

\vspace{1mm}
\noindent\textbf{MS COCO} is a large-scale dataset for object detection, segmentation, and image captioning, containing over 330k images with more than five human-written captions per image. It is widely used for training and evaluating vision–language models, particularly for tasks such as image captioning and text-to-image or image-to-text generation. 

\vspace{1mm}
\noindent\textbf{CelebA-Dialog} is a large-scale vision–language face dataset comprising 30,000 high-resolution facial images and 202,599 original images, each paired with detailed textual annotations describing facial attributes. It is broadly used for training and evaluating vision–language models.

\vspace{1mm}
\noindent\textbf{COCO 2017} is a split of the MS COCO dataset, containing over 160,000 images that depict complex everyday scenes with multiple objects and rich contextual interactions. It is widely used for training vision–language models across tasks such as image captioning and multimodal generation.

\vspace{1mm}
\noindent\textbf{CC\_SBU\_ALIGN} is a web-crawled vision–language dataset containing roughly 12 million image–text pairs formed by merging Conceptual Captions and SBU captions. It is commonly used for pretraining multimodal models for image–text alignment and image-to-text generation.

\end{document}